\documentclass{article}
\usepackage [utf8]{inputenc}
\usepackage{authblk}
\usepackage{setspace}
\usepackage [margin=1.25in]{geometry}
\usepackage{graphicx}
\graphicspath{ {./figures/} }
\usepackage{subcaption}
\usepackage{amsmath}
\usepackage{lineno}
\usepackage{color}
\usepackage{caption}
\usepackage{graphbox}
\usepackage{float}
\usepackage{svg}
\usepackage{enumerate}

\DeclareUnicodeCharacter{0301}{\'{e}}

\newcommand\keywords[1]{\textbf{Keywords}: #1}

\newtheorem{defn}{\textbf{Definition}}

\usepackage [style=ieee, citestyle=numeric-comp, sorting=none]{biblatex}
\title{Intelligent Computing: The Latest Advances, Challenges and Future}

\author [$\dag$1]{\normalsize Shiqiang Zhu\thanks{Corresponding author. Email: zhusq@zhejianglab.com; chenzuoning@vip.163.com; t.durrani@strath.ac.uk; whm\_w@163.com; wjx@mail.ndsc.com.cn; zhangty@shu.edu.cn; panyh@zju.edu.cn}}
\author [$\dag$1]{Ting Yu}
\author [1]{Tao Xu\thanks{These authors contributed equally to this work.}}

\author [1]{\\Hongyang Chen}
\author [2]{Schahram Dustdar}
\author [3]{Sylvain Gigan}
\author [4]{Deniz Gunduz}
\author [5]{Ekram Hossain}
\author [6]{Yaochu Jin}
\author [1]{Feng Lin}
\author [1]{Bo Liu}
\author [1]{Zhiguo Wan}
\author [1]{Ji Zhang}
\author [1]{Zhifeng Zhao}
\author [1]{Wentao Zhu}

\author [*7]{\\Zuoning Chen}
\author [*8]{Tariq S. Durrani}
\author [*9]{Huaimin Wang}
\author [*1,10]{Jiangxing Wu}
\author [*11]{Tongyi Zhang}
\author [*12,1]{Yunhe Pan}

\affil [1]{\normalsize Zhejiang Lab, Hangzhou, China}
\affil [2]{TU Wien, Austria}
\affil [3]{Sorbonne University, Paris, France}
\affil [4]{Department of Electrical and Electronic Engineering, Imperial College London, London, UK}
\affil [5]{Department of Electrical and Computer Engineering, University of Manitoba, Manitoba, Canada}
\affil [6]{Faculty of Technology, Bielefeld University,
Bielefeld, Germany}
\affil [7]{Chinese Academy of Engineering, Beijing, China}
\affil [8]{Department of Electronic and Electrical Engineering, University of Strathclyde, Glasgow, UK}
\affil [9]{National University of Defense Technology, Changsha, China}
\affil [10]{National Digital Switching System Engineering \& Technological R\&D Center, Zhengzhou, China}
\affil [11]{Materials Genome Institute, Shanghai University, Shanghai, China}
\affil [12]{Zhejiang University, Hangzhou, China}

\date{}

\begin{document}

\maketitle

%%%%%% Abstract %%%%%%
\begin{abstract}

Computing is a critical driving force in the development of human civilization. In recent years, we have witnessed the emergence of intelligent computing, a new computing paradigm that is reshaping traditional computing and promoting digital revolution in the era of big data, artificial intelligence and internet-of-things with new computing theories, architectures, methods, systems, and applications. Intelligent computing has greatly broadened the scope of computing, extending it from traditional computing on data to increasingly diverse computing paradigms such as perceptual intelligence, cognitive intelligence, autonomous intelligence, and human-computer fusion intelligence. Intelligence and computing have undergone paths of different evolution and development for a long time but have become increasingly intertwined in recent years: intelligent computing is not only intelligence-oriented but also intelligence-driven. Such cross-fertilization has prompted the emergence and rapid advancement of intelligent computing. Intelligent computing is still in its infancy and an abundance of innovations in the theories, systems, and applications of intelligent computing are expected to occur soon. We present the first comprehensive survey of literature on intelligent computing, covering its theory fundamentals, the technological fusion of intelligence and computing, important applications, challenges, and future perspectives. We believe that this survey is highly timely and will provide a comprehensive reference and cast valuable insights into intelligent computing for academic and industrial researchers and practitioners.

\end{abstract}


%%%%%% Main Text %%%%%%
\section{Introduction}
Human society is ushering into an intelligent society from an information society, in which computing has become a key element in formulating and promoting the development of society. In the new era of digital civilization with the internet of all things, traditional computing on data is far from being able to meet the growing endeavor for a higher level of intelligence by humans. The growing interest in intelligent computing, coupled with the development of computing science, the intelligent perception of the physical world, and the understanding of the cognitive mechanism of human consciousness, has collectively elevated the intelligence level of computing and accelerated the discovery and creation of knowledge.

Recent years have witnessed the rapid development of computing and information technology, from which artificial intelligence (AI) has been established as the frontier of human exploration of machine intelligence thanks to the unprecedented popularity and success of deep learning. Based on this, a series of breakthrough research results have been produced, including the convolutional neural network (CNN) proposed by Yann LeCun and contributions by Yoshua Bengio in the area of causal inference in deep learning~\cite{lecun1995convolutional,glorot2010understanding}.
Geoffrey Hinton, one of the pioneers of AI, proposed the deep belief network model and the backward propagation optimization algorithm in 2006~\cite{hinton2006fast}.
Jürgen Schmidhuber, another significant AI researcher, proposed the most widely-used recurrent neural network (RNN), long short-term memory (LSTM)~\cite{Schmidhuber1997LSTM}. It has been successfully applied in many fields to process entire sequences of data, such as speech, video, and time-series data.
In March 2016, AlphaGo, an AI Go program launched by DeepMind, battled with Lee Sedol, the world's top human Go master, and has attracted unprecedented worldwide attention. 
This epoch-making man-machine battle ended with a crushing victory of AI and has become a catalyst to push the wave of AI to a whole new level.

Another significant promoter of AI is the emergence of large pre-training models that have started to be widely used in natural language and image processing to deal with a wide variety of applications with the assistance of transfer learning. For example, GPT-3 has demonstrated that a big model, with a high level of structural complexity and a huge number of parameters, can improve the performance of deep learning. Inspired by GPT-3, a host of large-scale deep learning models have emerged~\cite{ethayarajh2019contextual,lepikhin2020gshard,bender2021dangers}.

Computational capacity is one of the important elements underpinning intelligent computing. Given the astronomical data sources, heterogeneous hardware configurations, and changing computing requirements in our information society, intelligent computing mainly meets the computational capacity requirements of intelligent tasks through vertical and horizontal architectures. Vertical architectures, which feature homogeneous computing infrastructure, mainly boost the computational capacity by applying intelligent methods to improve resource utilization efficiency. In comparison, horizontal architecture coordinates and schedules heterogeneous and wide-area computing resources to maximize the effectiveness of collaborative computing. For example, in April 2020, in response to the computing demands of COVID-19 research around the world, Folding@home achieved 2.5 Exaflops in computation by combining 400,000 computing volunteers in three weeks, more than any supercomputer in the world~\cite{beberg2009folding}. It is a success of horizontal computing collaboration to achieve such a huge computational capacity.

Despite the great success that has been achieved in intelligence and computing, we are still facing some major challenges in two respective areas, as follows:

\textbf{Challenges in intelligence.} 
AI using deep learning currently faces major challenges in interpretability, generality, evolvability, and autonomy. Most of the current AI technologies only work weakly compared to human intelligence and only work well in specific areas or tasks. Achieving strong and universal AI still has a long way to go. Finally, there are also major theoretical and technical challenges to upgrading from data-based intelligence to a more diverse form of intelligence, including perceptual intelligence, cognitive intelligence, autonomous intelligence, and human-machine fusion intelligence, to name a few.

\textbf{Challenges in computing.} 
The wave of digitalization brings an unprecedented growth of applications, connections, terminals, and users, as well as the amount of data generated, all requiring enormous computational capacity. For example, the computing power required for AI is doubling every 100 days and is projected to increase by more than a million times over the next five years. With the slowing down of Moore's Law, it becomes challenging to keep up with such a rapid increase in computational capacity requirements. In addition, the giant tasks in intelligent society rely on an efficient combination of various specific computing resources. Moreover, traditional hardware modes cannot fit intelligent algorithms well, which restricts software development. 

To date, there is no universally accepted definition of intelligent computing. Some researchers regard intelligent computing as the combination of AI and computing technology~\cite{bengio2009learning,schmidhuber2015deep,lecun2015deep}. It marks three different milestones of intelligent computing systems according to the development of AI. This perspective limits the definition of intelligent computing within the field of AI while ignoring the inherent limitations of AI and the vital role of ternary interactions between humans, machines, and things. Another school of thought views intelligent computing as computational intelligence. This area imitates human or biological intelligence to realize optimal algorithms to solve specific problems~\cite{poole1998computational} and treats intelligent computing primarily as an algorithmic innovation. However, it fails to consider the essential roles that the computing architecture and the internet-of-things (IoT) play in intelligent computing.

We present a new definition of intelligent computing from the perspective of solving complex scientific and societal problems considering the increasingly tight fusion of three fundamental spaces of the world, i.e., human society space, physical space, and information space.

\begin{defn}
\textit{(Intelligent Computing.)}
Intelligent computing is the area that encompasses the new computing theoretical methods, architecture systems, and technical capabilities in the era of digital civilization that supports the interconnection of all the world. Intelligent computing targets computational tasks with the minimum cost according to the specific actual needs, matching adequate computational power, invoking the finest algorithm, and obtaining optimal results.
\end{defn}

\begin{figure}[htbp]
	\centering
	\includegraphics[width=1\textwidth]{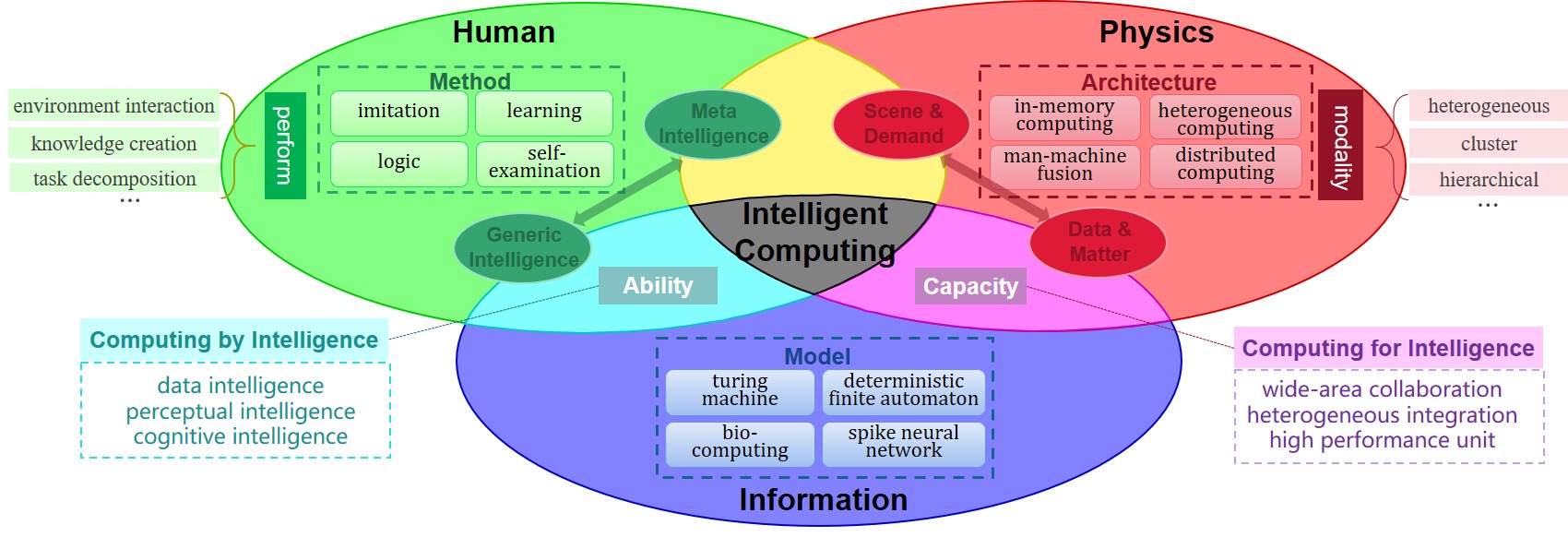}
	\caption{An overview of intelligent computing based on the fusion of human social space, physical space, and information space.}
	\label{fig:Overall_Structure}
\end{figure}

The new definition of intelligent computing is proposed in response to the fast-growing computing needs of the triple integration of human society, the physical world, and information space. Intelligent computing is human-oriented and pursues high computing capability, energy efficiency, intelligence, and security. Its goal is to provide universal, efficient, secure, autonomous, reliable, and transparent computing services to support large-scale and complex computational tasks. Figure~\ref{fig:Overall_Structure} shows the overall theoretical framework of intelligent computing, which embodies a wide variety of computing paradigms in support of human-physics-information integration.

First, intelligent computing is neither substitution nor a simple integration of the existing supercomputing, cloud computing, edge computing, and other computing technologies such as neuromorphic computing, optoelectronic computing, and quantum computing. Instead, it is a form of computing that solves practical problems by optimizing existing computing methods and resources systematically and holistically according to task requirements. In comparison, the major existing computing disciplines, such as supercomputing, cloud computing, and edge computing, fall into different domains. Supercomputing aims to achieve high computing power~\cite{caulfield2010future}, cloud computing emphasizes cross-platform/device convenience~\cite{armbrust2010view}, and edge computing pursues quality of service and transmission efficiency. Intelligent computing dynamically coordinates the data storage, communication, and computation among edge computing, cloud computing, and supercomputing domains. It constructs various cross-domain intelligent computing systems to support end-to-end cloud collaboration, inter-cloud collaboration, and supercomputing interconnection. Intelligence computing should make good use of existing computing technologies and, more importantly, promote the formation of new intelligent computing theories, architectures, algorithms, and systems. 

Second, intelligent computing is proposed to address problems in the future development of human-physics-information space integration. With the development of information technology applications in the big data era, the boundaries between physical space, digital space, and human society have become increasingly blurred. The human world has evolved into a new space characterized by the tight fusion of humans, machines, and things. Our social system, information systems, and physical environment constitute a large dynamically-coupled system in which humans, machines, and things are integrated and interact in a highly complex manner, which promotes the development and innovations of new computing technologies and application scenarios in the future.

We present the first comprehensive survey in the literature on intelligent computing, covering its theory fundamentals, the technological fusion of intelligence and computing, important applications, challenges, and future perspectives. To the best of our knowledge, this is the first review article to formally propose the definition of intelligent computing and its unified theoretical framework. We hope this review will provide a comprehensive reference and cast valuable insights into intelligent computing for academic and industrial researchers and practitioners. 

\begin{figure}[H]
	\centering
	\includegraphics[width=0.99\textwidth]{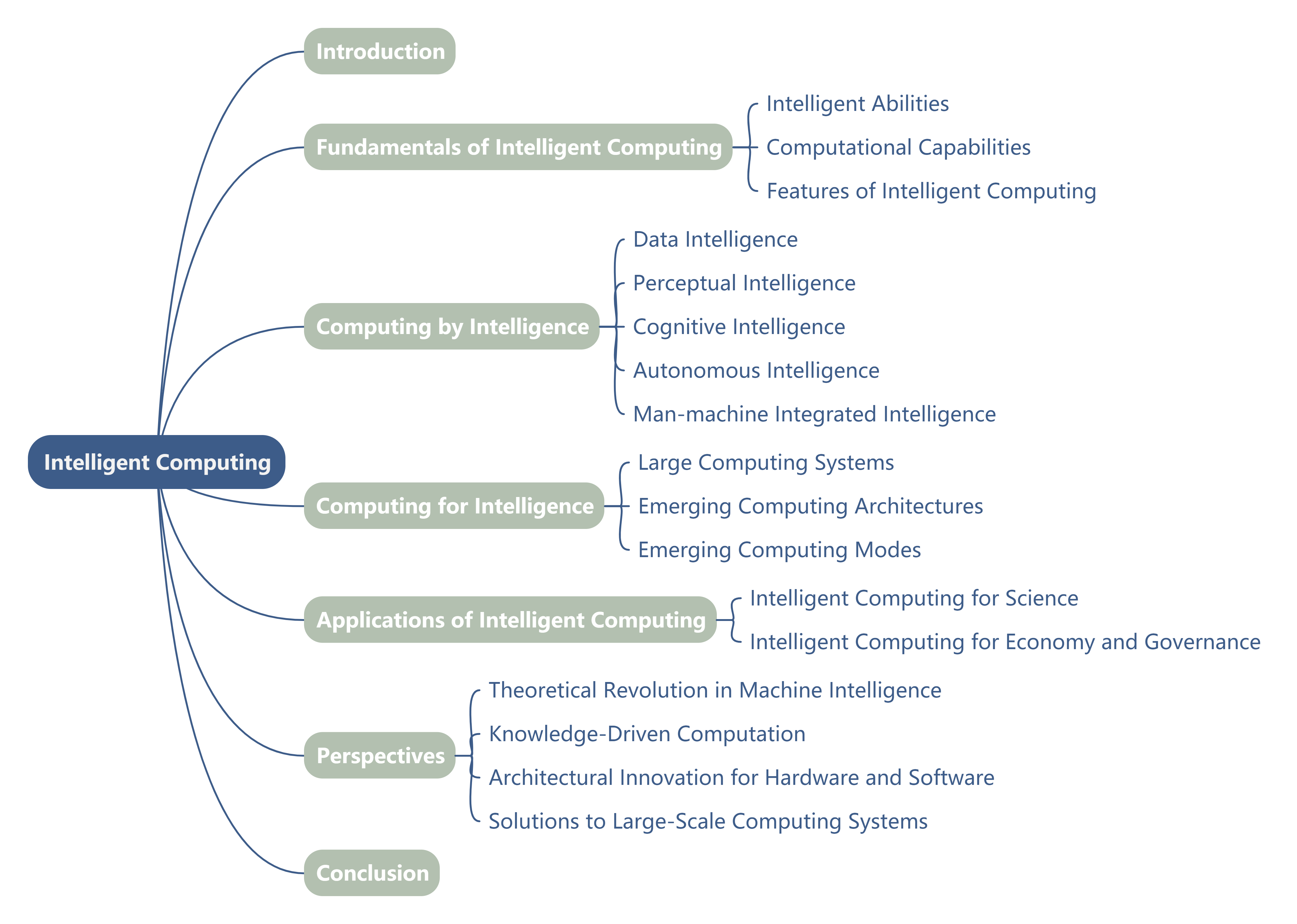}
	\caption{Main structure of the paper.}
	\label{fig:structure}
\end{figure}

The remainder of this paper is organized as follows.  
Section~\ref{Sec:FundamentalsOfIntelligentComputing} introduces the fundamentals of intelligent computing. 
Section~\ref{Sec:ComputingByIntelligence} summarizes the computing methods empowered by various intelligence aspects to boost computing performance. 
Section~\ref{Sec:ComputingForIntelligence} describes the large computing systems, emerging computing architectures, and modes to satisfy the urgent need for computing power from intelligent models. 
Section~\ref{Sec:Applications} exhibits several important applications of intelligent computing in both the scientific and social domains. 
Section~\ref{Sec:Perspectives} presents perspectives to cast light on the future development of intelligent computing. 
Finally, Section~\ref{Sec:Conclusions} concludes the paper.
Figure~\ref{fig:structure} shows the main structure of the paper.

%==============================================================================================
\section{Fundamentals of Intelligent Computing\label{Sec:FundamentalsOfIntelligentComputing}}

Intelligent computing is the general term for new computing theoretical methods, architectural systems, and technical capabilities in the era of digital civilization that support the interconnection of all things. It explores innovations in many classical and cutting-edge research fields to solve complex scientific and social problems. The basic elements of intelligent computing include human intelligence, machine capabilities, and the physical world composed of all things. In this section, we introduce the intelligent abilities and computational capacities expected of intelligent computing. We also describe the features of intelligent computing and how to combine intelligence and computation in the human-physics-information world.

\subsection{Intelligent Abilities}
In the theoretical framework, the human being is the core of intelligent computing and the source of wisdom, representing the original and inherent intelligence called \textbf{meta intelligence}. Meta intelligence includes advanced human abilities such as comprehension, expression, abstraction, inference, creation, and reflection, which contain the knowledge accumulated by human beings~\cite{taine1872intelligence,neisser1979concept,gardner2011frames,sternberg2021meta,barack2021two,dehaene2021consciousness,wilson2013embodied}.

\begin{figure}[htbp]
	\centering
	\includegraphics[width=1\textwidth]{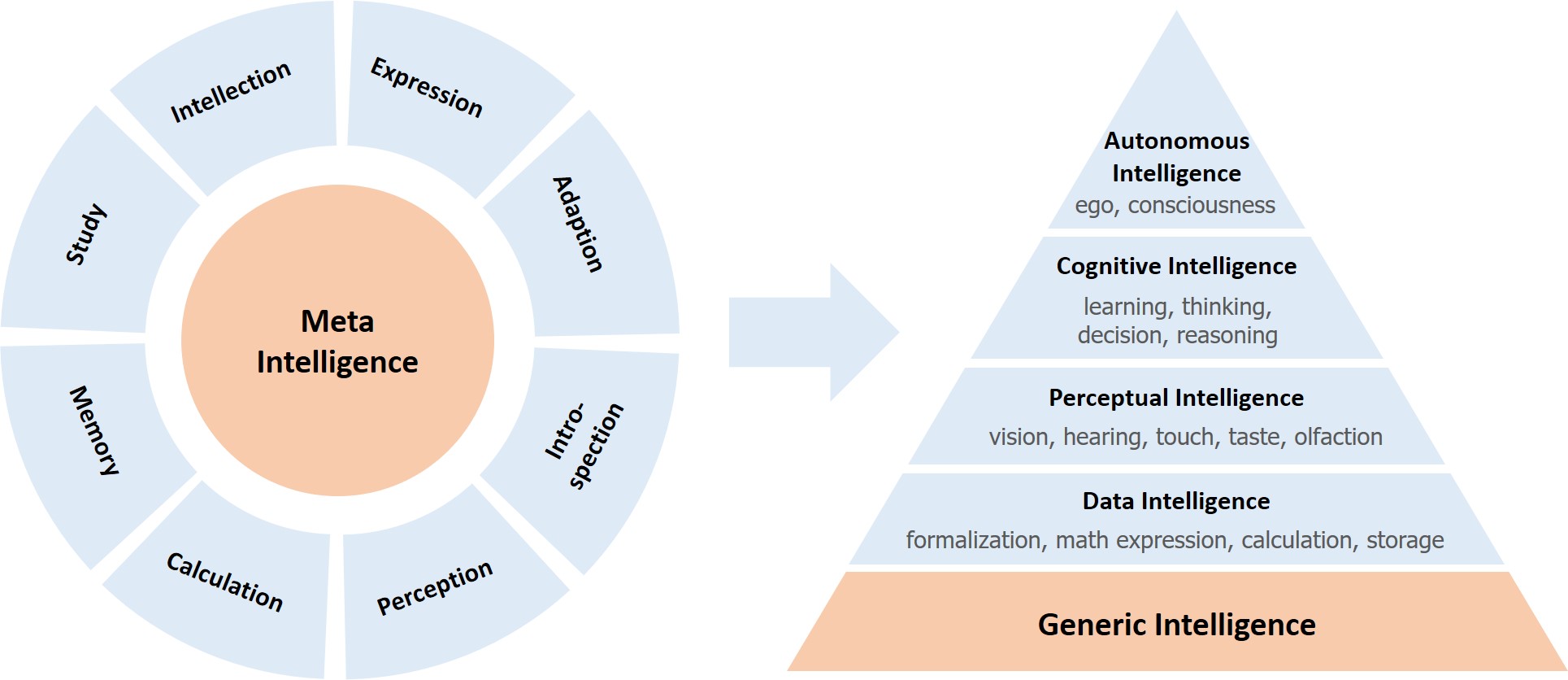}
	\caption{Meta intelligence and generic intelligence.}
	\label{fig:Meta_and_Generic}
\end{figure}

All intelligent systems are designed and built by humans. Therefore, in the theoretical system of intelligent computing, human wisdom is the source of intelligence, while computers are empowered by human intelligence. We call the intelligence of computers \textbf{generic intelligence}. Generic intelligence represents the ability of computers to solve complex problems with the wide extension, including natural language processing~\cite{chowdhary2020natural}, image recognition~\cite{dosovitskiy2020image}, speech recognition~\cite{nassif2019speech}, target detection and tracking~\cite{robin2016multi}, etc. The relationship between meta intelligence and generic intelligence is shown in Figure~\ref{fig:Meta_and_Generic} and is detailed in the following parts.

\subsubsection{Meta Intelligence}
Meta intelligence, also called natural intelligence, takes the carbon-based lives as the carrier and is produced by individuals and groups of organisms after millions of years of evolution. It includes biological embodied intelligence, brain intelligence (especially human brains), and swarm intelligence. Among them, biological embodied intelligence is widely obtained by organisms. They can receive the input of the environment to complete the specific tasks suitable for their physical form and perceive the changes in the environment to make the most advantageous intelligent behavior. Moreover, organisms may use tools and modify their environment to get a better chance of survival. The highest level of intelligence in nature is possessed by human beings, who have a solid ability not only to survive but also to feel and respond to the complex environment, for example, perceiving and identifying objects, expressing and acquiring knowledge, as well as complex reasoning and judgment. The intelligence of individual human beings is a comprehensive ability. More precisely,

\begin{itemize}
	\item They perform highly complex cognitive tasks.
	\item They can accomplish difficult learning, understand abstract concepts, do logical reasoning, and extract meaningful patterns.
	\item They can maximize the use and the transformation of the natural environment and can construct a cooperative of the order of millions of individuals.
	\item They have self-awareness.
\end{itemize}

Second, it is brain intelligence. The human brain is a complex and dynamic giant neural network system composed of a huge number of neurons. Its mystery has not been completely revealed yet, which leads to a vague understanding of intelligence. But in terms of the overall function, the intelligent performance of the human brain is recognizable. Abilities such as learning, discovering, and creating are clear manifestations of intelligence. Further analysis finds that the intelligence of the human brain and its occurrence is visible at its psychological level, expressed by some psychological activities and thinking processes~\cite{cacioppo1992social}. Thus, intelligence can be defined and studied on a macroscopic psychological level. We address the macro-psychological level of human intelligence performance as brain intelligence. Different areas of our brain in charge of varying perceptions or thinking functions cooperate as a unified whole.

Third, it is swarm intelligence. Swarm intelligence is a kind of high-level intelligence that low-level intelligent insects or animals usually generate through aggregation, coordination, adaptation, and other simple behaviors. Gerardo and Wang first proposed the definition of swarm intelligence~\cite{beni1993swarm}. The swarm intelligence optimization algorithm simulates the division and cooperation during the migration of natural organisms, foraging, and evolution. It stimulates the points in search space as individuals in nature and the search and optimization process as individuals' foraging or evolution process. The search and optimization swarm intelligence algorithms with the feature of generation and verification, which iteratively replace the less feasible solution with the better one, are inspired by ``survival of the fittest".

\subsubsection{Generic Intelligence}
Generic intelligence, also called machine intelligence, takes silicon-based facilities as the carrier and is produced by individuals and groups of computing devices. Biological intelligence can be transplanted to a computer on the following four levels: data intelligence, perceptual intelligence, cognitive intelligence, and autonomous intelligence. Data intelligence includes the ability of a computer to formalize, express, calculate, memorize and store data quickly. Perceptual intelligence refers to acquiring information such as voice, images, and video through various sensors and I/O devices. Cognitive intelligence is the ability to understand, think, reason, and explain. Autonomous intelligence stands for the ability of a machine to obtain a self-driven ego and consciousness. The four types of intelligence usually cooperate in conducting complex tasks.

Data intelligence emphasizes the realization of biological internal intelligent behavior through computational methods, programming the law of nature~\cite{Siddique2013Computational}. It is mainly guided by the theory of computing and relies on the basic storage and computing capabilities of the computer hardware to realize the original intelligence of data~\cite{turing1936computable}. Data intelligence uses the combination of five leading complementary technologies: symbolic and numerical computation for basic mathematical functions, fuzzy logic that enables computers to emulate human reasoning in linguistic terms; probabilistic methods based on big data and statistical law; artificial neural network construction that learns experiential data by models with a large number of parameters; evolutionary computation inspired from nature for search and optimization. Integrating data intelligence into these relatively mature branches forms various scientific methods.

Perceptual intelligence indicates machines with perceptual abilities like sight, hearing, and touch to reach the external world. Signals from the physical world are mapped to the digital world via microphones, cameras, and other sensors, using speech and image recognition. Machines communicate and interact similarly to humans via structured multi-modal real-world data~\cite{Scott2019From,Karakaya2019Electronic}. Perceptual intelligence completes the collection of large-scale data and features extraction of images, videos, audio, and other data types to complete structured processing. Computers present the data more comfortably for the user-connected hardware and software. For example, automatic driving utilizes light detection and ranging methods (lidars), other sensing devices, and AI algorithms for driving information computation. Face payment is a device through the perception of face data for identity confirmation.

Cognitive intelligence denotes machines with human-like logical thinking and cognition abilities, especially to actively learn, think, understand, summarize, interpret, plan, and apply knowledge~\cite{Emery2010Cognition}. The development of cognitive intelligence is composed of three levels. The first level is learning and understanding, such as text parsing, automatic marking, question understanding, etc. The second level is analyzing and reasoning, such as logical connecting and connotation abstracting. The third level is thinking and creating. 

Autonomous intelligence implies that the machine can act like a human being with a self-driven ego, emotion, and consciousness. It frees machines from heavy data dependency and enables them to learn the learning skills and renew their problem-solving abilities according to the change of environment. The final target of autonomous intelligence realizes self-learning, purposeful reasoning, and natural interaction with little or even any prior human programming.

\subsection{Computational Capabilities}
Intelligent computing is faced with the challenges of big scenes, big data, big problems, and ubiquitous requirements. The algorithmic models are becoming increasingly complex and require supercomputing power to support increasingly large model training. At present, computing resources have become a barrier to improving the level of computer intelligence research. With the development of intelligent algorithms, institutions rich in computing resources may form a systematic technological monopoly. The classical supercomputer is unsuitable for AI’s demand for computing power. Although algorithm optimization can reduce the need for computing power to a certain extent, it cannot fundamentally solve this problem. A complete optimization from multiple dimensions, such as architecture, acceleration module, integration mode, and software stack, is needed.

\begin{figure}[htbp]
	\centering
	\includegraphics[width=0.9\textwidth]{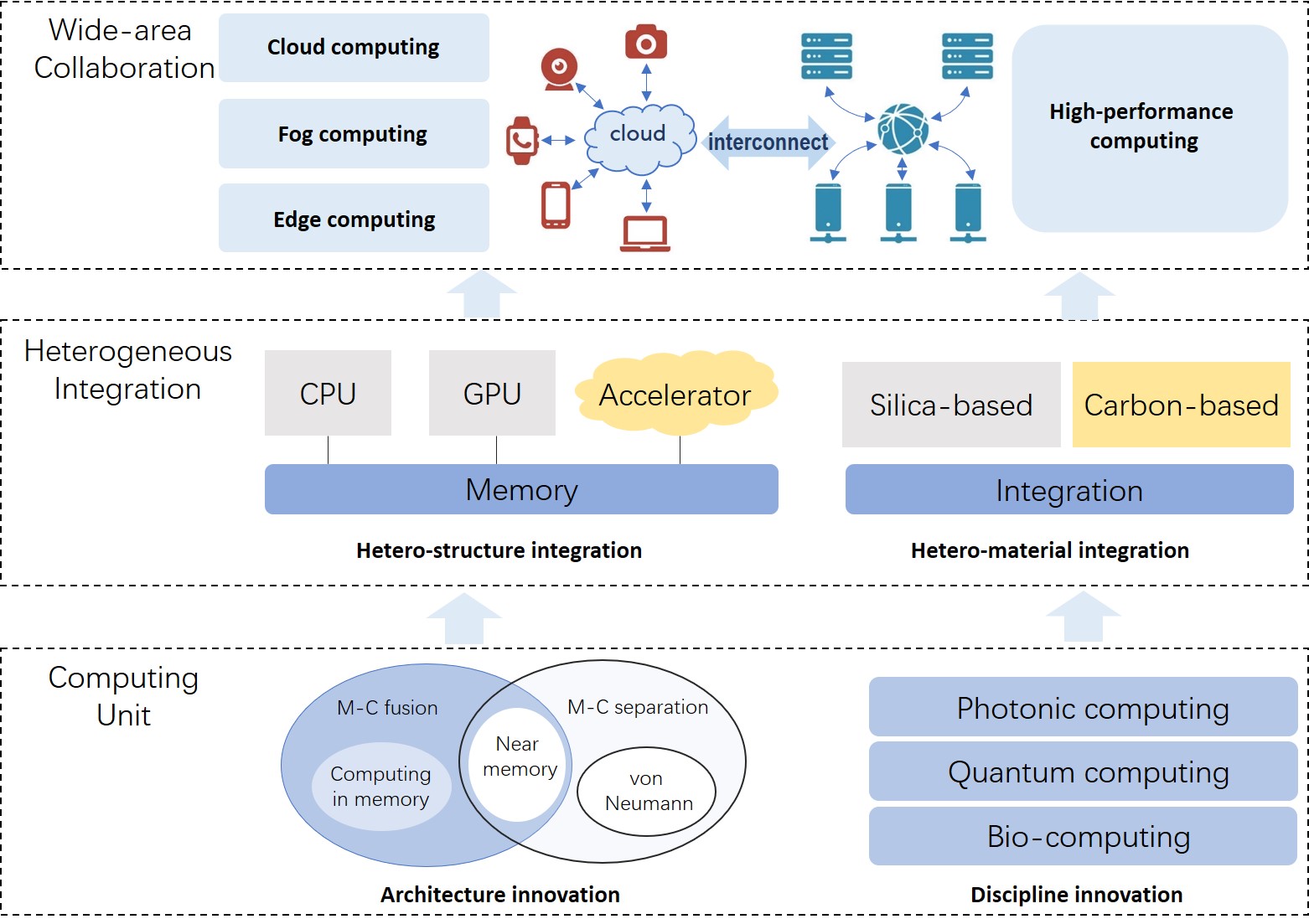}
	\caption{Computational capacities of intelligence computing.}
	\label{fig:Computing_Capacity}
\end{figure}

\subsubsection{Computing Units}
The most intuitive and effective method of wide-area collaboration is strengthening basic computing power through vertical lifting and horizontal expansion. At first, vertical improvement refers to the unit performance of computing components utilizing technological iteration, material innovation, and architectural design to increase the upper limit of the number of instructions a single chip can process per unit of time. Under the traditional von Neumann architecture, the performance limit is broken through technical means to meet the computational performance requirements for graphics rendering and deep learning training tasks. These chips have enough power to support the advanced deep learning algorithms and plug-and-play of the mainstream computers of today.

As Moore’s Law slows down, traditional von Neumann computing models will soon face a performance ceiling. The end of Dennard’s scaling law will result in power consumption and heat dissipation problems becoming obstacles to processor frequency growth. Traditional storage devices cannot obtain high speed and high density simultaneously. The existing computing-centered von Neumann architectures rely on a hierarchical storage structure composed of memory and storage to maintain a balance between computing performance and storage capacity. The structure needs to frequently deliver data between the processor and memory so that computing efficiency decreases and bandwidth is limited, causing the ``storage wall" problem. Under such circumstances, memory computing becomes an effective measure to break through the bottleneck of the von Neumann system and improve overall computing efficiency.

To break through the limitations of traditional chip architecture, intelligent computing needs to explore new chips through horizontal expansion. Given the challenges faced by traditional electronic computing methods, the emergence of integrated photonics, which is built on multidisciplinary areas such as materials science, photonics, and electronics, is exciting. Based on the principle of quantum mechanics, quantum computing realizes quantum parallel computing by using quantum superposition, entanglement, and quantum coherence, which fundamentally changes the traditional computing concept. Biocomputing is developed based on the inherent information processing mechanism of biological systems. In contrast with traditional computing systems, its structure is generally parallel and distributed.

Since the diversified computing power of data centers has become a trend, generalizing and specialized computing chips will develop in parallel. The traditional technology with CPUs and other general computing chips as the core is quite challenging to meet the requirements of mass data processing. The fusion of general technology and special technology has become a promising approach.

\subsubsection{Heterogeneous Integration of Systems}
Heterogeneous integration includes hetero-structure integration and hetero-material integration. Hetero-structure integration mainly refers to encapsulating chips manufactured by multiple processing nodes into one package to enhance functionality and performance. It can encapsulate components manufactured by different processes, functions, and manufacturers. The progress of semiconductor technology has reached the physical limit, and the circuit has become more complex. The traditional way, which improves the computational capacity by increasing the CPU clock frequency and the core number, has met the heat dissipation and energy consumption bottleneck. Heterogeneous integration can solve the problem. Through hetero-structure integration, different computing units adopt a hybrid computing architecture. Each computing unit performs its adequate task, effectively improving computing performance. Hetero-structure integration can be divided into the chip level and system level. Chip-level hetero-structure integration is a method to integrate different chips to improve the overall chip efficiency. Currently, the main-stream hetero-structure integration technologies mainly include 2D/3D packaging, Chiplet, etc. System-level hetero-structure integration provides various computing types in the form of single-machine multi-processor and multi-machine, including single-machine multi-computing, single-machine hybrid computing, homogeneous heterogeneous multi-machine, and heterogeneous multi-machine.

Hetero-material integration refers to the integration of semiconductor components of different materials for small size, good economy, high flexibility, and better system performance. It is considered an innovative exploration to use biological components for information processing and computation through the integration of silicon and carbon. As a basic unit of biological structure and function, a single cell is an independent and orderly system that can give feedback and self-regulate in response to external stimuli and environmental changes. Its operating mechanism has undergone long-term evolution and thus can meet its metabolic needs. As a natural storage carrier of genetic information, DNA in cells has high storage capacity and density characteristics. Over hundreds of millions of years of evolution, biological cells have also optimized their biochemical processes to minimize the energy consumption of metabolic processes. Biological components show the potential for storage capacity, computational parallelism, and ultra-low computing power consumption. The effective integration of carbon-based and conventional silicon-based chips is expected to reach new heights in computing power, storage density, and energy efficiency.

\subsubsection{Wide-Area Collaboration of Resources}
The data in the human-machine-thing integration scenario of wide-area collaboration has the characteristics of wide geographical distribution, complete scene coverage, and enormous collective value. The real-time acquisition, perception, processing, and intelligent data analysis from the time dimension require the support of distributed parallel computing power available anywhere. Thus, wide-area collaboration is highly needed. Wide-area collaborative computing connects computing resources such as high-performance computing (HPC), cloud computing, fog computing, and edge computing cost-effectively. It achieves automated horizontal expansion of supply-side resources. Demand-side diversified tasks require a new computing infrastructure across management domains and on-demand collaboration in a low-cost, efficient, and highly trusted way. Led by intelligent computing scenarios that support the interconnection of all things, wide-area collaborative computing supports vertical and horizontal convergence of resources in an autonomous and peer-to-peer manner. Significant challenges of intelligent matching, scheduling, and collaborating resources and tasks across domains exist in building a new infrastructure of secure and reliable intelligent computing.

Improving the computational capacity of wide-area collaboration mainly focuses on two scientific issues: the mechanism of the wide-area collaboration model and the realization of the wide-area collaboration system. The wide-area collaboration model primarily emphasizes resource abstraction, decoupling, and encapsulation and building a software-defined programmable entity abstraction method to shield the heterogeneity of device, computing, and data resources. It constructs a software-defined programmable collaboration model, rules, and processes based on interconnectivity and interoperability to support forming interaction orders of computing, data, and devices across independent stakeholders. The wide-area collaboration system mainly focuses on the task decomposition and scheduling of diversified jobs on the demand side; cross-domain fusion and management of computing and data resources; data privacy protection, identity trust, and security protection in an open environment; multi-dimensional intelligent operation and maintenance monitoring crack the invisibility of resource distribution, use, and business execution.

\subsection{Features of Intelligent Computing}

In this subsection, we first introduce the major characteristics of intelligent computing development and then reveal the innovation paths to obtain these critical characteristics.

\subsubsection{Objective-Oriented Intelligent Computing}
\begin{figure}[htbp]
	\centering
	\includegraphics[width=0.7\textwidth]{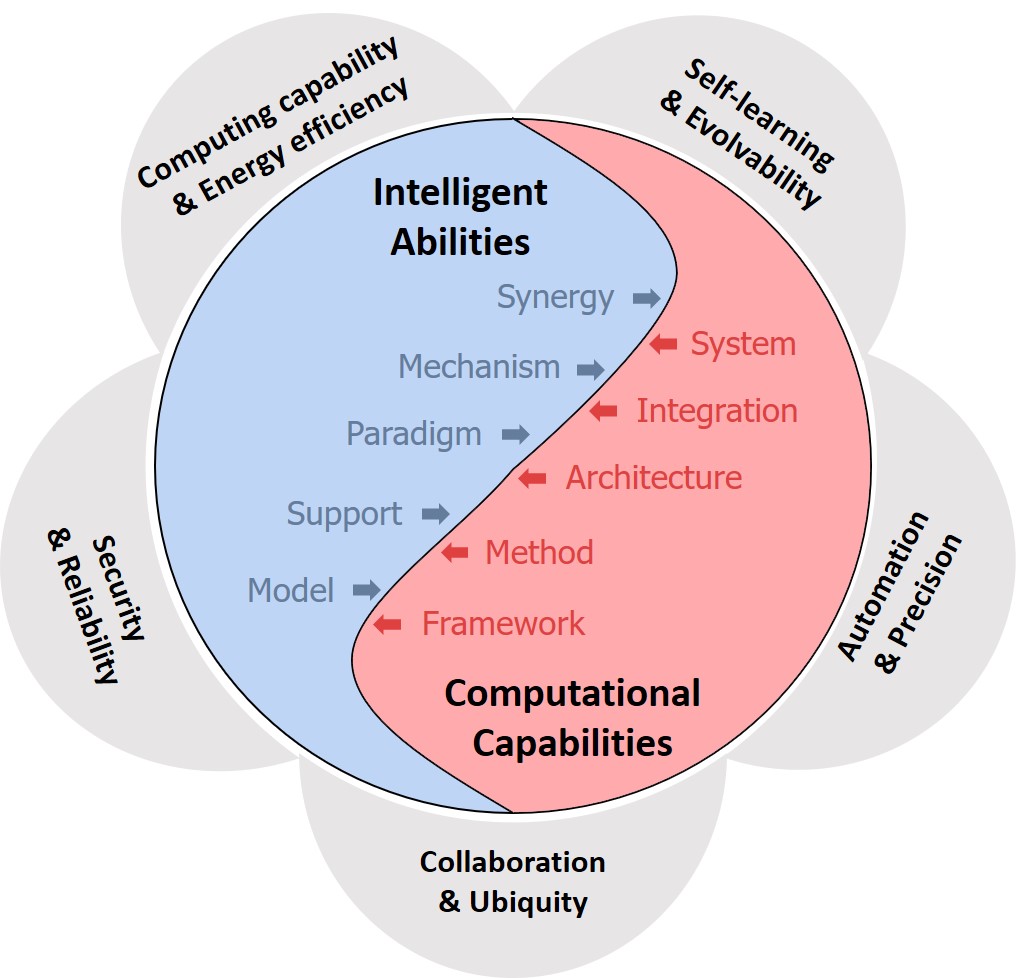}
	\caption{Features of intelligent computing.}
	\label{fig:Features}
\end{figure}

As depicted in Figure~\ref{fig:Features}, intelligent computing has the following characteristics: self-learning and evolvability in theoretical techniques, high computing capability and high energy efficiency in architecture, security and reliability in systematic methods, automation and precision in operational mechanisms, and collaboration and ubiquity in serviceability.

\textbf{Self-learning and evolvability.} 
Inspired by brain neuroscience, intelligent computing develops several novel techniques, such as neuromorphic computing and biological computing, to achieve breakthroughs in the principles and models of von Neumann’s computer structure. Self-learning refers to obtaining experience by mining rules and knowledge from massive data and optimizing the calculation paths with usable results. At the same time, evolvability represents a heuristic self-optimization ability that simulates the evolutionary process of organisms in nature, where the machines learn from the environment and subsequently make self-adjustments to adapt to the environment.

\textbf{High computing capability and high energy efficiency.} 
Aiming to exceed the traditional von Neumann’s architecture, intelligent computing evolves to new computing architectures concerning processing-in-memory, heterogeneous integration, and wide-area collaboration. High computing power refers to the computing capability that meets the needs of an intelligent society and serves as infrastructures like water and electricity. Moreover, high energy efficiency aims to maximize computing efficiency and reduce energy consumption as much as possible to ensure efficient processing of big data with large-scale characteristics, complex structure, and sparse value.

\textbf{Security and reliability.}
Intelligent computing supports cross-domain trust and security protection for large-scale ubiquitous interconnected computing systems. It establishes independent and controllable trusted security technology and support systems, realizing data fusion, sharing, and opening. High trust refers to the trust of identity, data, computing process, and computing environment through trusted hardware, operating system, software, network, and private computing. Particularly, high security means network security, storage security, content security, and circulation security of computing systems that can be guaranteed by integrating various privacy protection technologies.

\textbf{Automation and precision.}
Intelligent computing is task-oriented; it matches computing resources and realizes automatic demand calculation and precise system reconstruction. The system architecture is constantly adjusted to the task execution. Directed coupling reconstruction is performed at the software and hardware levels. Automation of the computing process includes automatic resource management and scheduling, automatic service creation and provision, and automatic management of the task life cycle, which is the key to evaluating the friendliness, availability, and service of intelligent computing. The precision of computing results anchors computing services; besides, it solves difficulties, including fast processing of computing tasks and timely matching of computing resources.

\textbf{Collaboration and ubiquity.} 
Intelligent computing integrates existing technologies to promote the penetration and integration of the physical, information, and social space using the various perception ability of heterogeneous elements, complementary computational resources, and the collaboration and competition of computational node functions. Cooperation between humans and machines improves intelligence levels in intelligent tasks, and ubiquity enables computing to be conducted everywhere by combining intelligent computing theoretical methods, architectural systems, and technical approaches.

\subsubsection{Fusion of Intelligence and Computation}
Intelligent computing includes two essential aspects: intelligence and computation, which complement each other. Intelligence facilitates the development of computing technologies, while computation is the foundation of intelligence. The paradigm of high-level intelligence technologies that improve the performance and efficiency of computing systems is ``computing by intelligence". The paradigm of efficient and powerful computational technologies that support the development of computer intelligence is ``computing for intelligence". The two basic paradigms are innovated from five aspects to improve computing power, energy efficiency, data usage, knowledge expression, and algorithm capabilities and achieve ubiquitous, transparent, reliable, real-time, and automatic services.

\textbf{The paradigm of computing by intelligence.} 
The computing power demand of complex models has exceeded that of general computers by one or two orders of magnitude. Moreover, there is a considerable gap between the underlying computing mechanism of traditional computers and the computing mode of intelligent models, resulting in low computing efficiency. The paradigm of computing by intelligence includes new models, support, paradigms, mechanisms, and synergy that utilize intelligent approaches to improve computing capability and efficiency. 

%\paragraph{New Model.}
Currently, intelligent systems can only handle specific tasks in a closed environment since they lack common sense, intuition, and imagination. Research on neuromorphic computing, graph computing, biological computing, and other new computing models is conducted to analyze the human brain, biological, and knowledge computing mechanisms. These new models can effectively improve cognitive understanding and reasoning learning abilities, adaptability, and the generalization effect for intelligent algorithms.

%\paragraph{New Support.}
Due to the limitations of computing system architecture and lack of end-to-end computational capacity, the computing and response speed of the current computing system needs further improvement. Intelligent computing can improve the real-time performance of the computing system by utilizing new computing support technologies, e.g., processing-in-memory, edge computing, and online learning. Moreover, new technologies, such as the fusion of perception and computation and processing localization, have also become promising research hotspots.

%\paragraph{New Paradigm.}
The deep integration of the triple-space leads to the diversities of computing tasks, so the computing scenarios and data are more unstructured, and the solutions of the tasks are more complex and challenging. Thus, the new computing paradigm enables the analysis and modeling of unstructured scenes and the adaptive processing of unstructured data. It achieves transparency computing through an automatic and intelligent process that concludes tasks understanding, decomposition, solving, and resource allocation.

%\paragraph{New Mechanism.} 
Intelligent computing explores new computing mechanisms, such as hardware and software refactoring and cooperative evolution, to deal with different types of tasks. During the execution of intelligent processes, the new mechanism configures the hardware through the organization of computing resources with different granularity and functions. The new mechanism will form an automatic computing system with autonomous learning and evolutionary iteration applying intelligent technologies, including elasticity design of software and hardware, flexible cooperation of algorithms and models, and adaptive allocation of data and resources.

%\paragraph{New Synergy.}
New synergy computing architectures, such as human-computer interaction, swarm intelligence, and human-in-loop, combine human perception and cognitive ability with the operation and storage ability of a computer. And such new architectures are effective in improving the sensing and reasoning ability of computers. Machines can have ultra-high computing speed and accuracy and also efficiently obtain information from the physical environment through various sensors. However, they cannot independently analyze the information and execute complex tasks. Notably, humans can study the physical environment at a higher level, recognize the laws of the physical world, and transfer knowledge to machines in human-computer interactions.

\textbf{The paradigm of computing for intelligence.}
The heterogeneity and complexity of hardware architecture hinder computing capability integration and service quality improvement. The computing for intelligence paradigm designs new frameworks, methods, integration, architectures, and systems to improve the intelligence level and provide ubiquitous, transparent, automatic, real-time, and security of computing services.

%\paragraph{New Framework.}
Given the diversity of intelligent devices, the discretization of computing resources, and the complexity of network connections, it becomes more difficult to integrate hardware and effectively improve intellectual computing power. The innovation of the computing framework adopts a non-von Neumann structure, which contains memory processing, heterogeneous integration, and wide-area collaboration. Additionally, the dedicated hardware building blocks are designed. To meet the demand for high computing power, the computing structure of the chip and system senses, schedules, and computing resources management are optimized.

%\paragraph{New Method.}
A new way to improve limited energy efficiency is to apply new computing methods, such as biocomputing and neuromorphic computing, to study low-power characteristics of living matter. With a power consumption of only 20W, the learning procedure of the human brain is much more effective than any AI. By learning the computing methods of biological and human brains and designing new computing hardware and software, intelligent computing may dramatically increase computing efficiency and decrease the energy consumption of the computing process.

%\paragraph{New Integration.}
Intelligent computing promotes effective coordination and develops integration of human-machine-thing by improving machine intelligence, sensing ability, and response to emergencies. Through the comprehensive connection of humans, machines, and things, intelligent computing creates a deeply integrated computing mode. The symbiotic integration, cooperation, and complementation of human-machine-thing provide more comprehensive, thoughtful, and accurate intelligent services for human beings.

%\paragraph{New Architecture.} 
The traditional cluster-centered computing architecture cannot provide timely services for the edge terminal nodes and users. 
New distributed computing architectures such as end-to-end cloud and wide-area collaboration are adopted to effectively integrate supercomputing, cloud computing, edge computing, and terminal computing resources. The problem of centralizing computing is solved through intelligent task decomposition to achieve efficient and ubiquitous computing services.

%\paragraph{New System}
In a human-physical-information integrated computing environment, more malicious attack surfaces could be exploited, making the system more vulnerable. Meanwhile, the massive multi-heterogeneous information also brings data security and privacy problems. A new secure and trusted intelligent computing system is established to tackle these problems by constructing endogenous secure methods and trusted computing mechanisms. It ensures the security and trust of the computing process, identity, data, and results.

%%%%%%%%%%%%%%%%%%%%%%%%%%%%%%%%%%%%%%%%%%%%%%%%%%%%%%%%%%%%%%%%%%%%%%%%%%%%%%%%%%%%%%%%%%%%%%%%%%%%%%%%%%%%%%%%%%%%%%%%%%%%%%%%%%%%%%%%%%%%%%%%%%%%%%%%%%%%%%%%%%%%%%%%%%%%%%%%%%%%%%%%%%%%%%%%%%%%%%%%%%%%%%%%%%%%%%%%%%%%%%%%%%%%%%%%%%%%%%%%%%%%%%%%%%%%%%%
\section{Computing by Intelligence \label{Sec:ComputingByIntelligence}}

In this section, we describe the four intelligent abilities of computers and the mode of integrated intelligence. For each intelligence, we present significant progress in typical research areas. 

\subsection{Data Intelligence}

Improving the universality of computing is critical for intelligent computing. Problems in real-world scenarios, such as analog, graph, etc., need various computations. Another critical point of intelligent computing is how to improve the intellectual level of computing. Empirically, we all need to learn from intelligent creatures in nature with no exception for computation, such as the three classical intelligence methods: artificial neural network, fuzzy system, and evolutionary computing. The theory of intelligent computing includes but is not limited to the above types of computing to achieve a high level of ubiquity and intelligence.

\subsubsection{Analog Computing} 
The simulated calculation model can have a wide range of complexity, in which the slide rule and nomogram are the simplest types. In contrast, the naval gun control calculation and large-scale hybrid digital/analog calculation are more complex~\cite{Analogue2021Farzad,haensch2018next}. Process control and protection relay system use analog calculations to control and protect. The simulated calculation is of various types according to different calculation methods and application fields~\cite{youssefi2016analog,abdollahramezani2015analog,fuller2017li,bayat2015redesigning,zhou2020optical}. 

Compared with common digital computing, analog computing has both advantages and disadvantages. It achieves real-time operation in both computation and analysis, which can operate multiple values at the same time. It has a simple hardware design with no sensor requirement to convert input/output to digital electronic form and less bandwidth consumption. Nevertheless, analog computing has poor transportability. An analog computer can only solve a preset type of problem. Since the calculation is affected by environmental factors, it is usually challenging to obtain the exact solutions.

\subsubsection{Graph Computing} 
Mathematically, graph theory is the study of the graph, which is the mathematical structure used to model pairwise relationships between objects. The graph is essential to mathematical theories such as algebra, geometry, group theory, and topology. Graph processing uses graphs as data models to express and solve problems, and it can completely depict the relationship between things. Graph computing architecture also shows excellent application value in mathematical and related fields such as dynamic systems and complexity computing. In recent years, graph processing has focused on the field of large-scale graph data and aimed to achieve data storage, management, and efficient computing of large-scale graphs.

\begin{figure}[htbp]
	\centering
	\includegraphics[width=1\textwidth]{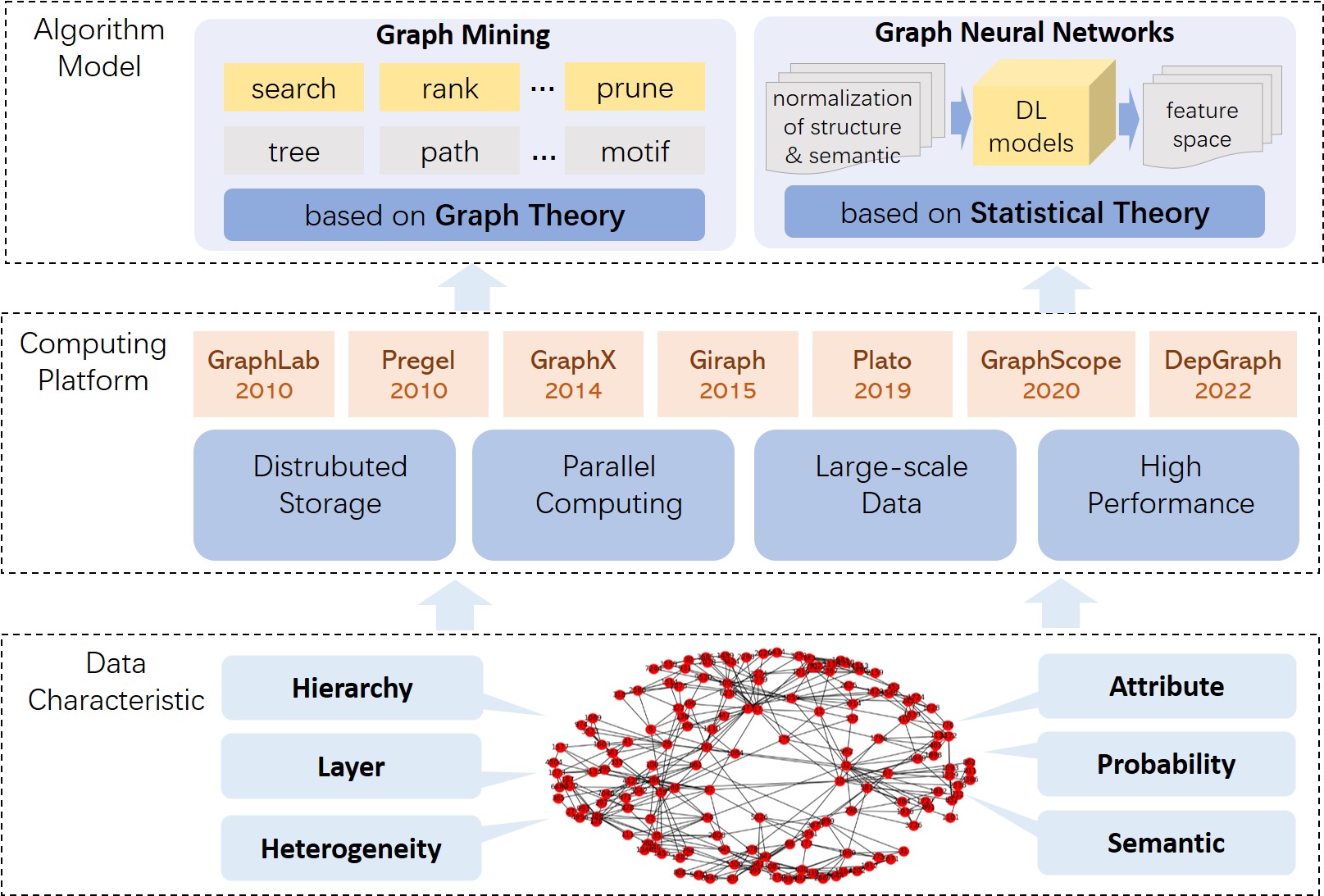}
	\caption{Technology architecture of graph computing.}
	\label{fig:Graph_Computing}
\end{figure}

Traditional graph processing is based on graph theory. It investigates various questions, including search, mining, statistics, analysis, transformation, and other issues based on the fundamental properties of the graph structure. These questions often take the query, traversal, sorting, and set operation of nodes or edges as basic operators to calculate the exact or optimal approximate solution of the target result.

With the increase of graph data scale, the mainstream research direction combines graph processing with big data-related technologies, such as distributed computing, parallel computing, stream computing, incremental computing, etc. Some node-centric parallel graph processing engines based on batch message processing have been designed to specifically handle parallel graph processing tasks, such as Pregel~\cite{malewicz2010pregel}, Giraph~\cite{ching2015one}, Graphx~\cite{ gonzalez2014graphx}, GraphScope~\cite{Fan2021GraphScope}, and DepGraph~\cite{Zhang2021DepGraph}. In addition to the data volume expansion, some studies expanded the graph data model. Attributes, labels, probability, hierarchy, and other characteristics are introduced to address more complex application requirements and modeling challenges. With the development of database technology, graph databases have risen strongly with their comprehensive application scenarios and flexible model expression. They have become one of the four core members of the emerging NoSQL data family. The expansion of the graph model is reflected in tasks such as storing and managing graph data. On the other hand, the connotation expansion of the model brings an increment in algorithm complexity for the algorithm evolution of various graph processing. Furthermore, it makes the results of computing problems more suitable for application in natural scenes.

In recent years, with the development of deep learning technology, graph data has been used as the input of the neural network models, and various types of graph neural network models and calculation methods have been derived. Graph neural networks, such as graph convolutional networks~\cite{kipf2016semi}, recurrent neural networks~\cite{li2015gated}, graph attention networks~\cite{velivckovic2017graph}, graph residual networks~\cite{li2019deepgcns}, evolve from the technical framework of deep learning. They transplant from structural data to semi-structural data retaining the characteristics of the structure and function of the model. At the same time, the core mathematical model is improved for the graph data structure to achieve good computational results in classification, prediction, anomaly detection, and other issues.

\subsubsection{Artificial Neural Network}
Since the 1980s, engineering techniques have been used to simulate the structure and function of the human brain's nervous system to construct artificial neural networks. The artificial neural network imitates the connection of brain neurons through many nonlinear processors. It simulates the signal transmission behavior between synapses with the input and output between computing nodes. W. S. McCulloch of psychology and W. Pitts of mathematical logic developed the neural network and mathematical model known as the MP model~\cite{mcculloch1943logical} in 1943. They suggested using the MP model as the basis for rigorous mathematical description and network structure for neurons. Artificial neural network research was founded on their discovery that a single neuron could carry out logical operations. The BP algorithm was created by Rumelhart, Hinton, and Williams in 1986~\cite{rumelhart1986learning}. The back propagation of loss and the forward propagation of signals make up the BP algorithm. Because the multi-layer feed-forward network is often trained by the back propagation algorithm, multi-layer feed-forward networks are often referred to as BP networks.

\begin{figure}[htbp]
	\centering
	\includegraphics[width=1\textwidth]{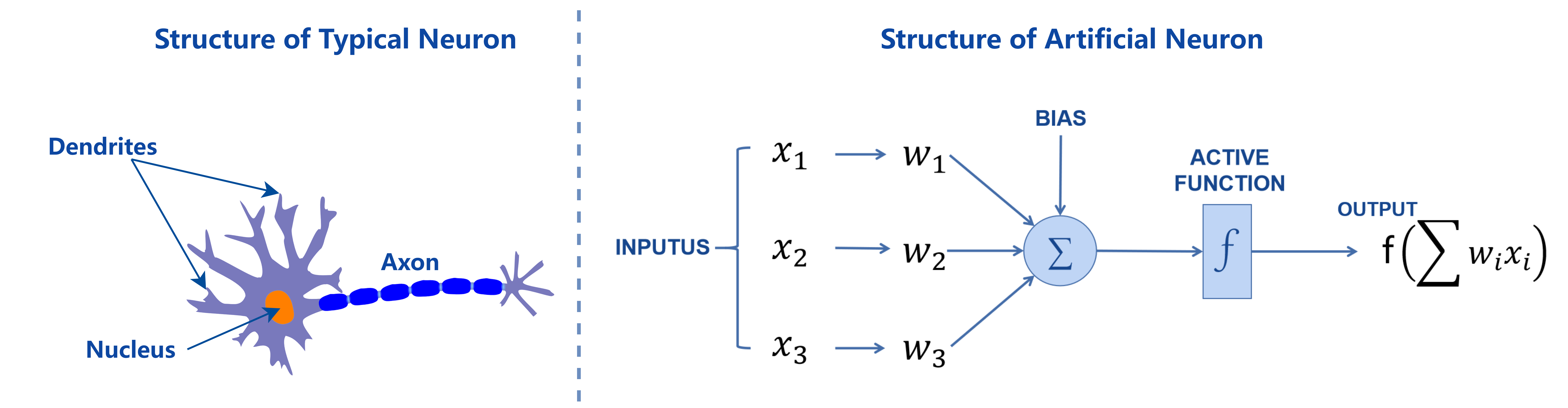}
	\caption{Structure of a typical neuron and structure of an artificial neuron.}
	\label{fig:Artificial_Neural_Network}
\end{figure}

After decades of development, nearly 40 artificial neural network models have been proposed, including back propagation networks, perceptron, self-organizing maps, Hopfield networks, Boltzmann machines, etc. In recent years, many classical models, such as CNNs~\cite{kattenborn2021review}, RNNs~\cite{mikolov2010recurrent}, and LSTMs~\cite{sherstinsky2020fundamentals}, have been widely used in various classification and prediction tasks in the fields of image, voice, text, graph, and so on. The training of artificial neural network models depends heavily on the amount of data. With the explosion of data volume and the deepening of model complexity, people began to separate the training and application of the model. Models are pre-trained based on large offline datasets, saved, and then applied to the problems using the transfer learning technique for quick solutions. BERT, proposed by Google AI Research Institute, and GPT-3, developed by OpenAI, are the two most well-known pre-trained models~\cite{Radford2018Improving,devlin2018bert}. They have achieved great success in natural language processing~\cite{borges2021strategic}.

Artificial neural networks are key building blocks for deep learning (DL) systems, including deep reinforcement learning (DRL) systems. DRL systems use multi-layer neural networks to solve Markov decision problems (MDPs)~\cite{Feriani2021Single}. Both single-agent and multi-agent DRL models are increasingly being used to solve a variety of computing problems (e.g., decision/control and prediction problems) intelligently, which would otherwise be infeasible to solve in a real-time manner.

\subsubsection{Fuzzy Systems}
Lotfi Zadeh initially introduced the notion of fuzzy logic in 1965~\cite{zadeh1965fuzzy}. The fuzzy system is a technique of computing based on fuzzy logic of ``degrees of truth" rather than the typical ``true or false" (1 or 0) Boolean logic on which the contemporary computer is built. The absolute values of 0 and 1 do not provide a good analogy for natural language, nor do they adequately describe most other activities in life or the cosmos. %While it's a fascinating philosophical debate whether the physical world can be described in binary terms, most of the information we may wish to feed a computer is in some intermediate state, and so are the outputs of computing quite often. 
Fuzzy logic might be regarded as the way thinking really operates, with binary or Boolean logic being a subset.

The word ``system" refers to a group of interdependent parts interacting and having a clear structure~\cite{mahfouf2001survey}. Systems can be identified as a complex whole from the external environment. Inputs and outputs are the channels through which a system interacts with its surroundings. Fuzzy systems are information processing architectures built using fuzzy approaches when it is either impractical or difficult to use conventional set theory and binary logic~\cite{czabanski2017introduction}. Their primary feature is the representation of symbolic information as fuzzy conditional (if-then) rules.

The four functional building blocks, i.e., the fuzzifier, fuzzy inference engine, knowledge base, and defuzzifier, make up the conventional structure of a fuzzy system~\cite{czabanski2017introduction}. A fuzzy system may take both crisp data and linguistic values as inputs. If you are working with crisp data, you should focus on the fuzzification stage rather than the inference phase, when the corresponding fuzzy set is assigned to the non-fuzzy input. The appropriate approximation reasoning approach is used to translate the input values into the language values of the output variable. Fuzzy conditional rules are used to reflect the expert's knowledge. When the fuzzy system requires numerical output information, the defuzzification methods are utilized to match the appropriate data set to the resulting fuzzy set.

Fuzzy systems have practical applications when there is a lack of comprehensive mathematical description or when it is very costly or difficult to use a precise (non-fuzzy) model. A fuzzy system is a great tool to process incomplete data, for example, for signal and image processing~\cite{czogala2000fuzzy,rutkowski2004new}, system identification~\cite{hellendoorn2012fuzzy,marszalek2014modeling}, decision support~\cite{stachowiak2016bipolar,yu2003applied}, and control processes~\cite{kacprzyk1983multistage,mamdani1975experiment}.

\subsubsection{Evolutionary Computation}
Evolutionary computation is a unique type of computing that takes its cues from the course of natural evolution. It is not unexpected that some computer scientists have looked to natural evolution for inspiration since the numerous organisms that make up our planet have all been specifically designed to thrive in their niches, which demonstrates the force of evolution in nature. A key point in evolutionary computing is to compare this potent natural development with a specific problem-solving approach called trial-and-error (also known as generate-and-test). The value of potential solutions, or how effectively they address the issue, influences the likelihood that they will be retained and utilized as building blocks for developing other potential solutions. Later, descriptions of pertinent sections of genetics and evolutionary theory are provided.

In the 1940s, long before the invention of computers, there were already ideas of using Darwinian principles to automate problem-solving~\cite{Fogel1998Evolutionary}. It was in 1948 that Turing coined the phrase ``genetic or evolutionary search", and by 1962, Bremermann had conducted computer tests on ``optimization by evolution and recombination". Throughout the '60s, three different implementations of the core idea emerged. Holland named his approach genetic algorithms~\cite{de1975analysis}, while Fogel, Owens, and Walsh presented evolutionary programming in the USA~\cite{fogel1965artificial,fogel1998artificial}. Rechenberg and Schwefel in Germany created evolution strategies for optimization at the same time~\cite{schwefel1993evolution}. These fields evolved independently for roughly 15 years. Since the early 1990s, they have been considered as different technology dialects later known as evolutionary computing~\cite{back1996evolutionary}. Early in the 1990s, a fourth stream, i.e., genetic programming, was promoted by Koza~\cite{banzhaf1998genetic} following the main notions. According to the current terminology, evolutionary computing (or evolutionary computation) refers to the entire field, the algorithms involved are known as evolutionary algorithms, and evolutionary programming, evolution strategies, genetic algorithms, and genetic programming are considered as subfields falling under the umbrella of the corresponding algorithm variants.

\subsection{Perceptual Intelligence}
An intelligent system first starts intelligent perception before it starts to work. Thus perceptual intelligence plays a vital role in all intelligent systems. Perceptual intelligence focuses on multi-modal perception, data fusion, smart signal extraction, and processing. Typical examples include smart city management, automatic diving system, smart defense system, and autonomous robots. A most consistent enthusiasm for perceptual intelligence is human-like five-sense capabilities, including vision, hearing, smelling, tasting, and tactile. In addition, intelligent sensing covers temperature, pressure, humidity, height, speed, gravity, etc., whenever a significant effort in computing or data training is required to advance its performance.

\begin{figure}[H]
	\centering
	\includegraphics[width=0.95\textwidth]{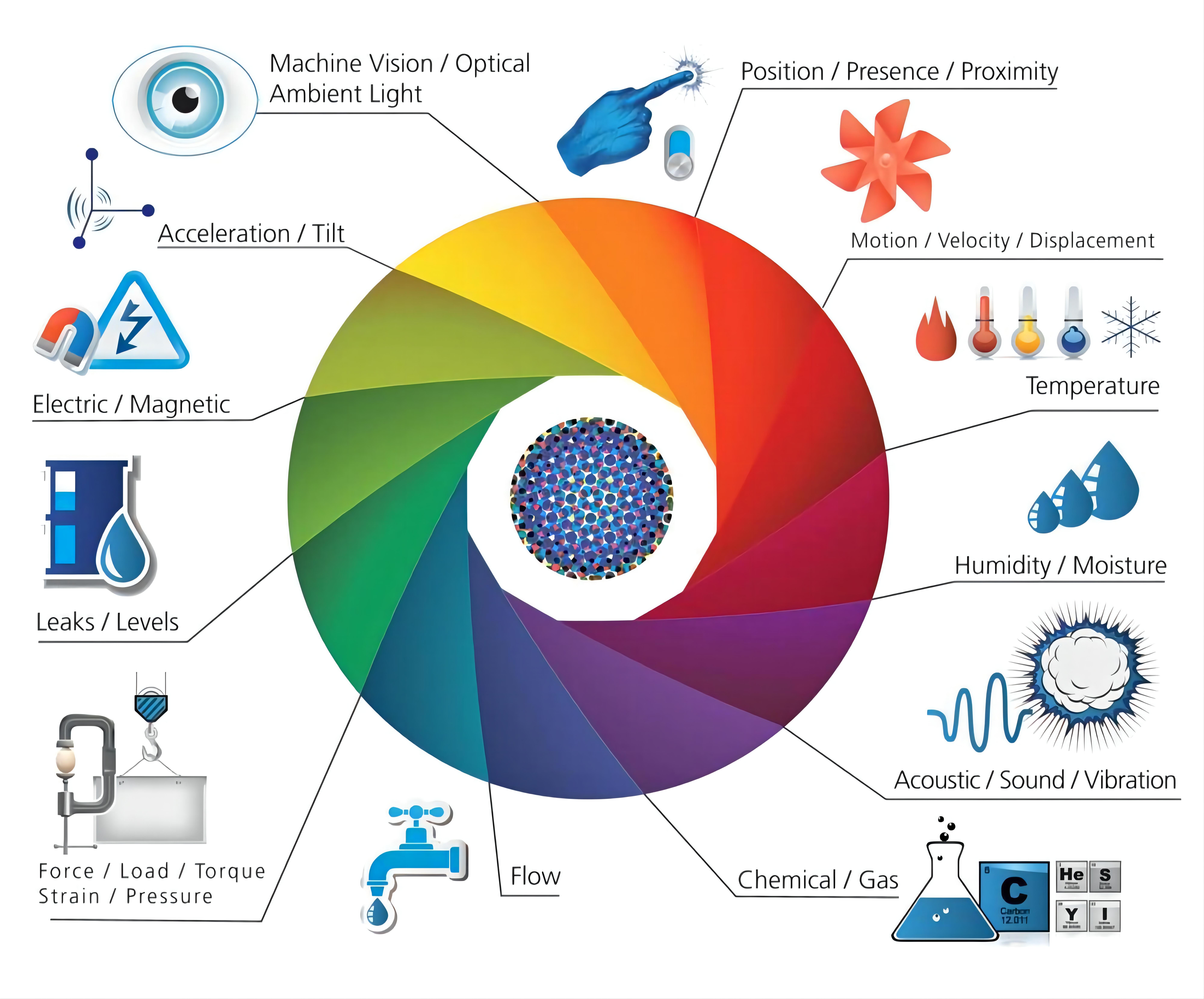}
	\caption{A wide variety of sensor types used in the industry that need the connection to the IoT~\cite{postscapes}.}
	\label{fig:Postscapes}
\end{figure}

Significant advances have emerged in machine vision during the last several decades, emphasizing the creation of devices that can see and understand their surroundings independently. The observation of industrial processes is no longer a challenge in manufacturing contexts due to the constrained range of states and clearly defined circumstances~\cite{velik2007model}. However, the situation becomes a significant challenge from industrial processes to free surroundings perception in the real world. Because there are an unlimited number of situations and unexpected events that may occur at any time, it is still a challenge to handle these scenarios by a fully autonomous robot. On the contrary, even a toddler can effortlessly observe the world. Our brains possess the most effective circuits and processing systems, which allow us to process sensory data from millions of sensory receptors. Machine intelligence would undoubtedly undergo a revolution if these circuits and processes could be understood and implemented technically. Applications include safety and security monitoring in public and private structures as well as in the observation of the behavior and health of elderly or physically or psychologically impaired individuals in nursing homes and hospitals~\cite{bruckner2007probabilistic}. Additionally, it may enable seniors to remain in their homes for longer~\cite{burgstaller2007interpretation}. A model like this would also be highly helpful for autonomous robots that must traverse their surroundings and control things in them, as well as interactive environments that make users more comfortable by sensing their demands~\cite{pratl2006processing}.

%\begin{table}[htbp]
	%\centering
	%\caption{Comparison of visual sensors}
	%\resizebox{\textwidth}{!}{%
%	\begin{tabular}{|p{2cm}|p{3cm}|p{3cm}|p{3cm}|p{3cm}|}
		%\hline
		%& \textbf{Monocular camera} & \textbf{Structured light} & \textbf{Time of flight} & \textbf{binocular visual} \\ \hline
		%Active/Passive  & Passive  & Active & Active  & Passive\\ \hline
		%Working principle  & Use a convex lens to focus the scene on the camera focal plane (map 3D to 2D).  &  Encode the optical pattern of the object and measure the pattern change.  &  Directly measure the time delay or phase delay of reflected light.  &  RGB image feature point matching and indirect triangulation calculation. \\ \hline
		%Deep sensing accuracy  &  /  &  0.01-1 mm  &  1-10 cm  &  $<$1 mm  \\ \hline
	%	Deep sensing range  & /  &  $<$5 m  &  $<$10 m  &  $<$20 m  \\ \hline
		%Frames rate (fps)  & 15-400  & 30-60  & $>$100  & 15-50\\ \hline
		%Software complexity  & low  & high  &  low  & high \\ \hline
		%Environmental parameter immunity  & Affected by ambient light.  & Affected by ambient light, seriously affected by the reflection of objects, & Affected by ambient light, weakly affected by the reflection of an object. & Affected by ambient light, affected by the texture of the object.\\ \hline
		%Applications  & Indoor and outdoor, not applicable at night.  & Indoor & Indoor  & Outdoor, not applicable at night\\ \hline
	%\end{tabular}}
	%\label{tab:Visual_Sensors}
%\end{table}

With the complete use of pattern recognition and deep learning technologies, machines have grown more perceptually intelligent than humans in recent years, making significant advances in voice, visual, and touch recognition. Because of their increasing importance and many possible applications, smart sensors have received a lot of attention. Ordinary sensors have become smart because of the integration of computers and IoT in manufacturing, allowing them to do intricate computations with the data gathered~\cite{bibby2018defining}. Smart sensors have expanded in capability, size, and flexibility, transforming cumbersome machinery into high-tech intelligence. Smart sensors have evolved into objects with detection and self-awareness capabilities because they are fitted with signal conditioning, embedded algorithms, and digital interfaces~\cite{choy2020ubiquitous}. These sensors are designed as IoT components that transform live data into digital data that can be sent to a gateway~\cite{herrojo2019chipless}. Process control and quality evaluation are only two of the many functions provided by these devices. Smart sensor data may also be used to minimize manufacturing costs via process optimization and predictive maintenance thanks to cloud-based analysis tools and AI. From supply management to global resource coordination, sensor data may be used in several ways once transmitted online. Smart sensors come in various shapes and size to meet the needs of diverse applications, as shown in Figure~\ref{fig:Postscapes}, and new and better models are always being developed. One of the most prevalent sensor types is a light sensor that measures the light intensity and color temperature. From large portfolio firms like TE Connectivity to more specialist vendors like Aceinna, there is a sensor type for almost any sort of process or environmental situation.

Smart sensors can also foresee, monitor, and immediately respond to remedy situations. The primary tasks of intelligent sensors include raw data collection, sensitivity, filtering adjustments, motion detection, analysis, and communication~\cite{mulloni2020chipless}. For instance, one use of smart sensors is wireless sensor networks, whose nodes are coupled with one or more additional sensors and sensor hubs to create some communication technology. Additionally, data from several sensors may be used to draw inferences about a problem already present; for example, temperature and pressure sensor data might be used to predict the beginning of a mechanical breakdown.

\subsection{Cognitive Intelligence}
Cognitive intelligence refers to machines having the capacity to logically understand and cognize like humans, especially the ability to think, comprehend, summarize, and actively apply knowledge. It describes the abilities and skills to process complex facts and situations in the real environment, like interpretation and planning. Data recognition is the core function of perceptual intelligence, which requires large-scale data collecting and feature extraction of images, videos, sounds, and other types of data to complete structured processing. In contrast, cognitive intelligence requires understanding the relationship between data elements, analyzing the logic within the structured data, and responding based on the distilled knowledge. Cognitive intelligent computing primarily studies the topics of natural language processing, causal inference, and knowledge reasoning of machines. Through heuristic research on the neurobiological process and cognitive mechanism of the human brain, a machine can improve its cognition level to assist, understand, make decisions, gain insight, and discover.

\subsubsection{Natural Language Processing}
Natural language processing converts human language into machine language so that machines can understand and calculate. This research field has a very lengthy task link, from upstream information extraction, data cleaning, data retrieval, and pre-training~\cite{Radford2018Improving,devlin2018bert}, to downstream text classification, question-answering system, and automatic abstract, etc. Natural language processing concentrate on two major tasks: natural language understanding (NLU) and natural language generation (NLG). NLU understands the meaning of a text to the extent that the word and structure must be grasped. The specific steps include lexical analysis, syntactic analysis, and semantic analysis.

Lexical analysis plays a key role in the word segmentation module of Chinese natural language processing~\cite{wong2009introduction}. The critical components of lexical analysis contain word segmentation, part-of-speech tagging, named entity recognition, and word sense disambiguation. Part-of-speech and semantic tagging are the primary functions of lexical analysis. Word disambiguation mainly addresses the issue of word meaning across various contexts because a word may have numerous meanings depending on contexts, and it is necessary to choose the most appropriate word meaning for the task context at hand. The primary goal of named entity recognition is to locate and annotate words with specific meanings in the context, such as names of people or places. The foundations of lexical analysis are constituted by rules, statistics, and machine learning~\cite{church2021using}.

Determining the relationships between each component of a sentence (or its syntactic structure) is the primary goal of syntactic analysis. Chomsky's hierarchy of grammar is currently the prevalent context-free syntax model. It obtains the syntactic tree of a sentence through a complete set of analyses.

Research on NLU mostly focuses on semantic analysis. It covers every stage of natural language understanding. Semantic analysis refers to three primary tasks in several levels of granularity: sense disambiguating at the word level, semantic role labeling at the sentence level, and coreference digesting at the discourse level. As auxiliary means of semantic analysis, pragmatic analysis and affective analysis have also been extensively studied. The pragmatic analysis mainly studies the relationship between text and environment, including speaker, receiver, and context. Sentiment analysis can obtain user preferences, emotions, and the potential tendency of speech. Earlier studies contributed by establishing an emotional dictionary~\cite{tran2018hybrid, bravo2022incremental}. In recent years, methods based on machine learning and deep learning have begun to analyze the emotional features of texts by constructing learning models~\cite{ahmad2018svm, wei2020bilstm}.

NLG generates new text information from raw text, data, and images. The main applications are machine translation~\cite{hirschberg2015Advances}, question and answer system~\cite{khansa2015understanding}, automatic abstract~\cite{gambhir2017recent}, and cross-modal text generation~\cite{he2017deep}. The classic approach of NLG comes in three stages. First, identifying what goals should be established and deciding what should be included in the text. Secondly, planning for how to achieve the goal by evaluating the scenario and available communicative resources, such as text structuring, sentence aggregation, lexicalization, and referring expression generation. Finally, generating text following the plan. In recent years, end-to-end approaches have attracted more attention with the advances in neural networks and increased complexity and specificity of tasks~\cite{yin2017comparative}. End-to-end methods construct models from input to output directly, iteratively enhance models based on the feedback of training task data and form a procedure of a full closed-loop calculation~\cite{kalchbrenner2013recurrent}.

\subsubsection{Causal Inference}
Current machine learning relies heavily on associative models leading to the poor interpretability of AI. It is difficult for machines to distinguish true or false causal associations in data. The key to solving this problem is to use causal inference instead of inference by association so that machines can use appropriate causal structures to model the inference world. Pearl uses three hierarchical structures to categorize causal inference~\cite{Pearl2009Causality}. The first layer is the association, which involves data-defined statistical correlations. The second level is intervention, which involves what is visible and what will result from additional intervention or action. The third layer is counterfactual, which is the reflection and retroactivity of past events. It answers the question, ``what would have been different if we had acted differently in the past"? The counterfactual layer is the most powerful level. If a model can answer counterfactual questions, it can also answer questions about intervention and observation.

Hume offered a literal exposition and initially suggested discussing causality using a counterfactual framework~\cite{Hume1955An}. Lewis gave a symbolic expression of the counterfactual framework based on Hume’s research by combining the semantics of possible worlds with counter facts to characterize causal dependence~\cite{Lewis1973Causation}. Verma \textit{et al.} learned from actual data to predict counterfactual results~\cite{Verma2020Counterfactual}. Besserve \textit{et al.} proposed a non-statistical framework. They revealed the modularization structure of the network by counterfactual reasoning, which consists of the entangled internal variables~\cite{Besserve2018Counterfactuals}. Kaushik \textit{et al.} designed a human-in-loop system for the counterfactual operation of documents. They suggested eliminating misleading associations using feedback in the loop~\cite{Kaushik2019Learning}.

The potential outcomes framework is one of the most important theoretical models in causal inference. The model was proposed by Rubin, a well-known statistician from Harvard University~\cite{Imbens2015Causal}, and is also called the Rubin causal model. The core of the potential outcomes model is to compare the effects on the same subject with or without intervention. Whether a result appears or not for a target mainly depends on the assignment mechanism. The fact that we can only see one outcome does not mean the other does not exist. Therefore, it is more reasonable to describe events regarding potential outcomes. Except for the potential outcome models, the structural causal model is one of the most widely used models in causal inference. The structural causal model can describe the causality of multiple variables. Pearl developed a formal expression method of causality based on external intervention and created a way to explore the causality and data generation mechanism from data~\cite{Pearl2009Causality}. Causal Network mines causal patterns from a large text corpus by gathering causal terms to determine causality~\cite{Luo2016Commonsense}. Data-driven approaches, such as concept network, which manually collects information to encode causal events as common sense, derive causality from text~\cite{Havasi2010Open,Gordon2011Commonsense}. Causal reasoning and natural language processing can be combined to extract causal relationships between terms or phrases from large textual corpora, capturing and comprehending the causal relationships between events and actions. Luo \textit{et al.} used a data-driven approach to solve the problem of commonsense causal reasoning between short texts. They proposed a framework to automatically collect causal relationships from an extensive network corpus, which can correctly model the strength of causal relationships between items~\cite{Luo2016Commonsense}. Dasgupta \textit{et al.} trained a recursive network with model-free reinforcement learning to overcome cause and effect problems~\cite{Dasgupta2019Causal}. Recent advances in causal representation learning retrieve the real-world model without prior knowledge of manual partitioning~\cite{Rehder2003A}.

\subsubsection{Knowledge Reasoning}
Knowledge reasoning has always been a crucial component of cognitive intelligence. Traditional reasoning, which includes deductive and inductive reasoning, is derived from classical mathematical theory. Deductive reasoning starts from general premises and leads to specific statements or individual conclusions~\cite{Robinson1965Automatic}. Inductive reasoning is from the individual to the general. It derives generality principles and rules from concrete examples~\cite{Solomonoff1964A}.

\begin{figure}[htbp]
	\centering
	\includegraphics[width=1\textwidth]{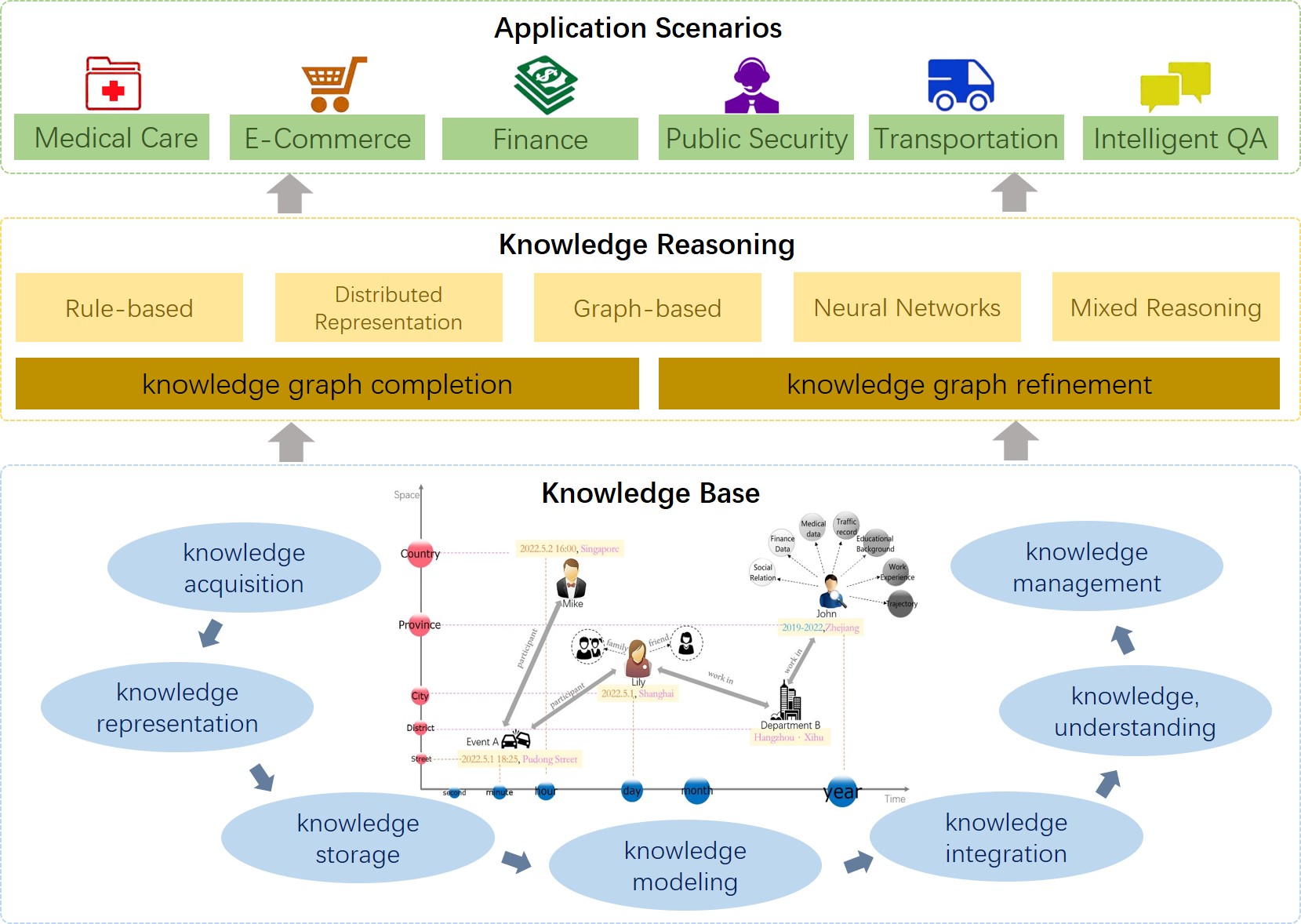}
	\caption{Overview of knowledge reasoning.}
	\label{fig:Knowledge_Reasoning}
\end{figure}

Knowledge reasoning builds the knowledge base using the graph data model or topology to integrate data, as shown in Figure~\ref{fig:Knowledge_Reasoning}. It stores the semantic entities with free form (objects, events, situations, or abstract concepts) and their relation description. The research contains seven aspects: knowledge acquisition, representation, storage, modeling, integration, understanding, and management.

As an effective way to express knowledge, a knowledge graph connects entities through rich semantic relations and constructs a systematic and semi-structured knowledge base. Knowledge graphs have been widely used in vertical application fields such as medical care, e-commerce, finance, public security, transportation, and intelligent question-answering. Typical knowledge graphs include YAGO~\cite{Suchanek2007YAGO}, DBpedia~\cite{AUER2007DBpedia}, Freebase~\cite{Bollacker2008Freebase}, Wikidata~\cite{Vrandecic2014Wikidata}, etc. These knowledge graphs extract, organize and manage knowledge from many data resources and then store and represent knowledge as triples. They help understand the semantics of search and provide accurate search answers.

Knowledge reasoning based on knowledge graphs mainly focuses on relations, that is, inferring unknown facts or relations based on existing ones in the graph or identifying and correcting errors in existing entities, relations, and graph structures based on prior knowledge and experience. It includes reasoning based on rules, distributed representation, graphs, and neural networks.

Reasoning based on logical rules mainly uses first-order predicate logic, description logic, and probability logic to deduce new entity relations. Typical methods include ProPPR~\cite{Wang2014Efficient}, TensorLog~\cite{Cohen2016TensorLog}, SRL~\cite{Getoor2007Introduction}, among others.

Graph-structure-based reasoning represented by path ranking algorithm~\cite{Lao2010Relational} combines semi-structured topological features of knowledge graphs with statistical criteria. This kind of method considers the path relation between entities, introduces statistical rules into the algorithm, and produces a strong inference effect.

Neural-network-based reasoning uses deep learning models to infer knowledge. Neural tensor network (NTN)~\cite{Socher2013Reasoning} constructs word vector average representation of entities. R-GCN~\cite{Schlichtkrull2017Modeling} captures the information of adjacent entities by convolution networks. IRN uses RNN as the control unit to simulate the process of multi-step reasoning and introduces an attention mechanism. DeepPath~\cite{Xiong2017DeepPath} initially introduces the reinforcement learning framework into the knowledge reasoning model.

Reasoning based on distributed representation learning learns fact tuples in the knowledge graph by representation model and obtains low dimensional vector representation of the knowledge graph. The inferential prediction is then converted into a simple vector operation based on the representation model. Its core is to map the knowledge graph to continuous vector space and deduce implicitly through calculating the distributed representation of each element. Most representation learning approaches, including TransE~\cite{Bordes2013Translating} and RESCAL~\cite{Nickel2011A}, build various learning models based on different spatial assumptions and use a single-step relationship, or a single triad, as their input and learning objective.

\subsection{Autonomous Intelligence}
Two critical ingredients drive machines from passive output to active creation: a strongly generalized model and continuous interaction with the external environment. 
The development path of autonomous intelligence starts from learning a single task to learning by drawing inferences from one example, gradually reaches active learning by dynamically interacting with the environment, and finally ends at the advanced intelligent goal of self-evolution.
This subsection focuses on developing technical fields such as transfer learning, meta-learning, and autonomous learning to look at feasible paths for generating autonomous intelligence.

\subsubsection{Transfer Learning}
The basic idea of transfer learning is to use the strategy of the solved problem to solve the new problem, that is, transfer the existing experience to the past. Currently, most neural network methods are used to train the model as a branch of machine learning. The parameters of the trained model are usually used as a set of initial values to reduce the complexity of model training. Transfer learning focuses on training the base model in the sample space by optimizing a single overall task as a transfer source. The appropriate model is directly transferred to the target domain, and then the target model is fine-tuned using a small number of labeled samples. The original intention of transfer learning is to save the time of manually labeling so that the model can transfer from the existing labeled data (source domain data) to the unlabeled data (target domain data). It can make the most use of the obtained data and reduce the sample size requirements for machine learning. 
 
In transfer learning, data are divided into the source and target data. The source data refers to other data not directly related to the unsolved task, usually with a large data set. The target data is directly related to the task, which is of a small amount. Transfer learning aims to establish a mapping relationship from the source domain to the target domain with some additional data or existing models. It applies the general knowledge to new tasks to fully use the source data to help the model improve on the target data. Transfer learning can also combine with other models, such as federated learning and reinforcement learning~\cite{chen2020fedhealth, da2019survey}. 

Transfer learning can be classified into four categories according to the learning style. Instance-based transfer learning selects instances from the source domain to help train the target domain~\cite{yao2010boosting}. Different weights are assigned to the instances. The more similar the instances are, the higher the weights are. Instances with higher weights have higher priority. Feature-based transfer learning maps the target and source domains into the same space, minimizing the distance between the distributions of the two domains~\cite{long2015learning}. 
The symmetric-space methods transform the source and the target domain feature space into a common subspace.
The asymmetric-space methods directly convert the source domain feature space to the target domain feature space (or, on the contrary) to achieve the alignment of the two domains.
This method can solve the problem of inconsistent data distribution between the source and target domain to solve the data-lacking problem completely. Model-based transfer learning reuses the model trained on the source domain and adjusts the model parameters via fine-tuning or fixed feature extractor~\cite{Yosinski2014HowTA}. Relationship-based transfer learning explores the relationship of similar scenes and uses the correlation implicit in the relationship between the source and the target domain~\cite{Davis2009Deep}.

\subsubsection{Meta-Learning}
Meta-learning aims to help machine learning to learn~\cite{Nichol2018On, Cruz2015META-DES} so that the machine can quickly learn various complex new tasks in the real environment. Traditional machine learning methods manually adjust the parameters in advance and directly train the deep model under specific tasks. While meta-learning will make the machine learn all the parameter variables that need to be set and defined by humans in advance, including how to pre-process data, choose network structure, set hyperparameters, define a loss function, and so on~\cite{GrantFLDG2018Recasting}. The experience gained from the learning history gives the machine meta-knowledge. As a result, it can quickly handle new tasks with only a few data samples. Meta-learning is mainly used in few-shot learning, zero-shot learning, unsupervised learning, and other fields with very little available data. Meta-learning is proposed to solve the traditional neural network models' problems of insufficient generalization performance in few-sample cases and poor adaptability to new tasks. The idea of meta-learning makes the machine-learning process more autonomous by reducing the model-design cost for various similar tasks.

Since the goal of meta-learning is to obtain the ability to learn new tasks by the meta-knowledge from training data quickly, meta-learning considers the entire task set as training examples. 
Meta-learning gets the initial network parameters with strong generalization on the training and the validation data set. It performs a few gradient descent operations on the test data to learn new tasks. Then, it tests the effect of the model after learning. Meta-learning obtains a good initial value of the model through preliminary training and then updates the weight of the specific task with a small amount of training data based on the initial value to achieve good results. Meta-learning can also be regarded as finding a set of high-sensitivity parameters. Based on the parameters, only a few iterations are needed to achieve desirable results on a new task.

The most influential meta-learning model to date is Model-Agnostic Meta-Learning (MAML)~\cite{Chelsea2017MAML}. MAML is not a deep learning model but more like a training technique. It targets training a set of fine-tuned parameters for a group of tasks rather than a model for a specific task. Thus, the inputs to feed in MAML are tasks, not data. MAML uses a set of adaptive weights, which can be adapted well to new tasks after a few gradient descents. Then, finding this weight is the training objective. MAML iteratively trains a batch of tasks. In each iteration, it first trains each task in the batch, then returns to the original status, makes a comprehensive judgment on the loss of these tasks, and then selects a direction suitable for all tasks in the batch.

\subsubsection{Autonomous Learning}
Meta-learning can handle the general solution model of a specific type of task by learning from similar task sets and can transfer learning between tasks. However, this learning ability can only transfer between homogeneous tasks, where even the support and query set sizes of tasks are strictly aligned. Autonomous learning aims to transform from passive data acceptance and training to active learning and improve learning efficiency, which is the direction considered by Turing Award winner Yann LeCun~\cite{lecun2022path}. In addition to higher-level transfer learning capabilities, models of the external open world are incorporated into the design of autonomous intelligent architectures. 
 
Humans and other animals have always been able to learn a great deal of background knowledge about how everything works in an unsupervised manner through observation and a small amount of interaction. This knowledge is what we call common sense, which is the basis of the model of the world. LeCun designed a learning framework that allows machines to learn a ``model of the world" in a self-supervised manner (i.e., without labeled data). He used that model to make predictions, reason, and act. In this model, he extracts valuable ideas from various disciplines and combines those ideas to propose an autonomous intelligence framework consisting of six modules (configuration module, perception module, world model module, cost module, action module, and short-term memory module). Each module can easily calculate objective functions, estimate the corresponding gradient, and propagate the gradient information to the upstream modules.
 
Most of the modules in this cognitive framework of autonomous intelligence have an analogy to the animal brain. The perception module corresponds to the cortex processing visual, auditory, and other sensory pathways. The world model corresponds to some partial high-level processing units of the prefrontal cortex. The intrinsic cost module corresponds to the amygdala. The trainable critic cost module, however, corresponds to the part of the prefrontal cortex responsible for reward prediction. The short-term memory module can correspond to the hippocampus. At the same time, the configurator corresponds to the central control and attention regulation mechanism in the prefrontal cortex. The actor module corresponds to the premotor cortex. Through this highly brain-like design, not only does the transfer of learning ability across tasks look promising, but it also introduces common sense and emotion into the framework in a modular way, allowing the machine to take a big step towards the ``conscious" reasoning and planning.

\subsection{Man-machine Integrated Intelligence}
Despite significant progress in the four levels of intelligence, more is needed to get critical insights from extremely complicated scenarios only by calculation/statistical models. In these scenarios, humans should continue to play an indispensable role in problem-solving and decision-making, explore the elements involved in human cognitive processing, and combine them with machine intelligence. The following will focus on human-computer interaction, human-machine integration, and brain-computer interface.

\subsubsection{Human–Computer Interaction}
Computers have appeared in various forms in daily life and industrial operations. Various methods and products have been designed to improve the usability of computers. The development of human-computer interaction technology further releases the potential of computers and improves users' work efficiency. Human-computer interaction has gone through the early stages of humans adapting to computers via manual work, command language, graphical user interface (GUI), network user interface, and so on. GUI is simple and easy to learn by reducing typing operations. Ordinary users who need help understanding the computer can also skillfully use it. It realizes the actual standardization and brings unprecedented development to the information industry due to the expansion of the user population. 

With the universal development of the network and the development of wireless communication technology, the human-computer interaction field is facing enormous challenges and opportunities. The users require a more convenient interaction pattern in the multimedia terminal. At the same time, the operation interface has innovations in aesthetics and forms. It has reached the multi-modal and imprecise interaction stage and is constantly developing in the direction of human-centered natural interaction. In this stage, human-computer interaction uses multiple communication channels. Modality covers a variety of communication methods for users to express intentions, perform actions, or perceive feedback information. Computer user interfaces that take this approach are called multi-modal user interfaces (MMI). MMI uses a variety of human sensory channels and action channels (such as speech, handwriting, posture, sight, expression, touch, smell, taste, and other inputs) to interact with the computer environment in a parallel and imprecise manner. It frees people from the shackles of traditional interaction methods and enables people to enter a period of natural and harmonious human-computer interaction~\cite{Chen2021Interactive,Cui2019User}.

\subsubsection{Human-Machine Integration}
The theory of human-machine fusion intelligence focuses on a new form of intelligence produced through interaction between humans, machines, and environments. It is a brand-new generation of intelligent scientific systems that combine physical and biological characteristics. Human-machine fusion intelligence, which effectively mixes objective data collected by hardware sensors and subjective information perceived by human senses, integrates the profound cognitive way of people and the superior computing power of computers~\cite{wang2019human,krishna2017visual}. It utilizes human prior domain knowledge as important learning clues to construct new understanding approaches and enhances computer-based decision-making. With the new methodology, computers can manage complex problems in professional fields that neither humans nor machines are capable to handle alone.

Human-machine integration, with humans and machines directing the integrated learning process until consensus, is conducted interactively and collaboratively rather than statically. It requires intuitive communication between the two components via an interactive platform. Machine intelligence can be interpreted and sent directly to humans. Experts can easily submit feedback in a natural form. Furthermore, the fusion should automatically adapt to a dynamic environment so that integrated intelligence can continuously evolve with updates in human knowledge. Thus, self-evolving integrated intelligence is critical for handling dynamic scenarios so that the tasks and data can change rapidly~\cite{sun2021brain}.

To effectively bring the computer into the real-time thinking process, humans and machines must be more closely coupled. It is not easy to achieve high real-time performance in integrated computers traditionally. An advanced mode to combine human and machine abilities is human-machine symbiosis. A human-machine symbiosis system should better understand the human intention in terms of better interaction and cooperation~\cite{gill2012human}. The hardware, such as sensors, bracelets, wearable devices, and other computers, is formalized in invisibility like air. For example, wearable devices can be attached to clothes and shoes to realize human-machine symbiosis~\cite{gill2012human}. In software, the development of meta-universe technology will provide people with a fully immersive experience of human-machine symbiosis~\cite{gerber2020conceptualization,sun2020potential}. Computer technology will continue to serve us in the future, and interactive interfaces and tasks will become more natural and intelligent. The cloud-side-end distributed interaction and collaboration system will be built with virtual and real integration. It will obtain human functions in the loop of information perception, modeling, simulation, deduction, prediction, decision-making, presentation, interaction, and control. It will provide the continuous learning ability of human-machine collaboration. It will provide platform support capabilities for significant applications such as remote exploration and operation in uncharted environments, collaborative command and operation of complex systems, human-machine co-driving environment, and research and governance of social problems integrating virtual and reality.

\subsubsection{Brain-Computer Interface}
Brain-computer interface (BCI) is an interactive system established by analyzing the electroencephalogram (EEG) signals of humans (or animals). It breaks through the limitation of traditional neural reflex arc structure and enables the brain nerve signal to communicate directly with the computer by wire or wireless to control and communicate directly with external electronic application equipment. According to the signal acquisition method, BCI is divided into three categories: non-invasive, semi-invasive, and invasive. Non-invasive BCIs utilize signal sources, including surface EEG, magnetoencephalography (MEG), functional magnetic resonance imaging (fMRI), functional near-infrared spectroscopy (fNIRS), etc. Semi-invasive BCIs use electrocorticography (ECoG). Invasive BCIs utilize intracortical EEG. Due to the simple acquisition of equipment, convenient operation, safety, and easy clinical use, EEG technology is greatly valued. The apparent advantage of EEG signals is that they can achieve high temporal resolution at the millisecond level, which is suitable for real-time monitoring and online transmission~\cite{Ramadan2017Brain}.

The single-modal brain-computer interface faces some challenges, including poor robustness of long-term operation, classification accuracy affected by the number of commands, human-machine adaptability, and system stability to be improved. For example, the number of tasks that a single-modality BCI system can achieve is limited, which restricts the completion of complex tasks by external output devices. With the increase in the number of function instructions, the classification accuracy decreases, the system stability is limited, and it is difficult to obtain good results in practical applications. Given the above problems of single modality brain-computer interface, the hybrid brain-computer interface (HBCI) concept has been proposed in recent years. HBCI is also known as a multi-modal brain-computer interface (MBCI). It refers to a system combining a unimodal brain-computer interface with another system (BCI system or non-BCI system)~\cite{Erwei2013Anovel,Qiang2017Noninvasive}. 
HBCI can satisfy the demand for multi-instruction and real-time in the multi-degree-of-freedom control system to break through the problem of limited instruction and low accuracy of multiple classification and recognition in single-mode BCIs.
It extends motion commands quantity, increases the applicability and output characteristics of human-computer interaction, and perfects the human-computer interaction system function. It has a broad application prospect in aerial teleoperation and equipment control~\cite{Allison2014A, Jie2014Evaluation}.

The HBCI system has two essential features: information fusion and control strategy. Information fusion includes data level fusion, feature level fusion, and decision level fusion according to the level of information representation. Data level fusion directly fuses the signal data obtained by different sensors. Feature level fusion combines the feature vector extracted from the data obtained by each sensor. Decision level fusion outputs the decision results of the overall system according to voting or weight calculation of classified decisions, which are processed by each sensor separately. According to the control strategies, HBCI system processes input signals by adopting the simultaneous mode~\cite{Yang2020A, Zuo2019Novel} or sequential mode~\cite{Chai2020A, Huang2019An}.

A collaborative BCI (CBCI), which can be applied to group collaboration to improve system performance by increasing the user dimension, is proposed~\cite{Groux2010Disembodied}. The advantages of CBCI are not only to effectively integrate group EEG features, improve decoding accuracy and robustness. It can also improve the decision-making confidence of human-computer hybrid intelligence in cutting-edge tasks. The CBCI usually has precise application scenarios. By designing efficient group brain information fusion algorithms, it can achieve more accurate and faster target control than single brain information~\cite{Kurvers2016Boosting}. The application value of CBCI is to enhance the information processing ability of multi-user brain-computer collaboration systems for specific tasks. It includes strengthening the decision-making ability of the system based on human visual information and the control ability of the system based on human kinesthesia information~\cite{Saugat2019Collaborative, Wang2011A}.

The structure of CBCI is divided into two types: centralized and distributed. A centralized CBCI structure is to perform multi-person EEG joint feature extraction for one or more features~\cite{Kyongsik2016Improved}. The design idea of distributed CBCI structure is group decision-making. The group performs the same task at the same time. Different decision weights are assigned to each user according to their task performance to avoid individual EEG differences. The setting of weights is the critical issue of decision fusion~\cite{Solon2018Collaborative}. Through the design of joint tasks, the exploration of brain activation characteristics, and the analysis of the causal relationship between multi-person cooperation,  the traditional interaction research between a single person and environment/task is gradually transformed into group interaction research between multiple people and environment/task.
It is a sign that BCI technology breaks through the limitations of engineering application. The group-brain collaborative joint operation in CBCI is more in line with the future human-computer interaction socialization and will be unprecedentedly developed and widely used.

%%%%%%%%%%%%%%%%%%%%%%%%%%%%%%%%%%%%%%%%%%%%%%%%%%%%%%%%%%%%%%%%%%%%%%%%%%%%%%%%%%%%%%%%%%%%%%%%%%%%%%%%%%%%%%%%%%%%%%%%%%%%%%%%%%%%%%%%%%%%%%%%%%%%%%%%%%%%%%%%%%%%%%%%%%%%%%%%%%%%%%%%%%%%%%%%%%%%%%%%%%%%%%%%%%%%%%%%%%%%%%%%%%%%%%%%%%%%%%%%%%%%%%%%%%%%%%%
\section{Computing for Intelligence \label{Sec:ComputingForIntelligence}}
AI discoveries are coming out of the woodwork on a regular basis, owing largely to ever-increasing computing power~\cite{powerai}. Compared to the groundbreaking 2012 model that initially popularized deep learning, the biggest model revealed in 2020 required six million times as much computing power. After highlighting this tendency and attempting to quantify its pace of rising in 2018, OpenAI researchers have concluded that this rapid rising cannot be maintained forever. Indeed, the looming slowdown may already be underway.

Historically, the rapid pace of change in AI has been fueled by new ideas or revolutionary theories. Often, the newest state-of-the-art models rely only on bigger neural networks and more powerful processing systems than those previously used in efforts to achieve the same goal. A study to track the growth of the biggest models based on computing power was made by OpenAI researchers in 2018~\cite{amodei2018ai}. Using the amount of computing necessary to train some of the most prominent AI models during the history of AI research, they discovered two trends with the rapid growth of computing resources.

\begin{figure}[H]
	\centering
	\includegraphics[width=0.95\textwidth]{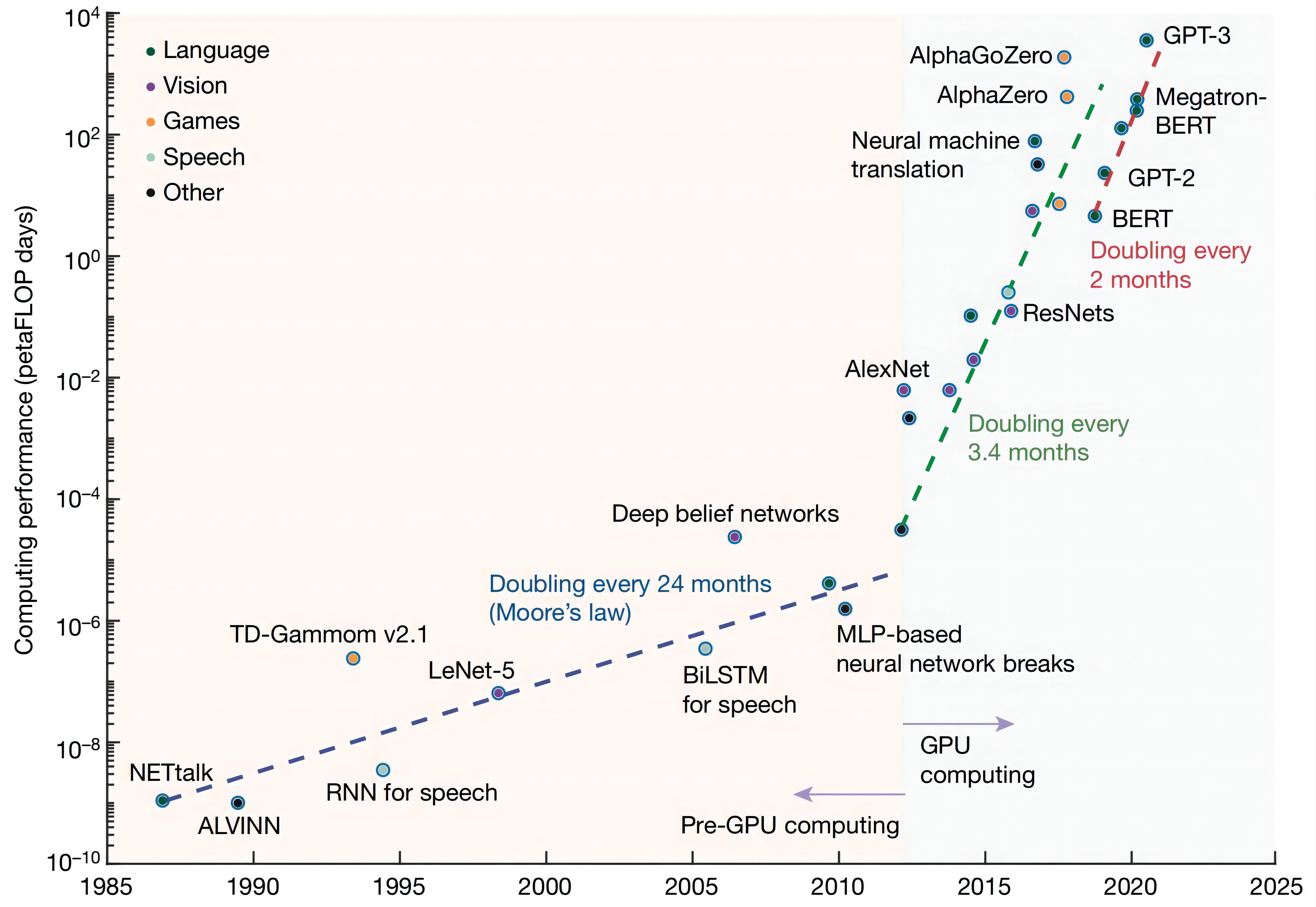}
	\caption{Growth in computing power demands over the past decade substantially outpaces macro trends~\cite{mehonic2022brain}.}
	\label{fig:compting_power_trends}
\end{figure}

Their study shows that the amount of computing power required to develop a breakthrough model has grown at about the same pace as Moore's law, the long-standing observation that the computational capacity of a single microchip has tended to double every two years, before 2012. Though deep learning techniques have been the driving force behind most of the AI developments over the previous decade, AlexNet, an image recognition system, attracted new interest in them in 2012 when it was released. The introduction of AlexNet spurred a dramatic increase in the computational needs of top models, which doubled every 3.4 months between 2012 and 2018, as seen in Figure~\ref{fig:compting_power_trends}.

Research into picture categorization provided the first evidence that growing computational power consistently improved performance in the earliest years of deep learning. However, when image recognition algorithms started to outperform humans at certain tasks by increasing the computing resources~\cite{krizhevsky2012imagenet}, attention switched to other areas. Reinforcement learning techniques were used in huge AI models to play games like Atari or Go in the middle of the 2010s~\cite{mnih2013playing}. Later, a new architecture called the transformer emerged, refocusing emphasis on language tasks~\cite{devlin2018bert}. OpenAI's GPT-3~\cite{brown2020language}, a text generator, has become one of the most popular AI models in recent years. 

Despite advances in algorithms and architectures that allowed for more learning to be accomplished with fewer computations, the processing needs remained high even after these advancements. AlexNet through GPT-3 requires the same 3.4-month doubling period in computing requirements. Computing power is therefore becoming a bottleneck for intelligent computing. At the same time, the energy efficiency of the AI/ML platforms will become increasingly important to reduce the cost of computation~\cite{Hossain2021Editorial}. 

Distributed machine learning (DML) approaches are being developed to make the computations scalable by reducing the computational load on a single server~\cite{Hu2021Distributed}. One category of DML, which is referred to as federated learning (FL), is particularly promising for distributed learning while preserving data privacy at the servers~\cite{Khan2021Federated}. It also avoids the overhead of transmitting a large volume of data from the distributed locations to a central server. After all, different DML paradigms will be vital in future intelligent computing systems.

%%%%%%%%%%%%%%%%%%%%%%%%%%%%%%%%%%%%%%%%%%%%%%%%%%%%%%%%%%%%%%%%%%%%%%%%%%%%%%%%%%%%%%%%%%%%%%%%%%%%%%%%%%%%%%%%%%%%%%%%%%%%%%%%%%%%%%%%%%%%%
\subsection{Large Computing Systems}
When Moore's law loses efficacy, super large computing power primarily depends on parallelly stacking up massive computing, memory, and storage resources. For example, the term ``high-performance computing" is used to describe the practice of rapidly networking a large number of computers into a single ``cluster" to do intensive computations. Thanks to cloud computing, organizations now have the option to increase the capacity of their high-performance computing programs.

\subsubsection{High-Performance Computing}
High-performance computing (HPC) allows users to handle massive volumes of data faster than a traditional computer, allowing for greater insight and competitive advantage. Over the next decade, scientists will see a $10$-$100$ times increase in sensitivity and resolution from their instruments, necessitating a comparable scale-up in data storage and processing capacity. The data derived by these upgraded instruments will push Moore's law to its limits, posing a threat to conventional operating models predicated primarily on HPC in data centers~\cite{petty2022facing,brayford2020deploying}. 

Conventional HPC architectures were developed for simulation-based methods like computational fluid dynamics. On the contrary, applications were developed to use the underlying technology accessible to programmers. Modern HPC systems include a wide variety of hardware components (e.g., processing, memory, communication, and storage). A measure of this heterogeneity can be seen in the diverse characteristics of applications integrated with techniques such as machine learning. The convergence of HPC and AI has led to the development of novel approaches to old issues and the formulation of new applications.

The AI platform, which increases the effectiveness of scientific discoveries through AI, provides an integrated workspace for development and computing. Researchers can avoid laborious environment settings and computer resource management thanks to the AI platform~\cite{mucha2020artificial}. Although researchers would want to submit AI workloads to HPC clusters directly, doing so is impractical due to the need for extensive administration and scheduling procedures. To encapsulate heterogeneous infrastructure and create a consistent setting for researchers, HPC-based AI platforms are becoming more popular. In the future, researchers will increasingly use interdisciplinary approaches that use a variety of resources (including data, HPC, and the physical world) to address a wide range of problems ~\cite{cheng2020large,wang2008linear,wang2020special}.

Recently, a growing number of technology firms have looked at platforms that use comparable AI technologies. Numerous AI platforms have been established for various study disciplines because of the development of architecture and the constant expansion of processing power~\cite{yao2022venusai}.
\begin{itemize}
	\item IDrug~\cite{IDrug} by Tencent offers a drug development platform that integrates the strengths of the latest algorithms, databases as well as hardware. The operating time for computer-aided drug search iteration is significantly reduced by utilizing powerful computing resources (e.g., NVIDIA GPU). IDrug facilitates creating and aggregating fresh data while integrating several current databases. IDrug offers services related to preclinical drug development, covering protein structure prediction, visualization, synthesis routing, and molecular design. 
	\item EasyDL~\cite{EasyDL} features a thresholdless deep learning platform from Baidu Brain that utilizes P4 and P40 GPUs from NVIDIA's Tesla series for the majority of machine learning workloads. For fundamental tasks, the PaddlePaddle framework and the AI workflow engine are combined~\cite{yang2021automatic}. Typically, business researchers with training in AI development should use the EasyDL.
	\item Amazon AI~\cite{AmazonAI} (AWS) leverages Amazon Web Services via the cloud. The major characteristics of Amazon AI are flexibility, configurable, and simplicity of installation. AWS supplies a full range of resources, including a variety of popular Python tools and libraries, besides security features.
	\item VenusAI~\cite{yao2022venusai} is a supercomputer-based method that extends the virtualization and containerization of primary hardware. VenusAI provides a technology mechanism for aggregating and allocating diverse resources. VenusAI also has a uniform interface for resources at the layer of application services.
\end{itemize}

The abovementioned platforms range from commercial cloud deployments to industry-specific platforms requiring complex integration with scientific investigations. This necessitates the creation of an AI platform with powerful processing capabilities for scientific research.

\subsubsection{Edge, Fog, and Cloud Computing}
Cloud computing has existed as a well-established paradigm since 2006~\cite{mell2011nist}. It allows application deployment and scalability by abstracting underlying computation, storage, and network infrastructure. In a cloud data center, numerous homogeneous, highly-capable computers are linked together by a highly-reliable, redundant network~\cite{yousefpour2019all}.

The IoT era has been shaped by the widespread addition of computing capability thanks to recent advancements in processors, memory, and communications technology~\cite{taivalsaari2017roadmap}. Smartwatches, smart city power grids, and smart building devices that monitor physiological data are all examples of this emerging field. In light of advances in mobile computing and the widespread desire for these devices to function together, a circumstance has emerged in which many different types of devices are all involved in providing the same service or program (e.g., a health monitoring app). These new computational needs are typified by the requirement of a local computation paradigm, which is not met adequately by cloud computing owing to its aforementioned features~\cite{costa2022orchestration}.

Fog Computing is a kind of distributed cloud computing in which resources such as data, computing, storage, and applications are located not in a centralized data center but rather in other nodes across the cloud and its underlying data sources. It's a method for controlling many dispersed networks, some of which may be virtualized, all of which provide data processing and transmission facilities between sensors and cloud storage facilities~\cite{bonomi2012fog}.

Edge computing enables remote devices to process data locally, at the network's ``edge" on their own, or with the help of a nearby server. Moreover, only the most crucial data is transported to the central data center for processing, drastically reducing latency~\cite{cao2021survey}. In an edge computing scenario, terminal devices may communicate with a nearby base station to offload processing-intensive jobs. After completing a job, the edge server sends the results to the terminal device. While the end result of this job handling is comparable with those of cloud computing, edge servers rather than centralized cloud servers are responsible for delivering the required services to terminal devices. By moving distributed services closer to the physical locations of events, edge computing can potentially significantly decrease service latency for end-user devices.

\begin{figure}[H]
	\centering
	\includegraphics[width=0.95\textwidth]{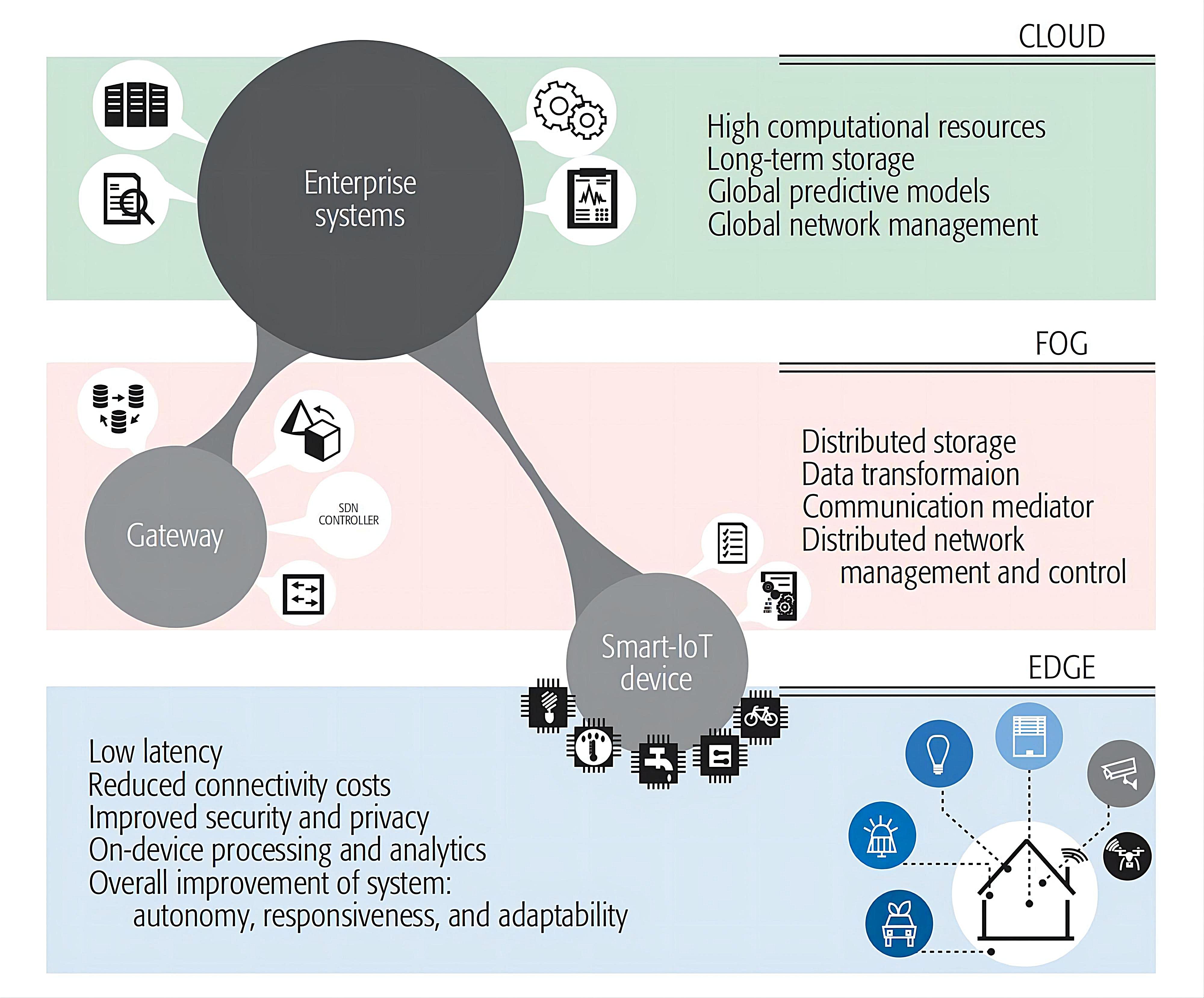}
	\caption{Representation of cloud, fog, and edge computing
\cite{alam2018orchestration}.}
	\label{fig:cloud_fog_edge}
\end{figure}

Figure~\ref{fig:cloud_fog_edge} illustrates the representation of cloud, fog, and edge computing. In certain cases, the terms ``edge" and ``fog" are used synonymously~\cite{definingStefan2022}. Contrary to popular belief, fog computing does not only rely on edge computing. Conversely, fog computing might be used via edge computing. In addition, the cloud is included in the fog when it is not in the edge. Accordingly, the fog must exist at a position intermediate somewhere between the edge devices and the cloud. It acts as an intermediary between the network and the edge devices, supporting local computing for analysis.

In most cases, edge servers can't compete with the power and flexibility of cloud servers when it comes to computation. As the number of endpoints continues to rise, the demand for edge servers might become too much to handle. And since edge computing is a distributed computing paradigm, an edge server can only utilize the information locally to the node where it is located instead of the entire data set. According to these results, edge computing is not optimal for global decision-making. However, due to its centralized nature, cloud computing has the potential to provide not just substantial computing capabilities but also a service for international decision-making. Based on these findings, researchers have proposed the concept of edge-cloud computing, which brings together the advantages of both edge and cloud computing~\cite{cao2021survey}. However, note that in an edge computing system, multiple servers can cooperate with each other securely (e.g., by utilizing a Blockchain platform) to serve the terminal devices and thereby improve the utilization of the edge servers~\cite{Angelo2021Task,Angelo2021Blockchain}. However, the coordination among the servers will involve some overhead.

There are several advantages to using a hierarchical and collaborative edge-fog-cloud architecture, such as the ability to spread intelligence and computing to find an optimum solution within the bounds of the given restrictions (such as the tradeoff between delay and energy)~\cite{firouzi2022convergence}. Obtaining a sustainable integration of edge, fog, and cloud computing necessitates overcoming several problems with design, implementation, deployment, and assessment because of the hierarchical, cross-layer, and dispersed structure of this paradigm.

\subsection{Emerging Computing Architectures}
The goals of architectural innovation to boost digital computing include more efficient energy management, reduced power consumption, cheaper total chip cost, and quicker detection and correction of errors. AI accelerators may drastically cut down on training and execution time when it comes to certain AI operations that can't be performed on a CPU. In-memory computing is an extremely favorable choice because it facilitates memory cells to conduct primitive logic operations so they can compute without the necessity to interact with processors, which is a major contributor to the widening speed gap between memory and processor.

\subsubsection{Accelerators}
In the near term, architectural specialization using a variety of accelerators will be the best way to keep computing power growing. Because a transistor prototype built in the laboratory typically takes around ten years to be integrated into a general manufacturing process. However, no viable alternatives have been exhibited so far. Consequently, it is almost a decade past the deadline to find a practical post-CMOS solution to this issue. Architectural specialization is the only viable option for hardware in the next decade without a viable alternative. Hardware specialization was hard to keep up with in an evolving universal computing environment. As a result of long lead times and expensive development, specialization was not an appropriate solution. While the slowing of Moore's law, as argued by Thompson and Spanuth~\cite{thompson2018decline}, renders architecture specialization a practical and affordable substitute for full universal computing, it will have far-reaching consequences for algorithm design and programming environments ~\cite{shalf2020future}.

\begin{figure}[H]
	\centering
	\includegraphics[width=0.99\textwidth]{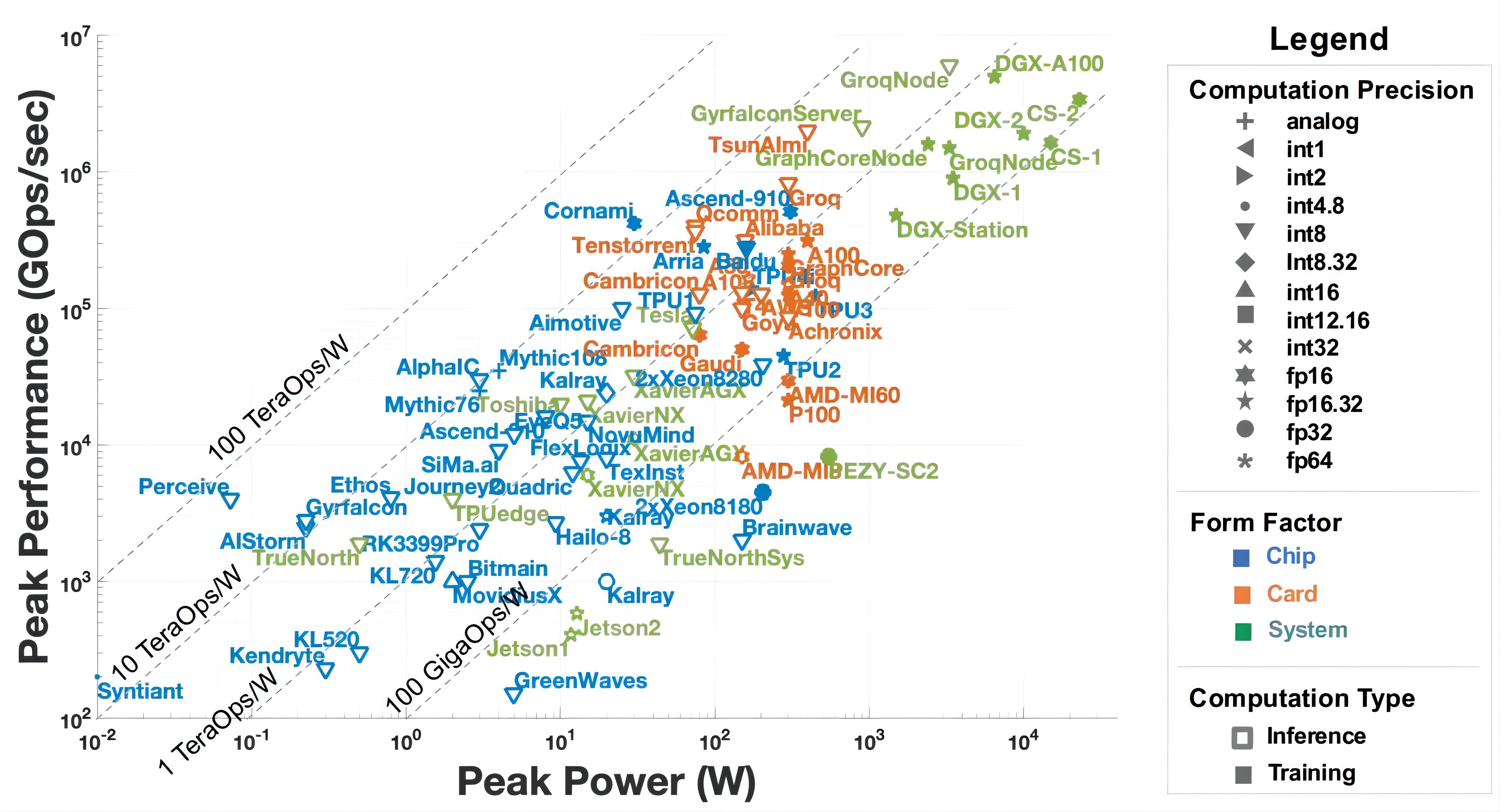}
	\caption{Peak performance vs. power scatter plot of publicly announced AI accelerators and processors~\cite{reuther2021ai}.}
	\label{fig:accelerators}
\end{figure}

As shown in Figure~\ref{fig:accelerators}. Peak power (x-axis) and peak giga-operations per second (y-axis) are shown on a logarithmic scale. Take heed of the caption on the right, which explains the numerous characteristics used to categorize computation accuracy, form factors, and inference/training. The geometry used to indicate the precision of the calculation may take many different forms, including analog, int1, int32, fp16, and fp64. It's easier to see what volume of computing power is being crammed into a computing element when the form factor is represented with different colors. This study only includes setups with a single motherboard and one physical memory slot. Finally, solid geometric figures represent the performance of accelerators built for both training and inference, whereas hollow geometric objects represent the performance of inference-only accelerators.

Some of the newest and finest chips from companies like Alibaba and Groq, as well as recent offerings from NVIDIA and Intel, have peak power consumption far beyond 100W and were developed with inference in mind. The trend over the last several years has changed with this. Both accelerators are designed for driverless cars and data centers, indicating that the power budget for these technologies has increased to more than 100W. Previously, other numerical precisions were the norm for integrated devices, autonomous vehicles, and data centers; however, int8 has since supplanted them. Several accelerators support not just int8 for inference but also fp16 and/or bf16. Finally, the ellipse representing data center systems reveals rising rivalry for high-end training nodes. Nodes from NVIDIA and Cerebras are among the most performing, and there are also notable contributions from Graphcore and Groq. Though Google TPUs and SambaNova are also competitors, they have only reported multi-node benchmark results rather than the peak capability of their systems on a single node.

Accelerators are, therefore, the most effective tools to ensure constant performance gains expected by all scientific computing users; however, Accelerators should be driven by a clearly defined use case. As a result, there is a special need for the fields of study to emphasize certain features of data science for purposes of analysis and simulation. Some of the biggest names in the IT sector have been discussing how next-gen HPC systems are becoming much more diverse. As a result of these long-term improvements in hardware design, it won't be easy to maintain the efficacy and performance increase of HPC systems in the future.

\subsubsection{In-Memory Computing}
It's clear that the way computers are used is rapidly evolving. According to von Neumann's model of computation, a computer retrieves the data and code it needs to carry out its instructions from a central repository called memory. Nevertheless, the performance gap between the memory and processor is widening despite improvements in memory devices. Breakthroughs like deep learning and IoT have a particularly severe case of this issue. Since handling such massive datasets exceeds the capabilities of the von Neumann architecture, such applications provide significant difficulty.

When memory cells are given the ability to execute elementary logic operations, computing-in-memory (CIM) becomes a viable option since it can compute independently of a central processor unit~\cite{nii201413,xue201928,chang201517,lin20167}. Several alternative computer architectures based on CIM that break from the von Neumann paradigm have been suggested. Such designs often use cutting-edge technology and a thoughtful mix of tried-and-true and novel techniques to boost computing performance. Improving the performance of such systems demands considerable synthesis, which provides significant obstacles to application translation. Similarly crucial is the issue of verifying such CIM frameworks.

An appropriate CIM design for the requirements of future computing demands requires a careful balancing of technological and architectural options. One such modern tech that has already impacted the computer industry is the memristor. Scientists are investigating implementations of memristors for many reasons, including their low-temperature manufacturing technique, non-volatile resistive switching, and compatibility with CMOS. Because of its status as a relatively new technology, memristors have several limitations, such as geographical and temporal variations in device performance and an absence of reliable simulations. 

Static random access memory (SRAM)~\cite{jeloka201628,dong20170} or nonvolatile memory~\cite{xue201924,yao2020fully} may both be used to implement CIM, namely SRAM-CIM or nvCIM~\cite{jhang2021challenges}. Static random access memory (SRAM) may be used to construct CIM. SRAM-CIM or non-volatile memory (nvCIM). nvCIM allows storing weight data even while the system is inactive, so it is unnecessary to retrieve data from a processor upon powering it up. Because of its low durability and high write energy, nvCIM can only be used on systems with adequate memory to retain all data required for the specific application. In contrast, SRAM-CIM is well suited for low to medium-capacity systems. It can be configured to function with various neural networks thanks to its quicker write rates, cheaper write energy, and significantly better (nearly infinite) endurance. The latest logic technologies may also be used with SRAM-CIM, reducing latency and improving power efficiency.

\begin{figure}[H]
	\centering
	\includegraphics[width=0.95\textwidth]{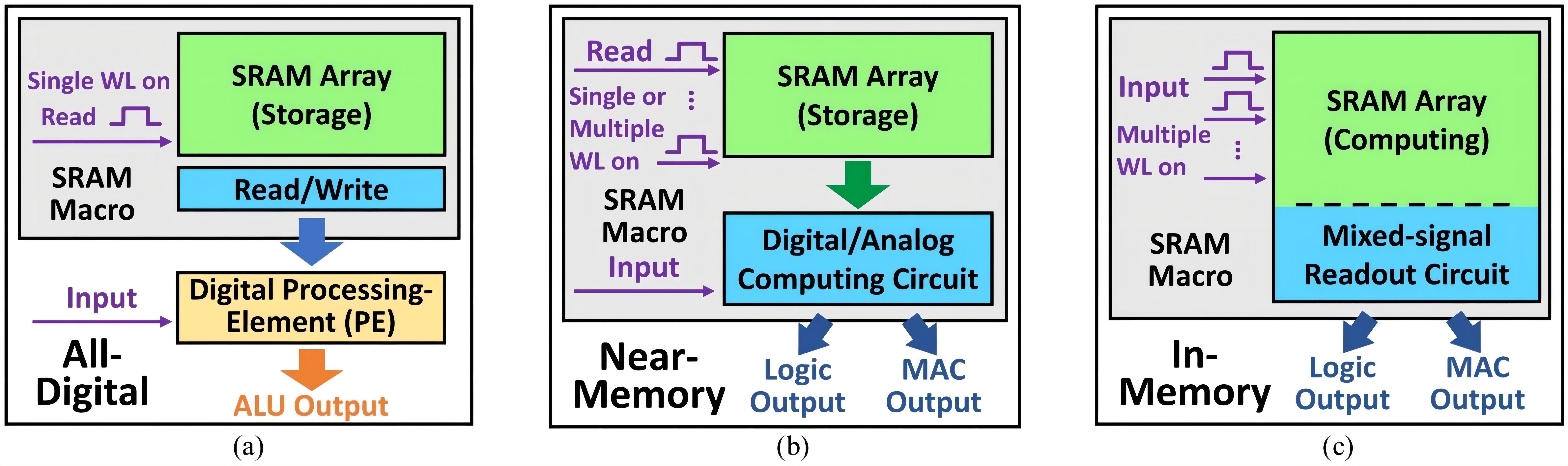}
	\caption{Three conceptual approaches to computing: (a) conventional digital computing, (b) near-memory-array computing (NMAC), and (c) in-memory-array computing (IMAC)~\cite{jhang2021challenges}.}
	\label{fig:compute_in_memory}
\end{figure}

According to~\cite{jhang2021challenges}, in terms of computational structure. Memory macro and digital processing elements are set up as two distinct blocks in the traditional von Neumann design~\cite{chen2016eyeriss,park2016energy,jouppi2017datacenter}. On the other hand, CIM macros do both information exchange and computation throughout the single memory window. As shown in Figure~\ref{fig:compute_in_memory}, CIM is classified into either near-memory array computing (NMAC) and in-memory array computing (IMAC). 

\begin{itemize}
	\item  NMAC: Data is stored using the NMAC structure's memory cells like those of a conventional memory device~\cite{gonugondla201842pj,wang201914,yang201924}. The memory macro is outfitted with an individual interface to connect a simulant or electronic circuit with the memory cells. NMAC circuits are used to compute the digital or analog MAC using the output weights and inputs from outside the circuit. Digital MAC procedures store NMAC's outputs in output registers. 
	\item  IMAC: Various input techniques are used to input data and execute analog computations using the memory cell array~\cite{zhang2016machine,khwa201865nm,guo20195}. Each SRAM cell multiplies a binary weight and an input once during MAC computation. The results of analog computation on the bitlines are subsequently converted into digital outputs. For example, consider the binary fully connected network-using MAC computing approach in~\cite{khwa201865nm}. All the IMC data for a particular column are added together to get the accessible bitline's analog voltage because of the MAC operation. Then, an ADC circuit transforms this voltage of the accessible bitline into a digital output.
\end{itemize}

%%%%%%%%%%%%%%%%%%%%%%%%%%%%%%%%%%%%%%%%%%%%%%%%%%%%%%%%%%%%%%%%%%%%%%%%%%%%%%%%%%%%%%%%%%%%%%%%%%%%%%%%%%%%%%%%%%%%%%%%%%%%%%%%%%%%%%%%%%%%%%%%%%%%%%%%%%%%%%%%%%%%%%%%%%%%%%%%%%%%%%%%%%%%%%%%%%%%%%%%%%%%%%%%%%%%%%%%%%%%%%%%%%%%%%%%%%%%%%%%%%%%%%%%%%%%%%%
\subsection{Emerging Computing Modes}
The presence of complexity is frequently to blame for the failure of traditional computers. If a supercomputer gets stumped, it is likely because a particularly difficult task was presented to the large classical machine. Moreover, the ubiquitous use of today's highly complex AI models (e.g., DNNs) in edge devices remains elusive. It is attributed to the deficiencies that there is a power and bandwidth crunch for premium GPUs and accelerators operating these models, which results in long processing times and cumbersome architecture designs~\cite{wetzstein2020inference}. In light of these facts, researchers are spurred to create novel computing modes, such as neuromorphic and photonic computing, biocomputing, and mind-bogglingly disruptive quantum computing.

\subsubsection{Quantum Computing}
Since entering the big data era, demand for data processing speed is increasing. At the same time, the computational power of classical computers is gradually reaching its limit. Quantum computing, however, can surmount this limitation since it has the quantum advantage brought on by entanglement or other non-classical correlations, achieving exponential speed in many complex computational problems. This advantage of quantum computing could bring a huge potential to deal with extensive information in a short time and become a promising candidate for next-generation computing technology.

In the early 1990s, Elizabeth Behrman started combining quantum physics with AI. Most scientists thought the two disciplines were just like oil and water and could not be combined. But now, when chemists and biologists begin to learn quantum mechanics, the combination of computer science and quantum mechanics seems very natural. Additionally, the evolution of computers is greatly influenced by quantum information technology.
Compared with the classical bits 0 and 1 considered by computer science, quantum physics began to consider whether such classical bits could be replaced by quantum bits for operation. Due to its superposition state, this qubit may stand for either 0 or 1, depending on the context. Superposition exists in many quantum systems, including the two orthogonal polarization directions of photons, the spin directions of electrons in a magnetic field, and the two spin directions of nuclear spin, all of which have applications in quantum computing.
The superposition of quantum systems gives quantum computing the advantage of parallel computing, which improves its speed significantly. The quantum computer can improve the computing power exponentially compared with the classical computer~\cite{arute2019quantum,zhong2020quantum}. Besides, if we study AI from the perspective of quantum computing, we may not need a very advanced general-purpose quantum computer. Most of the time, a specific-function quantum processor can satisfy an AI algorithm and exhibits the quantum advantages~\cite{huang2022quantum,huang2021power,bennett1996mixed,aaronson2007learnability,chen2022exponential,degen2017quantum}, and this can be achieved very soon.

\begin{figure}[H]
	\centering
	\includegraphics[width=0.95\textwidth]{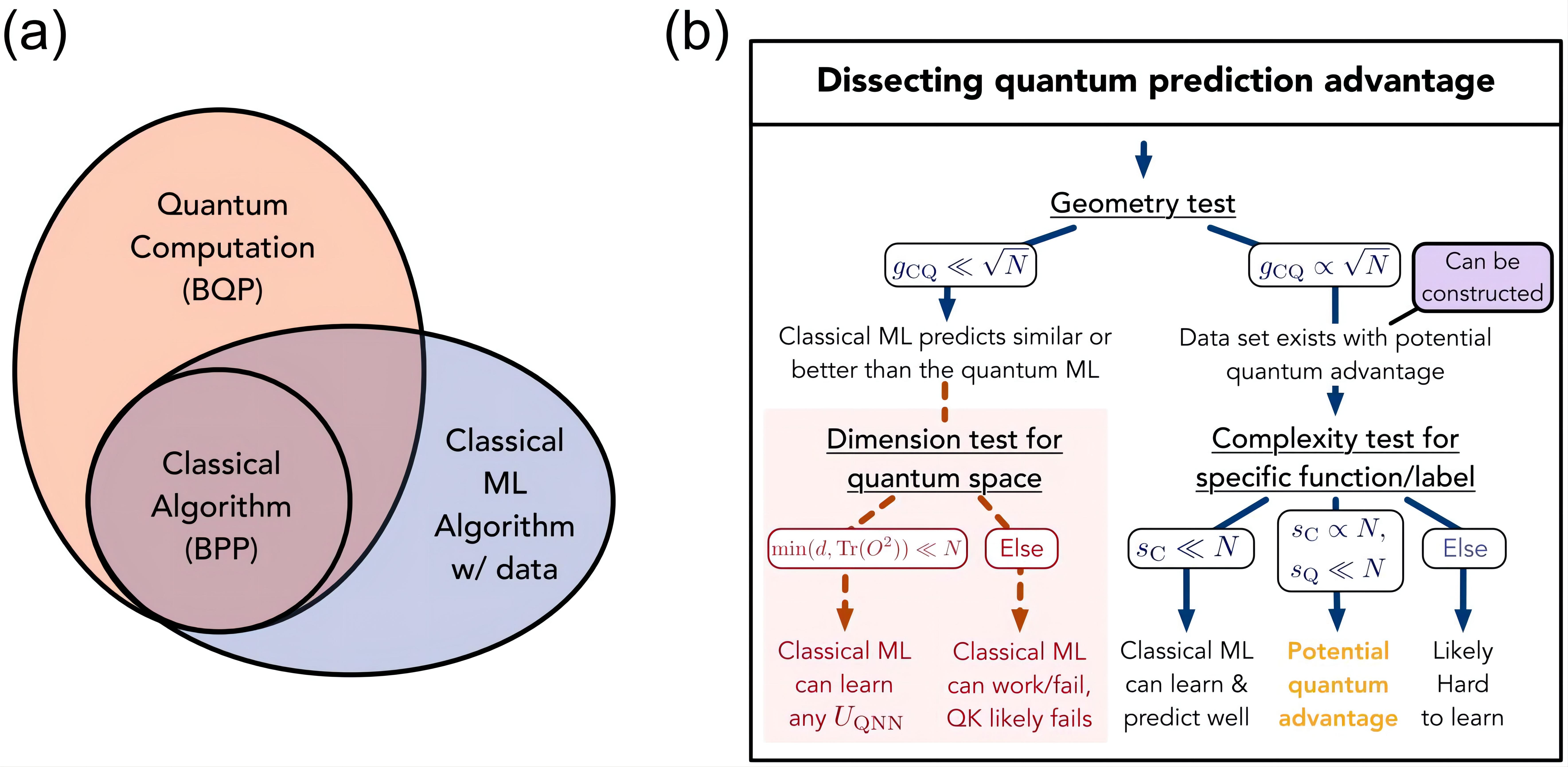}
	\caption{Diagram showing the relationship between complexity classes and a flowchart for identifying and evaluating possible quantum advantages~\cite{huang2021power}. a) it is illustrated how adding more data may increase the complexity in several ways. It is believed that quantum computing can efficiently solve issues that conventional ML algorithms with data cannot. Because classical algorithms that are able to learn from data belong to a complexity class that can handle issues that go beyond classical computation. b) the flow chart is the developed methodology for analyzing the feasibility of a quantum prediction advantage. Quantum and classical procedures with associated kernels, as well as $N$ samples of data from a QNN with potentially unlimited depth using encoding and function circuits $U_{enc}$ and $U_{QNN}$, are supplied as input. The importance of the data for a potential prediction advantage is emphasized by presenting the tests as functions of N. Before even thinking about the function to learn, a geometric quantity called $g_{CQ}$ may be evaluated to determine the likelihood of a positive quantum or classical prediction separation. If the test is successful, we demonstrate how to build an adversarial function that reaches this limit effectively; otherwise, the traditional technique is guaranteed to provide the same level of performance regardless of the data function. After the model complexity $s_C$ and $s_Q$ have been determined, a label/function-specific test may be conducted to evaluate the actual service provided. The red dashed lines show whether the quantum kernel (QK) approach can determine whether a simple classical function may represent any given encoding of data.}
	\label{fig:qunatum ai}
\end{figure}

In recent years, AI and quantum computing have continued to heat up and gradually become two major research hotspots. Quantum artificial intelligence is an interdisciplinary frontier subject combining these two hot topics. At present, people believe that if one of the data or algorithms is quantum, it can be summarized into the category of quantum artificial intelligence. Two significant concerns exist in this emerging discipline. One is using advanced classical machine learning algorithms to analyze or optimize quantum systems and solve problems related to quantum mechanics. The other one is establishing a quantum learning algorithm based on quantum hardware and using the parallelism of quantum computing to improve the speed of the machine learning algorithm. Finally, there is another situation: the algorithm is quantum, and the data is quantum, but there is no substantive progress in this field. 

Quantum artificial intelligence has broad application prospects~\cite{arrazola2021quantum}. For example, quantum artificial intelligence has been applied in the synthesis of drugs~\cite{banchi2020molecular} and the treatment of various chemical reactions~\cite{huh2015boson}. Despite the rapid development of quantum artificial intelligence, it is still in its initial stage. Many applications are still limited by the number of quantum bits and the bit error rate caused by environmental noise in the specific function quantum computer. Therefore, now we study quantum artificial intelligence from the quantum perspective. We all consider how to build a scalable system and ensure that quantum bits receive the least noise in the calculation process. We believe that quantum artificial intelligence will bring the fifth wave to the world after several years or even decades of development.

For now, we must deal with a lot of data every day. There is an association between these data. Graph algorithms can get much helpful or hidden information from the relationship between these data. Graph computing, the core technology of next-generation AI, has been widely used in many fields such as medical treatment, education, military, finance, and so on. However, when the scale of the graph is large, the computing resource requirements will greatly increase. For example, the problem of finding the largest fully connected subgraph is an NP-hard problem. Now, we think about whether we can use quantum computing to improve the speed of graph computing. Gaussian Boson sampling has been proven with quantum superiority many times, and at the same time, we find that the graph can be encoded into a Gaussian Boson sampling machine~\cite{arrazola2018using,sempere2022experimentally}. Namely, we can use the sampling results to quickly find the maximum number of fully connected subgraphs (cliques)~\cite{banchi2020molecular}.

\subsubsection{Neuromorphic Computing}
Carver Mead first uttered the term neuromorphic in the 1980s~\cite{mead2020we,mead1990neuromorphic}, at which time it mainly involved hybrid analog-digital forms of brain-inspired computing. Nevertheless, a considerably wider spectrum of hardware is now considered to fall under the umbrella of neuromorphic computing because of the growth of the field and the appearance of significant funding for brain-inspired computer systems.

Non-von Neumann computers are those that resemble neurons and synapses. Their construction and operation are inspired by neurons and synapses in the brain. Alternatively, a neuromorphic computer has neurons and synapses that control processing and memory. In contrast to von Neumann computers, neuromorphic computers construct their programs utilizing parameters and the neural network's topology instead of predefined instructions. One subclass of neuromorphic approaches relies on generating and manipulating ``spikes" in analog neural networks. The frequency that spikes appear, their magnitude, and their shape can be employed to store numerical data in neuromorphic computers, whereas von Neumann computers encode information as binary values. The conversion of binary values into spikes and vice versa is still a subject of study in neuromorphic computing~\cite{schuman2019non}. 

\begin{figure}[H]
	\centering
	\includegraphics[width=0.99\textwidth]{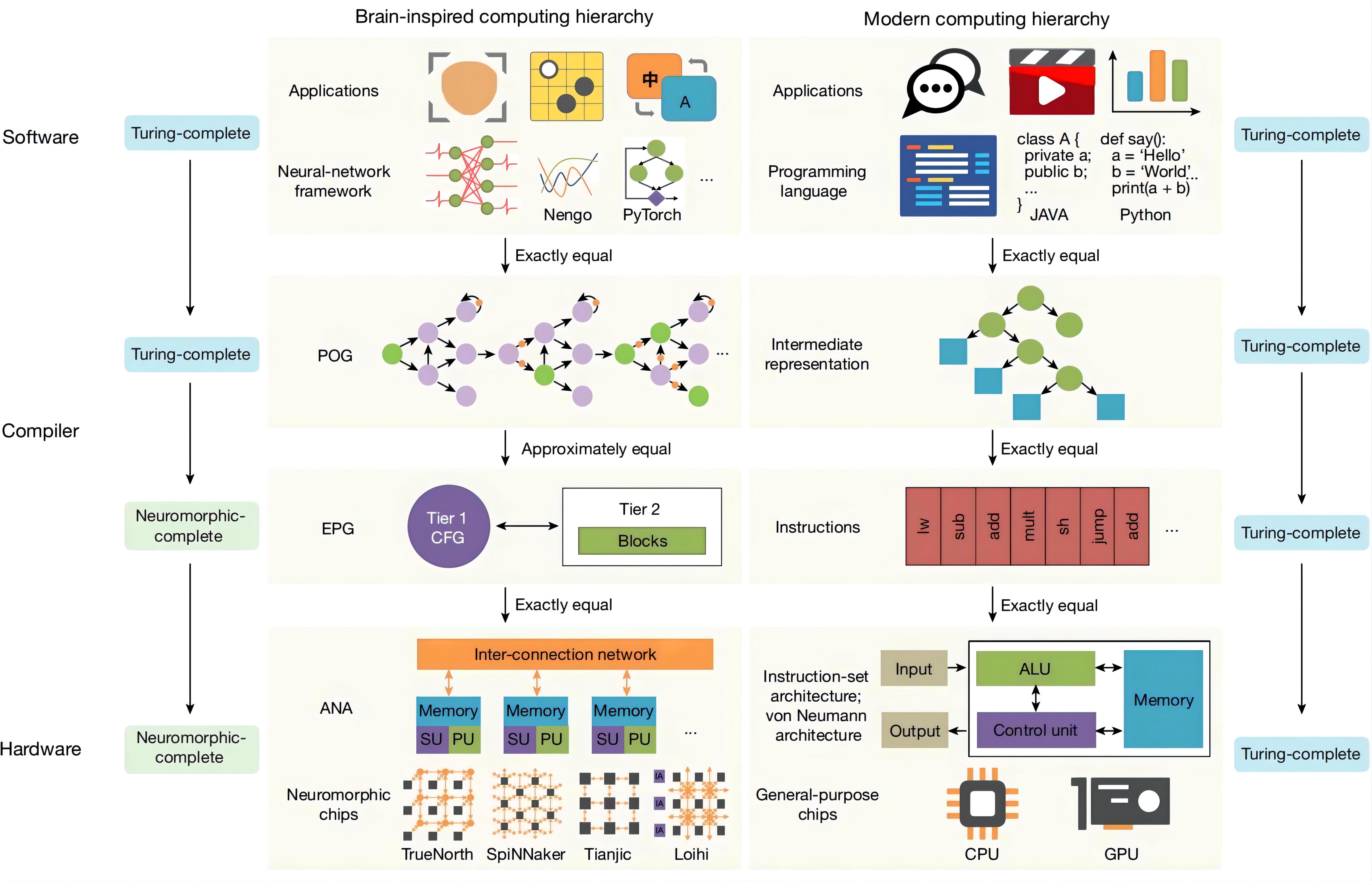}
	\caption{Structures of conventional computing systems and brain-inspired computing systems~\cite{zhang2020system}. The figure illustrates that based on the three-tiered structure of conventional computing systems (right), a brain-inspired computing system (left) consisting of software (on the top), a compiler (in the center), and hardware (at the bottom) were proposed. Applications and Turing-complete programming languages (like JAVA and Python) make up the software layer of the architecture of a conventional computer system. Intermediate software representations, such as the abstract syntax tree, are transformed into hardware representations, such as instructions, throughout the compilation process. The instructions are executed by CPUs or GPUs that adhere to the von Neumann architecture at a hardware level. ALU, CPU, ROM, RAM, and I/O are all von Neumann architecture components. Turing completeness guarantees the exact equality of all stacks. The neuromorphic applications and frameworks for their development constitute the software layer of a computer system inspired by the human brain (such as Nengo and PyTorch). The POG represents software at this stage, and the EPG represents hardware at this stage (CFG, control-flow graph). The POG is presented before the compilation tools are used to convert it to the EPG. To abstract the neuromorphic hardware, a hardware layer called ANA was proposed, which consists of scheduling units (SUs), processor units (PUs), memory, and an inter-connection network (TrueNorth, SpiNNaker, Tianjic, and Loihi). Neuromorphic completeness, on the other hand, offers not only exact equivalence but also approximation equivalence to account for the approximation feature of brain-inspired computing.}
	\label{fig:neuromorphic}
\end{figure}

The two designs operate differently from one another because of their differing traits~\cite{schuman2022opportunities}.

\begin{itemize}
	\item High parallelism: Since all neurons and synapses have the ability to function concurrently, neuromorphic computers are by their very nature parallel. In contrast to von Neumann systems, neurons and synapses carry out comparatively straightforward calculations.
	\item Co-located processing and memory: In neuromorphic hardware, processing and memory are not separated. In many cases, synapses and neurons carry out processing and storing values, although neurons are often regarded as processing units and synapses as memory units. Incorporating processors and memory units can alleviate the von Neumann limitation regarding processor or memory division, resulting in slower maximum performance. Furthermore, co-location reduces the time data is accessed from main memory, a practice common in conventional computing that consumes a substantial volume of energy compared to computing.
	\item Intrinsic scalability: Since adding more neuromorphic chips entails adding more neurons and synapses, neuromorphic computers are naturally scalable. One may think of a combination of many physical neuromorphic chips as a huge neuromorphic system to operate ever larger networks. Numerous massive neuromorphic hardware systems have been effectively put into use, such as SpiNNaker~\cite{mayr2019spinnaker,furber2014spinnaker} and Loihi~\cite{davies2018loihi}. 
	\item Event-driven computation. Neuromorphic computers are able to perform extremely efficient computations due to event-driven computation~\cite{mostafa2015event,amir2017low}. During the execution of the network, neurons, and synapses only carry out computations when spikes are present, and spikes are relatively sparse.
	\item Stochasticity: Neuromorphic computers can incorporate stochasticity.
\end{itemize}

Neuromorphic computers are extensively described and cited in publications as motivations for adoption~\cite{schuman2017survey,davies2021advancing}. Neuromorphic computers are ideally suited for computation because of their energy efficiency: they typically run with a fraction of the power of traditional computers. They consume very little power because they are event-driven and highly parallelized, meaning that only a fraction of the system works simultaneously. Energy efficiency is a sufficient motivation to explore the implementation of neuromorphic computers, given the increasing energy consumption of computing and the emergence of energy-constrained programs (i.e., edge computing). In addition, neuromorphic computers are ideally suited to modern AI and ML applications, as they inherently perform neural network-like operations. Additionally, neuromorphic computers have the potential to handle multiple kinds of computations~\cite{aimone2018non}.

\subsubsection{Photonic Computing}
Architecture specialization is bringing more data center demands like accelerator technologies for machine learning workloads, and rack disaggregation approaches are also putting pressure on current interconnect technologies. Even though the newest high-throughput processor chips feature multiple CPU/GPU cores that can perform extremely difficult computations, they lack the off-chip bandwidth needed to make the most of their resources. Taking on this challenge requires overcoming packaging limitations, which are directly related to the limited bandwidth density of current electrical packages~\cite{shalf2020future}.

\begin{figure}[H]
	\centering
	\includegraphics[width=0.95\textwidth]{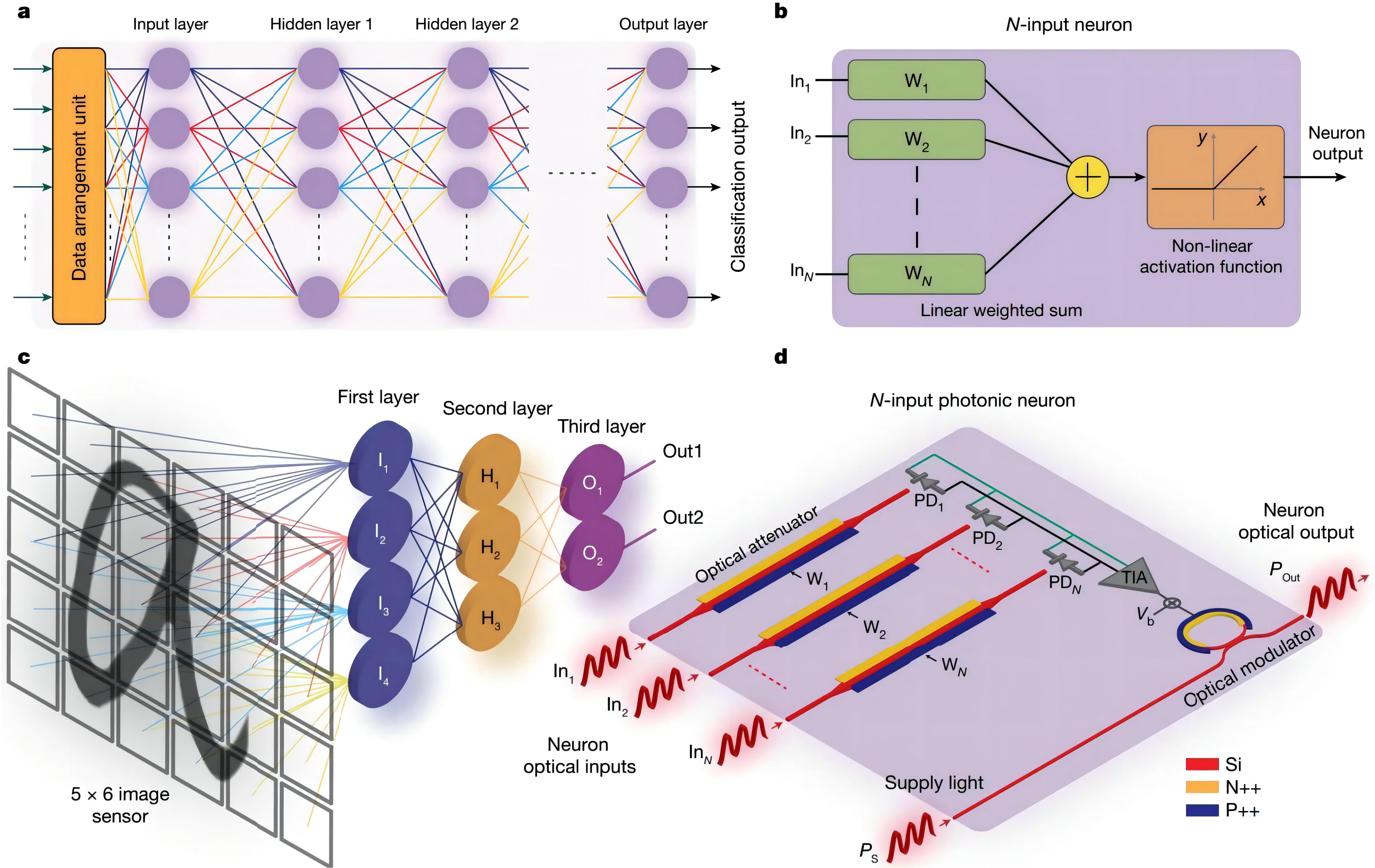}
	\caption{Deep neural networks, including conventional and electronic-photonic~\cite{ashtiani2022chip}. a) A typical block architecture for a deep neural network comprises an input layer, multiple hidden layers, as well as an output layer that produces outputs for classification or regression. b) In this network design, a conventional N-input neuron is employed. Its output is formed by processing the linear weighted sum of its inputs via a nonlinear activation function. c) In accordance with the PDNN chip architecture, separated from one another by overlap, the input picture is split into four smaller images on a five by a 6-pixel array. Pixels that constitute portions of images are sent to the primary layer of neurons. The connections between the second and third tiers and the layers below them are obvious. Two outcomes are conceivable for the network. d) The architecture of a real-world $N$-input photonic neuron, in which optical PIN attenuators are used to change the weights of N optical input signals, and the summed output of parallel PDs is then used to perform photodetection. A TIA is used to amplify and voltage-convert the photocurrent $i_{sum}$. Adjusting the supply light makes it possible to produce the optical output of neurons.}
	\label{fig:photon}
\end{figure} 

Optical neural networks (ONNs) offer many benefits over electrical neural networks, including ultra-high bandwidth, fast calculation speed, and high parallelism, all of which are realized by using photonic hardware acceleration to calculate complicated matrix-vector multiplication~\cite{kitayama2019novel,wetzstein2020inference,shastri2021photonics}. To keep up with the ever-increasing complexity of data processing techniques and the volume of datasets, we need deeply integrated and scalable ONN systems with compact sizes and decreased energy usage. Light's superposition and coherence features allow ONN neurons to be naturally coupled by interference~\cite{shen2017deep} or diffraction~\cite{lin2018all} in diverse contexts, while a wide range of nonlinear optical effects~\cite{zuo2019all} may be used to implement the activation function of the neurons physically. Because of these tools, other types of neural network topologies, such as fully connected~\cite{shen2017deep,lin2018all,meyer2017disaggregated}, convolutional~\cite{xu202111,feldmann2021parallel,miscuglio2020massively}, and recurrent~\cite{bueno2018reinforcement,larger2012photonic}, may now be realized optically. With today's state-of-the-art optical technologies, ONNs can perform ten trillion operations per second~\cite{xu202111}, which is comparable to electrical counterparts, with energy consumption that may be on par with or even less than one photon per operation~\cite{wang2022optical}, which is orders of magnitude lower than digital computation~\cite{spall2022hybrid}. Silicon photonic integrated circuits (PICs) are becoming an attractive option for building the massive and compact processing units needed in optical-artificial-intelligence computers because of their small size, high integration density, and low power consumption~\cite{o2009photonic,kues2017chip,feldmann2021parallel,zhu2022space}.

\subsubsection{Biocomputing}
Biological computing is a new computing model developed using the inherent information-processing mechanism of biological systems. In short, it is used to solve computational problems with biological methods. Biological computing mainly focuses on devices and systems. Devices, also known as molecular devices, are the basic units for information detection, processing, transmission, and storage at the molecular level; a system refers to the design of a new computing system utterly different from the traditional computing architecture. Generally, the system is a distributed system.

\begin{figure}
\centering
\subcaptionbox{The cell as a ``physical" computer.\label{fig:bio1}}
  {%
    \includegraphics[width=0.45\textwidth,align=c]{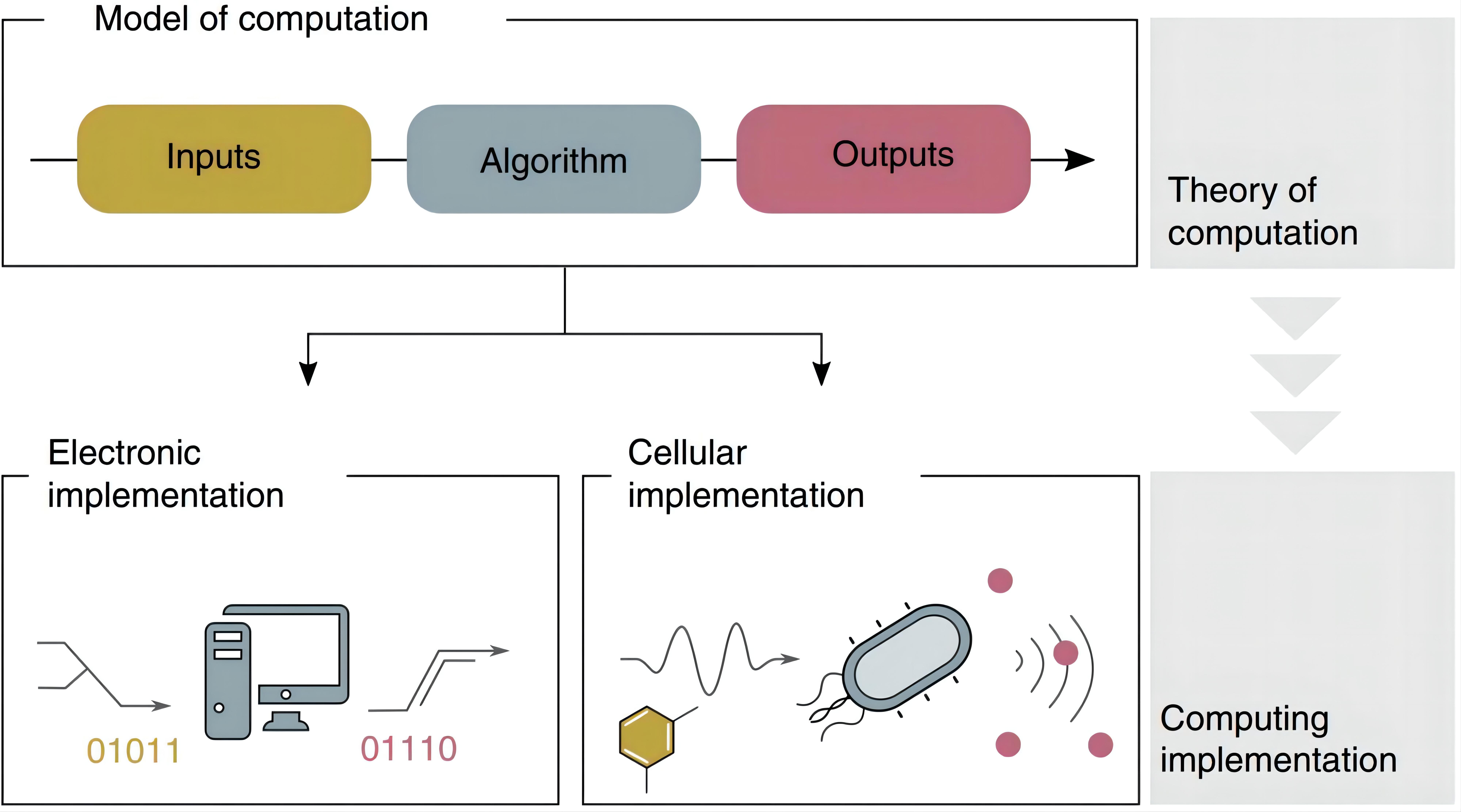}%
    \vphantom{\includegraphics[width=0.45\textwidth,align=c]{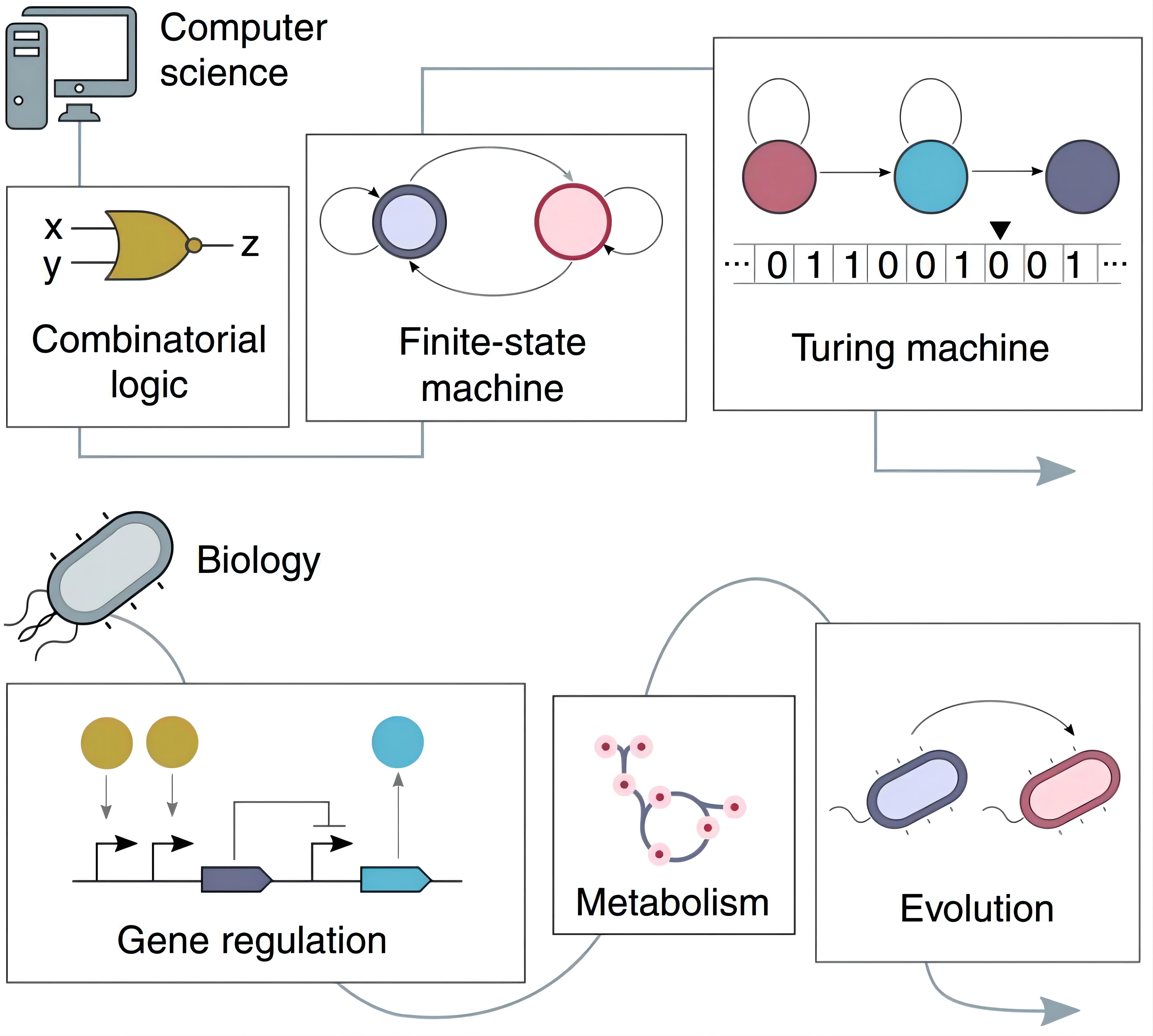}}%
  }\qquad
\subcaptionbox{Cells could provide more than logic circuits.\label{fig:bio2}}
  {\includegraphics[width=0.45\textwidth]{figures/biocomputing_b.jpg}}
\caption{Biocomputing might offer performance superior to that of traditional computers~\cite{grozinger2019pathways}. a) inputs and outputs and the processing of inputs by an algorithm are technically defined in a computing model. Although there are various physical implementations of the same theoretical computing model, the essence of computation is consistent regardless of the specific implementation. Electrical data also makes up the inputs and outputs for electronic implementations. Still, cells can also detect and transmit a wide range of physical, chemical, and biological data flow. Data may be encoded into inputs using a variety of techniques. Examples of temperature encoding systems. b) More complex computation models than combinatorial logic have been established in computer science. The Turing machine and finite-state machines are examples of this. These models outperform combinatorial logic because they enable the processing of a wider range of inputs into a wider range of outputs in a wider range of ways. Living systems' cells can process information because a variety of computing mechanisms have evolved throughout time. A simple model that serves as the basis for creating combinatorial logic circuits in cells represents the fundamental tenets of molecular biology. However, the model does not take into account essential biological systems like metabolism or processes like evolution that may pave the way for the development of more sophisticated, as-yet-unknown models.}\label{fig:bio}
\end{figure}

Biological computers mainly include protein computers, RNA computers, and DNA computers. The protein computer takes the law of protein motion as the basic prototype of computer operation. The researchers of Syracuse have used protein as the computer's core device and the laser to read the information. The storage capacity is 300 times larger than the electronic computer, opening the era of protein computers. RNA and DNA computers use the specific hybridization between nucleic acid molecules as the basic model. Because RNA is inferior to DNA in differentiating molecular structure and experimental operation, few people have paid attention to RNA computing. DNA computer takes biological enzymes as the basic material and biochemical reactions as the process of processing information, trying to improve the efficiency of computer processing information in the way humans process information. Adleman first proposed a DNA computer in 1994. After numerous studies and practices, although it is still in its infancy, the powerful storage capacity and parallelism shown by DNA give DNA computers a huge potential.

Biological computing has unique advantages compared to traditional computing, which can be summarized as the strong parallel and distributed computing ability and low power consumption. Parallel computing and distributed computing are the modes designed by traditional computers to solve large-scale and complex computing problems. But biological computing naturally has incomparable advantages over parallel and distributed computing. Second, the biochemical connection process in biological computing requires molecular energy and does not require additional external energy, and the overall energy consumption is very low. For example, the energy of 1 joule can complete more than 1000 calculations for DNA, while the traditional silicon-based computer can only complete more than 100 calculations. There is an order of magnitude difference.

Within the scope of current technical capabilities, biological computing inevitably has deficiencies. Limited by the existing biological technology, most current biological computers are designed on paper, and there are no suitable conditions for relevant experimental verification, let alone construction. For example, in the DNA computer, how to reuse DNA or protein to meet the requirements of continuous consumption of DNA in the calculation process; the existing DNA computers are all dedicated to a specific field. It is also complex in making standard and universal computer components; DNA involves biological privacy information. Protecting citizens' DNA information from being infringed on and used by criminals is a significant social problem.

\section{Applications of Intelligent Computing\label{Sec:Applications}}
\subsection{Intelligent Computing for Science}
Discovering innovative ideas with the same old methods isn't going to stagnate if we're going to keep up with the ever-increasing problems of our rapidly evolving environment. However, the pace of scientific discovery will be tremendously boosted like never before by the confluence of computer revolutions now underway.

\subsubsection{Computational Materials Science}
Computational materials (CM) have become a powerful means of studying materials' properties and designing new materials for several decades. Their applications, however, suffer from many challenges due to the complexity of materials and material behaviors, including lacking force fields and potentials for many atoms, ions, and atomic and ionic interactions, different thermodynamic phases in molecular dynamics (MD) simulations, and the huge search space for the optimization of material components and process parameters. AI integration into CM is shown to be a revolution to the traditional CM as a new research paradigm~\cite{agrawal2016perspective}. Intelligent CM is one major component of materials informatics~\cite{ramakrishna2019materials} and is becoming increasingly popular. The number of relevant publications has reached 50 thousand on the Web of Science. About 70 percent were published in the past five years.  

The integration of AI and CM is exhibiting great success in multiple lengths and time scales and multiple physical field coupling calculations. The most famous electronic and atomic scale calculation method is first-principle calculations by applying the Density Functional Theory (DFT). The key issue in DFT calculations is the huge demands on computational powers due to the multiple particles and nonlinear interactions in the Schrödinger equation for electronic structures. The Deep Neural Network might be an efficient way to accelerate the calculation process of the electronic Schrödinger equation~\cite{shen2018self,hermann2020deep}. The approximations of exchange-correlation (XC) energy in DFT limit the accuracy of the Kahn-Sham DFT calculations. Kernel ridge regression and deep neural networks can create more accurate XC approximations~\cite{margraf2021pure,kalita2021learning}. ML-based XC approximations can even be applied in systems with strong correlations. By substituting time-consuming electronic structure calculations with empirical potentials, MD can simulate much larger systems with defects in different temperatures. ML can provide a systematic method to derive force fields or potentials from first-principle calculations. The ML-based force fields are called ML potentials. ML potentials consist of two parts: Data and ML potential model. During data collection, prior knowledge of the studied system plays a central role in the design of candidate structures which should be calculated using first-principle calculations. The design of descriptors for local structures is the heart of the ML potential model~\cite{yang2021taking}. Several efficient ML potential packages have been published including Amp~\cite{khorshidi2016amp}, MLIP~\cite{novikov2020mlip}, MLatom~\cite{dral2022mlatom}, and DeepMD~\cite{zhang2020dp}. Xu \textit{et al.} investigated Li-Si alloys using molecular dynamics based on a model that was built from DeepMD~\cite{xu2020deep}. Various crystalline and amorphous Li-Si systems were analyzed for their structural and dynamic features. Their prediction was 20 times faster than ab initio molecular dynamics simulations with similar accuracy. This method can also be applied to insulating materials like liquid water. Grace \textit{et al.} introduced a model for insulating materials and applied it on liquid water~\cite{sommers2020raman}. It shows that the Raman spectra associated with classical 2-nanosecond trajectories under a fixed temperature could be computed, and the resolution of low-frequency Raman spectra was enhanced. Li \textit{et al.} applied DeepMD on the solid-state electrolyte Na3OBr~\cite{li2021theoretical}. The $Na+$ diffusion coefficients at finite temperature were obtained, suggesting the influence of temperature on the migration barrier. Their work also demonstrates the promising future of DeepMD in the study of transport properties of solid-state electrolytes. Recently, He \textit{et al.} studied the structure phase transitions of SrTiO3 using DeepMD~\cite{zhang2020deep}. The temperature-driven phase transition characters under different in-plane strains were studied via the model build using DeepMD.  

\begin{figure}[H]
	\centering
	\includegraphics[width=0.95\textwidth]{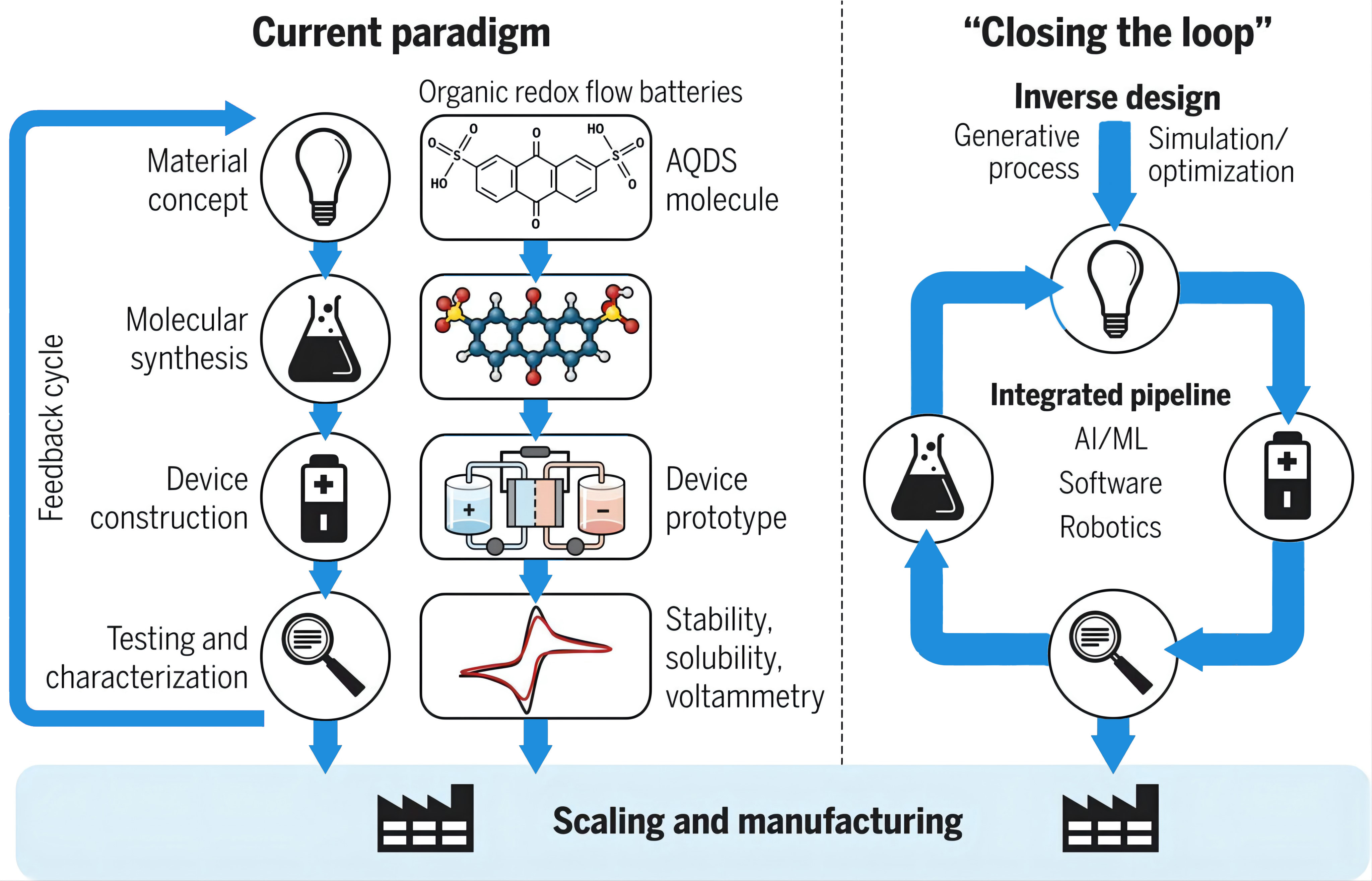}
	\caption{High-level comparison of paradigms for materials/molecular sciences~\cite{sanchez2018inverse}. The conventional paradigm is described on the left and illustrated using organic redox flow batteries in the middle. On the right is a model of a closed-loop system. Inverse engineering, intelligent software, AI/machine learning, embedded systems, and robotics are all necessary components of a closed-loop system.}
	\label{fig:materials}
\end{figure}

Phase field simulations can illustrate the microstructure evolution over time at the continuum thermodynamic and kinetical levels. But they require huge computational powers. LSTM networks, as a famous algorithm of gated RNN, were successfully applied to train a model to predict results of a long time evolution from data collected from calculations over a very short time period~\cite{montes2021accelerating}. In finite element calculations of materials, the key issue is constructing a constitutive model for specific materials in a given service environment. Many ML models are able to construct a constitutive model from data, for example, Gaussian processing~\cite{wang2021metamodeling}, Artificial Neural Networks~\cite{liu2021review,zhou2022frame,huang2020learning,liu2020learning,chen2021recurrent,masi2021thermodynamics} and symbolic regressions~\cite{kabliman2019prediction,sun2019data}. 

Another application of ML in computational materials is to train an ML surrogate model, especially a simple analytic surrogate formula, which is used to substitute the original true physical model~\cite{bhosekar2018advances,forrester2009recent}. Data are collected from several calculations of the physical model with different input parameters and applied in ML-model training. Usually, the ML surrogate model can be evaluated far faster than the original physical model, while both models have nearly the same accuracy. The several orders acceleration on computational speed allows for a global search in the design and optimization space. Typical ML algorithms that can be applied to train a multi-fidelity surrogate model include Kriging/Gaussian process~\cite{rocha2021fly}, LSTM Networks~\cite{montes2021accelerating,khandelwal2021machine}, Physics-informed Neural Networks~\cite{haghighat2021physics}, and CNNs~\cite{messner2020convolutional}.

For cases without available physical models, a surrogate model can be trained directly from experimental data to substitute a yet-unknown physical model. The situation is the most common in materials society, and the model is usually called the material microstructure-macroproperty relationship. Features or descriptors in machine learning of materials include electronic and atomic parameters, chemical composition, microstructural parameters, thermodynamic and kinetic parameters, processing conditions, service environment conditions, all material characterization conditions, optical and electron microscopy images, etc. The output of machine learning models can be either target properties or potential energy surfaces~\cite{zhang2020dp}. Depending on the studied problems, some parameters are features in some problems and become responses in other problems. Target properties include stability~\cite{park2019learn}, formability, bandgap~\cite{gladkikh2020machine}, Curie temperature, dielectric properties, flexoelectricity~\cite{li2017quantification} and so on. Stability plays an imperative role in predicting new materials and the formation energy, which can be obtained from the first principles calculations. Li \textit{et al.}~\cite{li2019computational} developed a transfer learning method to predict formation energy. After screening 21,316 perovskites, they found 764 stable perovskites with a tolerance factor of less than 4.8. Ninety-eight of them have already been proven stable by DFT calculation. Recently, Park \textit{et al.} studied the stability of hybrid organic/inorganic compounds using a series of machine learning models and proved that the combination of advanced electronic structure theory and machine learning promotes designing new materials~\cite{takahashi2018searching}. The bandgap is an important parameter in designing novel photovoltaic devices. The solar cell requires a bandgap that meets the wavelength of visible light. In 2018, Takahashi \textit{et al.} predicted perovskite bandgap to search candidates for solar cells using machine learning~\cite{park2019learn}. They predicted 9,238 perovskite materials to have the desired bandgap, and 11 of them were undiscovered.

\subsubsection{Computing for Astronomy}
Astronomy has gathered vast amounts of data in history as one of the most ancient observational sciences. Thanks to the breakthroughs in telescopic technologies that generate digital outputs, there has recently been a tremendous data explosion. The field of astronomy and astrophysics is characterized by a wealth of data and a variety of ground-based telescopes with big apertures, for example, the upcoming large synoptic survey telescope and the space-based telescopes~\cite{kremer2017big}. Data collection is now more efficient and largely automated using high-resolution cameras and associated tools. The system will collect roughly 15TB of data daily~\cite{tallada2020cosmohub}. With respect to effective decision-making, it is imperative to have more effective data analysis. Hence, intelligent computing techniques are needed to interpret and evaluate that dataset.

\paragraph{Morphological classification of galaxies.} After years of waiting and anticipation, the first images captured by the James Webb space telescope were finally released on July 12, 2022. A machine learning model called Morpheus creates morphological classifications of astronomical sources at the pixel level. Morpheus is trained on UC Santa Cruz’s Lux supercomputer, which consists of 28 GPU nodes with two NVIDIA V100 Tensor Core GPUs each. Machine learning models rapidly evolve into incredibly effective tools in cosmology and astrophysics. For example, CNNs and generational adversarial networks (GANs) have been successfully applied to facilitate the classification of galactic morphologies based on star formation and morphological properties~\cite{dieleman2015rotation,ascaso2015galaxy,dominguez2018improving,gan2021seeinggan}. It has been proven that these ML algorithms can achieve over 90\% accuracy and perform equally well or even superior to conventional methods with much fewer time budgets.

\paragraph{Radio frequency interference detection.}Newer studies have shown that U-Net~\cite{ronneberger2015u} and its variants provide a strong architectural foundation for semantic segmentation, which is a crucial component of deep learning-based radio frequency interference (RFI) detection. U-Net was initially implemented for RFI detection in radio astronomy~\cite{akeret2017radio}. A combination of computer-generated data and observed data captured by a signal antenna at Bleien Observatory was used to train and test the network~\cite{chang2017integrated}. After evaluating a U-Net variant in terms of detecting RFI using synthetic and real data collected at the HERA observatory~\cite{kerrigan2019optimizing}, for improved generalization to other representations, the authors split off the scale and the time period into separate elements in the model. Combining the scale and the time period depictions of the intricate visibility yielded only marginal benefits, as is demonstrated in~\cite{mesarcik2020deep,vafaei2020deep}. Transfer learning has consistently been found to be effective in situations where labeled data are absent. For instance, R-Net can be trained on simulated data and employ a tiny part of expert-labeled data. Its domain may be adapted from simulated to real-world data~\cite{vafaei2020deep}. GANs have been demonstrated to be useful for RFI detection in~\cite{vinsen2019using}. A very novel way to use GANs was proposed in~\cite{vos2019generative}. The authors suggested a source-separation strategy to differentiate astronomical signals from RFI based on employing two independent generators.

\subsubsection{Computing for Pharmaceutical Research}
AI has been affected by all drug design phases~\cite{hessler2018artificial,schneider2020rethinking,paul2021artificial}. Drug design benefits from AI as it helps scientists establish 3D structures of proteins, the chemistry between medications and proteins, and the efficacy of drugs. In pharmacology, AI is used to create targeted compounds and multitarget medications. AI can also design synthetic routes, predict reaction yields, and understand the mechanics behind chemical synthesis. AI has made it simpler to repurpose current medications to treat new therapeutic objectives. AI is vital for identifying adverse reactions, bioactivity, and other drug screening outcomes.

\begin{figure}[H]
	\centering
	\includegraphics[width=0.95\textwidth]{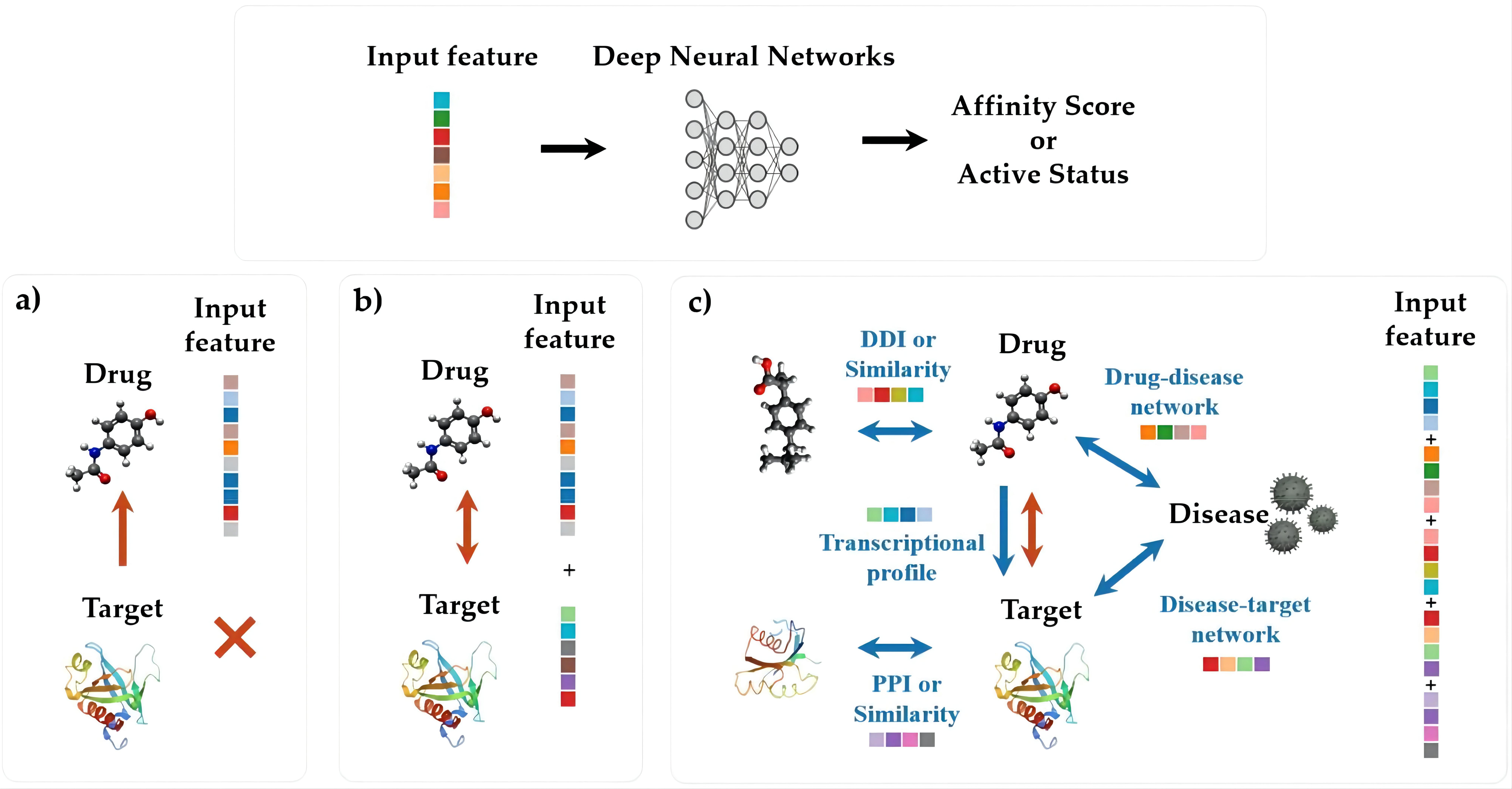}
	\caption{Different DL-based DTI prediction algorithms take different input characteristics into account~\cite{kim2021comprehensive}. Therefore, they may be divided into three categories: a) ligand-based approach, b) structure-based approach, and c) relationship-based approach.}
	\label{fig:medcine}
\end{figure}

Recent AI tools and platforms for drug design are as follows~\cite{doytchinova2022drug}.
\begin{itemize}
	\item AlphaFold, the groundbreaking computational model, estimates the 3D structure of proteins exclusively using their amino acid sequence generated by DeepMind and EMBL-EBI~\cite{jumper2021highly}. According to the latest CASP14 analysis, AlphaFold provides the most precise estimation of 3D protein structures~\cite{jumper2021highly}. In addition to considering different constraints (evolutionary, physical, and geometric) related to protein structures, AlphaFold implements a neural network architecture based on protein data banks.
	\item SwissDrugDesign~\cite{duvaud2021expasy}, a product of the Swiss Institute of Bioinformatics, is one of the most widely used AI platforms for drug design.
	\item Synthia by Merck, an upgraded version of Chematica, suggests potential synthesis routes based on compound information. The AI application can provide multiple synthetic routes for the target molecule by adjusting the search options. Chematica was developed by Klucznik \textit{et al.} to generate synthetic procedures for eight commonly occurring compounds and subsequently conduct experiments on them. Each of the compounds has shown a significant increase in productivity and cost reduction compared to conventional techniques~\cite{klucznik2018efficient}.
	\item Ligand Express from Cyclica identifies potential targets associated with certain macromolecules. Instead of screening large collections of macromolecules to locate the suitable ligand to bind to certain proteins, an advanced platform built on the cloud screens the human proteome to discover the optimal matching protein and proteins~\cite{mackinnon2021proteome}.
	\item AstraZeneca's AI platform REINVENT is used to design macromolecules from scratch. It can produce macromolecules that comply with a broad range of preferences entered by the user~\cite{zhong2018artificial}.
\end{itemize}

There are several different AI platforms and tools for drug research and discovery on the web, and new ones are continually emerging. The current study does not have the space to describe them all in-depth, but other excellent evaluations~\cite{hessler2018artificial,schneider2020rethinking,paul2021artificial,jimenez2021artificial,cavasotto2021artificial,chan2019advancing} analyze and compare them.

The detection of active compounds through huge chemical libraries is one of the initial stages in the drug development process~\cite{thomas2022applications}. High-throughput screening (HTS) now rules this phase~\cite{shoichet2004virtual}. Large chemical libraries are screened via HTS using assays relevant to the study. Instead of calculating behaviors in silico, it offers the benefit of testing them empirically. HTS is not constantly necessary yet. Large libraries are expensive to experimentally screen because they only contain a tiny portion of the chemical space.
Additionally, not all assays can be carried out on a large enough scale; generally, negotiation must be made between the amount and quality of experimental data acquired for each test to get the best possible results. Virtual screening (VS) is an alternate method that may be used in addition to or instead of HTS~\cite{chen2009pharmacophore,muthas2008possible}. By screening chemicals in silico rather than in vitro, which is more affordable and is not constrained by a physical library, VS aims to overcome the drawbacks of HTS. VS usually enriches actives, raises hit rates, and lowers the cost of subsequent tests~\cite{zhu2013hit}. This is particularly true when a distinct design hypothesis, like a verified target, is present.
Nevertheless, VS is imprecise and prone to producing inaccurate predictions, much like several other in silico techniques. Once this occurs, inactive molecules may be classified as false positives, wasting time and important resources on further research. Therefore, increasing VS's enrichment rates is still necessary.

Stephen Oliver and Ross King from the University of Manchester created two robots, Adam and Eve, which are the icing on the automation and present use of AI in drug creation~\cite{doytchinova2022drug}. Adam was built to do microbiological experiments, analyze the data on its own, propose hypotheses, and create experiments to examine the hypotheses until a correct theory was established~\cite{king2004functional}. The robot Eve is more sophisticated; it experimentally screens hundreds of compounds each day, identifies certain hits, constructs a specific cell line to test the hits, and then modifies the structures of the hits to produce lead compounds~\cite{williams2015cheaper}.

In the pharmaceutical industry, there is a general trend toward the use of advanced manufacturing technologies, with a strong emphasis on connected and efficient processes like continuous manufacturing, new technologies suited for personalized and on-demand medicine (like 3D printing), and an ongoing effort to find solutions for problematic compounds in the pipeline~\cite{seven2022}. The COVID epidemic made us reevaluate ways to quicken the processes of medication and vaccine research and development. Digitalization, difficult substances, and a quick pace bring a tendency for modeling, predictive methods, and digital cooperation in the pharmaceutical sector. Additional and unique difficulties, such as protein stabilization and purification, are brought on by the growing number of biomolecules in the pipeline.

\subsubsection{Computer-Aided Breeding}
Food security is now a world concern, partly because of the fast population expansion, which is anticipated to reach nine billion people by 2050~\cite{hatem2022artificial}. Approaches such as tissue culture mutagenesis and transformation have been used to improve crops. Functional genomics improves our understanding of the plant genome and opens new opportunities for tinkering with it. Promising methods, like nanotechnology, RNA interference, and next-generation sequencing, have been developed to boost agricultural output in response to future needs~\cite{rashid2017crop}.

Crop breeding has lately seen a growth in the use of AI technologies, which support the creation of services, the identification of models, and decision-making processes in agri-food applications and supply chain stages. The main objective of AI in agriculture is to anticipate outcomes with accuracy and improve yield while minimizing resource use~\cite{patel2021smart}. Therefore, AI tools provide algorithms that may evaluate performance, anticipate unforeseen issues or occurrences, and discover trends, such as water consumption and irrigation process management via the installation of intelligent irrigation systems, to handle agricultural concerns~\cite{suprem2013review}.

AI facilitates the whole agricultural value chain, from planting to harvesting to selling~\cite{ben2021artificial}. Therefore, AI advancements have aided the efficiency of agro-based firms by enhancing crop management. Weather forecasting, improving automated equipment for accurate pest or disease detection, and analyzing sick crops to boost the ability to produce healthy crops are common areas where AI is being used. It has paved the way for several tech firms to create AI algorithms that help the agricultural sector to deal with issues that include pest and weed infestations and yield decreases due to global warming~\cite{sujatha2021performance}.

\begin{figure}[H]
	\centering
	\includegraphics[width=0.95\textwidth]{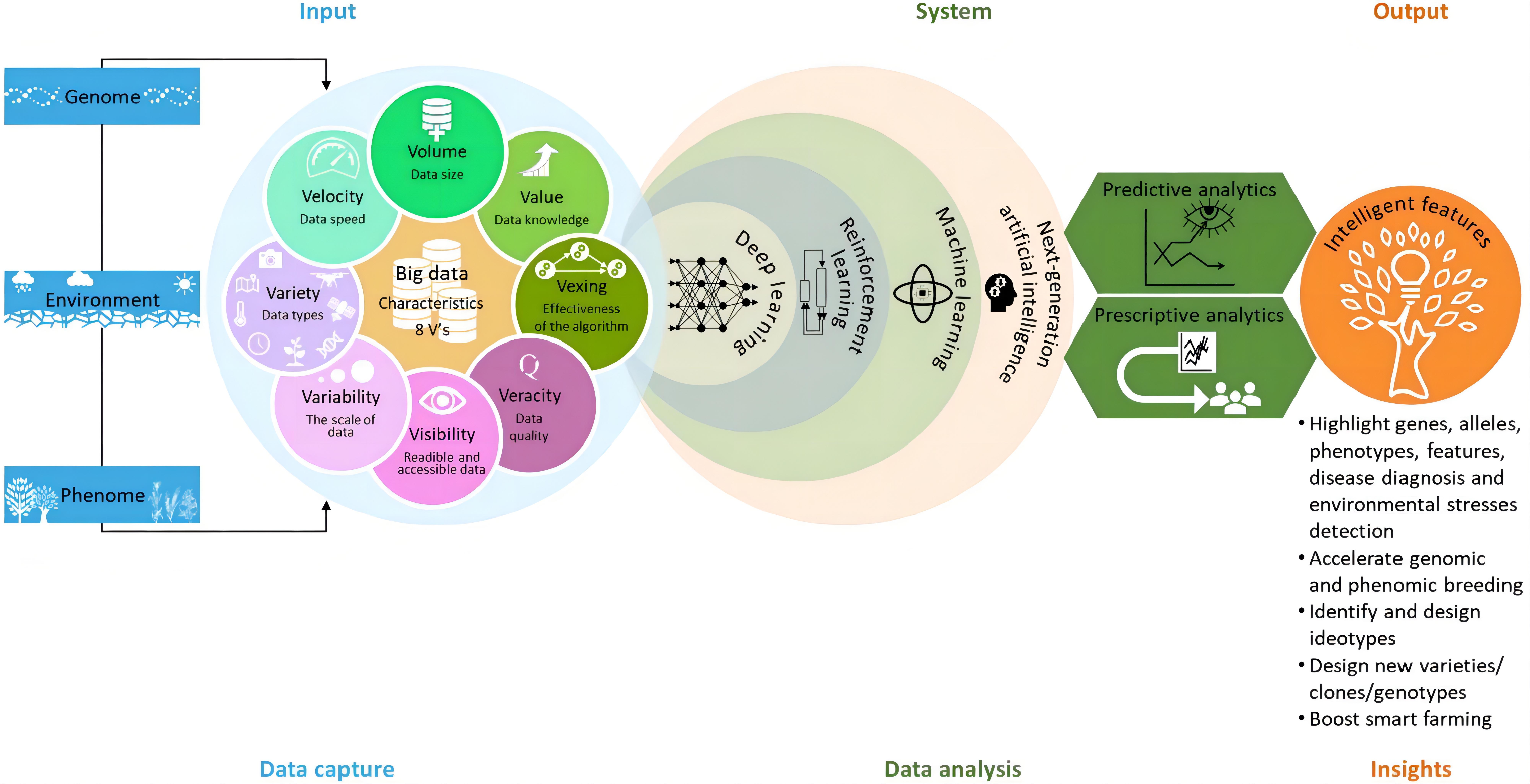}
	\caption{Combination of Big Data and next-generation AI in plant breeding~\cite{harfouche2019accelerating}.}
	\label{fig:plant}
\end{figure}

To maximize yields and profitability while reducing crop damage, farmers should employ technology to predict the weather. AI empowers farmers to gain greater knowledge and understanding by analyzing the data they collect and then acting by putting processes in place that help them make informed choices~\cite{crane2018machine}. Additionally, using either farm flora patterns or pictures taken with a camera recognition tool, AI techniques can monitor soil management and health by recognizing plant pests and diseases, as well as nutrient deficits and potential soil defects~\cite{ben2021artificial}. By lowering the use of pesticides, AI technology offers a huge functional advantage in environmental preservation. Farmers might, for instance, spray herbicides just where weeds are present using AI approaches, including robotics, ML, and computer vision, controlling weeds more effectively and precisely. This would lessen the chemical spray needed to cover the whole field.

The four central clusters of the agricultural supply chain (preproduction, production, processing, and distribution) are becoming more and more relevant for ML algorithms~\cite{ahumada2009application}. ML technologies are used during the preproduction phase to anticipate soil characteristics, crop output, and irrigation needs. In the succeeding stage, ML might be used to detect illnesses and predict the weather. To achieve high and secure product quality, production planning is forecasted using ML algorithms in the third cluster of the processing phase. Finally, the distribution cluster may benefit from ML algorithms, especially in terms of storage, customer analysis, and transportation~\cite{ben2021artificial}.

\subsection{Intelligent Computing for Economy and Governance}
Intelligent computing accelerates transformational change, resulting in the shift of economic and social order. Markets for goods and labor are changing drastically due to technological advancements. The newest developments in AI and associated advancements are pushing the boundaries of the digital revolution in new directions.

\subsubsection{Digital Economy}
There are several potential routes for advancement in AI systems. In general, AI should be at the heart of every data-driven strategy in the digital economy, including Industry 4.0. Predictive maintenance, for instance, may benefit greatly from AI~\cite{grall2002continuous,lee2019predictive}. Predictive maintenance deals with maintenance involving general or production machinery and aids in lowering operating expenses or downtime using sensor data from either production or operating lines.

It is possible to develop and apply AI-based prediction models to improve maintenance schedules. Furthermore, IoT and CPS applications should benefit from AI since these technologies were created for data collection rather than analysis. Finally, AI may contribute to the future development of robotics and automation for use in industrial, manufacturing, and service applications. For such unique AI techniques, deep reinforcement learning is now showing promising results~\cite{mnih2015human,gu2017deep}.
A more basic thing to notice is that general data analysis concepts must also be adjusted for the use of AI. Cross-industry standard procedure for data mining~\cite{shearer2000crisp} is a rudimentary standard that emphasizes feedback between successive analytical processes. This has recently been expanded to consider industry-specific demands and domain-specific expertise~\cite{tripathi2021ensuring}.

Three main issues are commonly brought up regarding the use of AI in business and the economy. The first is job losses because of the adoption of automated analytic systems~\cite{brynjolfsson2014second}; the second is the difficulty in understanding generic AI approaches; the third is the widening wealth disparity between rich and developing nations~\cite{bughin2018notes}. Interestingly, the first two arguments are virtually identical to digital medicine and health systems. AI governance, which must be designed appropriately, addresses the latter problem.

\subsubsection{Urban Governance}
According to recent research, urban governance is to develop novel strategies and methods to make cities smarter~\cite{jiang2019smart}. Smart cities include smart urban governance, Which aims to utilize cutting-edge IT to sync data, procedures, authorities, and physical structures that will benefit locals.~\cite{gil2012enacting}. Meijer and Bolvar~\cite{meijer2016governing} established four exemplary conceptions of smart urban governance based on a thorough examination of the literature: smart decision-making, governance of a smart city, smart administration, and smart urban cooperation. 

Also, numerous promising new avenues of research into the urban brain have been presented~\cite{grossi2020public,jiang2021smart,verrest2019elaborating}. Big data has made it common practice to combine data from various sources and different points of view to provide a complete picture of urban residents. Large-city population development has brought the additional challenge of managing more complicated road networks. Because of this, trafﬁc management also necessitates analysis, forecasting, and smart action~\cite{pereira2018smart}. For instance, cities have been developing integrated traffic management systems for the optimization of traffic flow in real-time, such as City Brain in Hangzhou, China. These technologies take advantage of copious urban monitoring data captured by a variety of sensors. In addition, the difficulty of simulating the urban brain system increases with the complexity of traffic networks. Important considerations include~\cite{deng2021systematic} (1) speeding up computations on enormous synchronized heterogeneous network configurations through parallel heterogeneous computing; (2) visually simulating an urban setting and building algorithms for flexible perception with strong environmental robustness. In terms of crisis management, it will be crucial for future cities to be able to conduct searches rapidly. As a result, it is crucial to search for and recognize individuals in monitoring data by conducting simulated analyses of their characteristics and activities. Urban planning and analyzing public resources are also fascinating fields. The urban brain can accumulate facts according to the rules of urbanization throughout time. It is possible to optimize the design of public facilities and the distribution of government funds via examining such information.

\begin{figure}[H]
	\centering
	\includegraphics[width=0.75\textwidth]{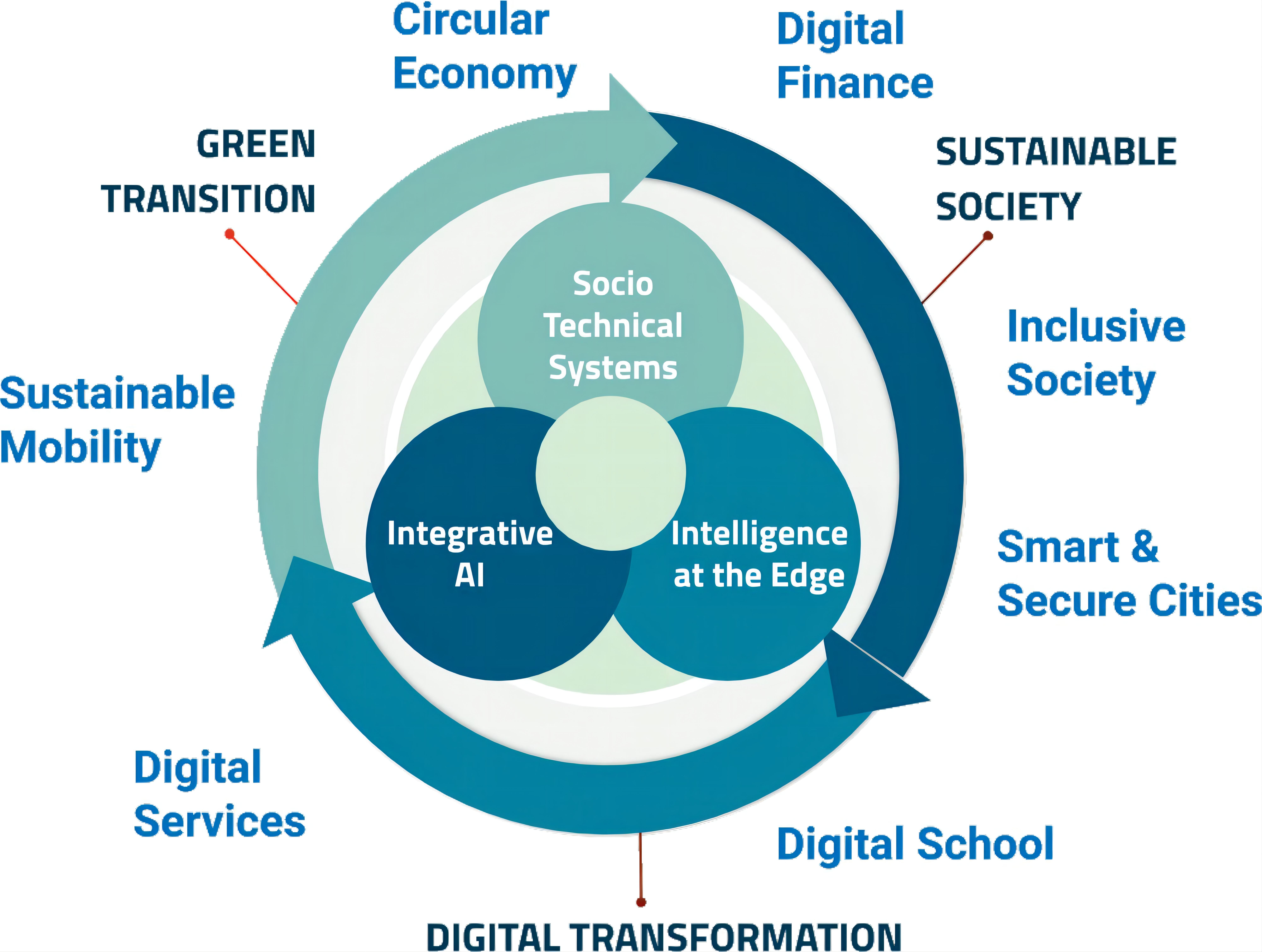}
	\caption{Components of digital society~\cite{DigitalSociety}.}
	\label{fig:smart city}
\end{figure}

Recent research has called for a greater emphasis on the ``urban" component of ``smart governance". Instead of placing excessive weight on technologically created neutral information, urban issues, relevant expertise is understood as socially formed via interaction with people~\cite{mcfarlane2017alternative}. For instance, information that is the product of collaborative efforts between the government, the business sector, and civil society is often poorly organized or, at most, semi-structured. It is required when attempting to solve strategic and unconventional problems. When addressing issues that affect a whole community, we must have access to technological tools that promote dialogue, debate, and the development of agreement~\cite{jiang2021smart}. Democratic institutions, social conditions, ethnic and political values, and the physical world are all examples of contextual elements that may foster or stifle the growth of creative and smart governance. It argues that context should be considered when considering alternatives to the existing ``smart" government~\cite{meijer2016governing}.

%%%%%%%%%%%%%%%%%%%%%%%%%%%%%%%%%%%%%%%%%%%%%%%%%%%%%%%%%%%%%%%%%%%%%%%%%%%%%%%%%%%%%%%%%%%%%%%%%%%%%%%%%%%%%%%%%%%%%%%%%%%%%%%%%%%%%%%%%%%%%%%%%%%%%%%%%%%%%%%%%%%%%%%%%
\section{Perspectives \label{Sec:Perspectives}}
The technical detail of intelligent computing and its main challenges from theoretical and experimental perspectives have been reviewed in the previous sections. In this section, the main challenges and future development of the intelligent computing industry are exposited from the view of an emerging industrial ecology. 

\subsection{Theoretical Revolution in Machine Intelligence}

Compared with conventional computing theory, intelligent computing is the application and development of linguistically and biologically motivated computational paradigms~\cite{goudos2021guest}. It means that machines can mimic the problem-solving and decision-making capabilities of the human mind based on different scenarios. 
However, there are fundamental differences in the underlying logic of silicon-based and carbon-based operations, and the mechanics of brain intelligence still need to be further revealed.
The next step in intelligent computing is to develop a radical theoretical overhaul through an in-depth exploration of the essential elements of human-like intelligence and its interaction mechanism at the macro level~\cite{pearl2018theoretical} and the computational theory underpinning the generation of uncertainty at the micro level. 

According to the theory of Multiple Intelligence by Howard Gardner, human intelligence can be differentiated into specific modalities of intelligence~\cite{gardner2000intelligence}. Depending on the expression of different machine intelligence, it can be disassembled into different combinations of basic abilities. For example, logical-mathematical intelligence is the combination of learning ability, computing ability, and memory ability. Since intelligent computing refers to the simulation and approximation of human intelligence in machines, a basic paradigm, which clarifies the definition and standardizes the definable and computable properties of multivariate intelligence, helps better realize human-like intelligence. It is necessary to design an axiomatic system of multivariate intelligence and prove that it has basic mathematical properties such as decidability and completeness. For multivariate computational intelligence, precise decomposition and quantitative description are needed. Moreover, the calculability and comparability of multivariate intelligence also should be provided through quantified mathematical expressions and measurement criteria of the atomic intelligence fusion. Scientists can develop better human-like machine intelligence through the integration, collision, and interaction between different theories~\cite{legg2007universal,tirri2012measuring,naglieri2015handbook}.

Turing computation, based on the theory of computation, is functional. Classical computation built on Turing computation produces deterministic results. However, the creativity of intelligence is built on uncertainty. Receiving the same background knowledge, different people will have diverse thinking on the same problem. Even in the face of the same problem, the same person makes various choices and judgments at other times and circumstances. This uncertainty is why human intelligence can continuously generate new data, knowledge, and tools. Randomness and fuzziness are two primary forms of uncertainty in the subjective and objective world. At present, the exploration of machine intelligence is built on the classical computing theory, trying to abstract the natural world through artificial preset symbolic systems and algorithmic models, and realize the approximation of randomness and fuzziness. The intelligence of the human brain grows out of the emergence of chemical phenomena such as proteins, particle channels, chemical signals, and electrical signals. To study evolvement from low-level perception to high-level logical reasoning, it is necessary to understand how uncertain emergence occurs and how to reproduce the randomness and fuzziness of the emergence. A neuromorphic network simulates the structure and function of neurons in the brain, but it cannot simulate the process of intelligence generation. Quantum computing may be one of the most promising directions. Quantum mechanics reveals the uncertainty of the fundamental particles that make up the world. This uncertainty may drive the emergence and development of human consciousness and can also be the theoretical basis for constructing high-level machine intelligence. Some research has been conducted to mimic the probabilistic behavior of quantum mechanics in a classical computer~\cite{camsari2019p}. To further develop intelligent computing, we need first to build a computational theory that can support uncertainty to realize a perfect mapping from theoretical computing space to physical space.

\subsection{Knowledge-Driven Computation}
To make the computer learn like humans, scientists have adopted two approaches: Symbolic AI, brought up by expert systems, and Connectionism, designated by deep neural networks~\cite{clancey1987knowledge,smolensky1999grammar}. These two approaches can be good solutions to intelligent computing problems to some extent. The key to the problem still requires prior knowledge inputs such as a pre-defined physical symbol system, neural network models, behavior rules, etc. Data-driven intelligence at the theoretical and methodological level relies mainly on mathematical models and large-scale data input to compute results. In essence, the machine does not produce new knowledge; it only performs a series of numerical calculations based on prior knowledge and, thus, the results. In other words, the machine is just an ``executor", while the actual strategies and logic that enable the derivation and computation of the knowledge are still specified by humans. However, data-fitting learning in small tasks with large data gradually shows limitations. Due to the problem of weak generalization, poor interpretation, difficulty in knowledge expression, lack of common sense, and catastrophic forgetting, the models are far from human understanding. In most cases, humans are better at summarizing from a few practices without learning from large-scale training data. According to their correlation~\cite{popham2021visual}, the brain can transform vision into multiple knowledge. However, the current deep learning frameworks can only simulate human intelligence on the surface~\cite{mitchell2018never}.

Integrating knowledge from different domains with algorithmic models can lead to better problem-solving, on which the prototype of the fifth paradigm of scientific research is based~\cite{hey2009fourth}. Therefore, it is important to explore how humans learn and apply it to the study of AI. The knowledge-driven machine intelligence can learn from human activities and mimic the decision-making capabilities of the human mind, enabling machines to perceive, recognize, think, learn and collaborate like humans. Exploring theories and key technologies for multi-knowledge-driven knowledge reasoning and continuous learning to enable intelligent systems with human-like learning, perception, representation, and decision-making capabilities can facilitate the evolution of intelligent computing from data-driven to knowledge-driven~\cite{pan2019visual}. Combining data-driven inductive abstraction with knowledge-driven deductive reasoning and constrained optimization of physical theorems is a key challenge in improving machine intelligence. 
To achieve the ability to summarize abstract concepts and reach higher levels of intelligence, more flexible system architectures need to be developed to explore the way knowledge is created, stored, and retrieved. At the theoretical level, the knowledge data model needs to be improved, while the model's ability to describe the real world needs to be enhanced. Human-like thinking models are introduced to learn human environmental perception, emotional preference, seeking advantages, and tendency to avoid harm and to construct a computational model of a self-learning system capable of perceiving the environment. 

%Deep learning models are one of the most popular models in the field of intelligent computing. Deep learning models are trained to fit sample data, producing results comparable to and sometimes surpassing human expert performance. Deep learning has been applied to different applications such as speech recognition, computer vision, natural language processing, bioinformatics, etc.~\cite{ciregan2012multi,krizhevsky2012imagenet}. In addition, humans have the abilities such as hypothesis, conjecture, reflection, creativity, association, and exploration to learn from experience continuously. 

%Therefore, it is very important to introduce intelligent ideas into computer software and hardware systems. The code modules and basic components should be treated as the raw materials of the intelligent algorithm, transplant the mature, intelligent methods and models to the software and hardware system, and effectively analyze and calculate the system's code optimization and hardware improvement.

\subsection{Architectural Innovation for Hardware and Software}
Various innovations have been proposed in hardware architectures. But the adaptation between hardware and software faces enormous challenges, such as accuracy loss, invoking difficulty, and collaboration inefficiency. From the theoretical perspective, neuromorphic computing is an effective technology system with apparent advantages in computing for intelligence~\cite{monroe2014neuromorphic, zhao2010nanotube}. Neuromorphic technology can reduce power consumption and improve real-time computational performance by several magnitudes. In addition, neuromorphic technologies are low-cost and easy to implement in many applications. However, the design of neuromorphic computing hardware (neuromorphic chips, SNNs, memristors, etc.) places obstacles in constructing algorithms and models. Although the traditional neural network model has achieved accurate results in modeling domain problems, when transplanting the trained neural network model to SNNs, structural incompatibility can lead to a loss of accuracy. The application of SNNs depends on the development of neurocomputing chips. Due to their new design structure and computing mode, the SNNs cannot achieve the theoretical results in traditional chips.

To narrow the gap between neuromorphic computing hardware and software, the co-design and coordinated development of software and hardware are necessary for data management and analysis in the new hardware environment. In the future, it is essential to break through the fixed input and processing paradigm under von Neumann's architecture for the computer and vigorously develop interdisciplinary intelligent computing and bionics. Design at the algorithm level, break through the existing architecture's limitations and try more flexible and human-like data processing with lower computational cost and hardware design. It is also important to develop new component design schemes with high-performance and low energy consumption to improve the computing ability and efficiency of both software and hardware to meet the rapid growth in demand and the application of intelligent computing.

%Both methods require mathematical abstraction of the principles of human intelligence into computer language. Since the mechanism of the brain has not been fully revealed, the means of machine intelligence generation, development, and computation are more complex and challenging to define. There is still a lack of effective basic intelligence mathematical models.

\subsection{Solutions to Large-Scale Computing Systems}
The theoretical-technical architecture of intelligent computing is a complex system with multiple subsystems that interact with other disciplines. Various hardware in the system requires a more complex system design, better optimization technologies, and many costs in system tuning. Lacking complexity in the theory of high-dimension computing is the main challenge for a large-scale computing system. In large-scale computing systems, the optimization problems can be simplified into multiple small tasks to reduce the system's complexity. However, no solid theoretical foundation exists in that aspect. For the optimization problem, the main target is minimizing the objective function. However, minimizing the objective function cannot capture those uncertainties when there are multiple uncertainties. Uncertainty can lead to significant variation in the system and thus increase the complexity, which is difficult to analyze. For example, in the problem of computational social science, the main objects to be modeled are groups of people. The mechanism of macro phenomena, such as ethnic group evolution and cultural transmission, can be explained by analyzing the interaction between humans and the environment. However, capturing all micro disturbances when operating such a large-scale system is problematic. Meanwhile, it is challenging to disassemble the social science computational process into multiple independent subsystems.

When finding solutions for large-scale applications, it is necessary to define the global parameters from a macro perspective at the beginning of the task. A new interactive task guidance method that introduces the role of humans in the task understanding process needs to be designed. Then, it is necessary to break the complex computing problems into subproblems and organize the problem sequence according to its logic. Finally, the results from multiple subproblems can be combined as a complete solution. Generally, there are three difficulties in solving large-scale problems. Firstly, new abstraction methods should be explored to build the micro-macro linkage causal analysis model instead of adding more parameters and building more complex subsystems. For system modules that are not directly connected, the implicit relationships between them can be revealed through dimension transformations. The problem of a large-scale complex system often involves multiple disciplines, which require prior knowledge and experimental experience from different fields to solve the general mathematical principles of their sub-problems. Second, in large-scale systems, the subsystem’s calculation mechanism is highly non-linear, which may cause competition in the computing resource in different subsystems or even constrain each other. The multi-level subsystems make it exponentially more difficult to study the non-linear effects. Studying the non-linearity of the multi-level subsystems can effective reveal the internal mechanism of large-scale systems. The last problem is the interpretability of the system. In general, the model complexity gradually grows with accuracy, and higher complexity leads to an unexplainable model. Since the large-scale system frequently exchanges information with the real world, the complex interaction between the subsystems leads to the evolution of the system structure. It is necessary to establish a new theory from a higher-order perspective to analyze its interpretability~\cite{zhang2020system}.

%%%%%%%%%%%%%%%%%%%%%%%%%%%%%%%%%%%%%%%%%%%%%%%%%%%%%%%%%%%%%%%%%%%%%%%%%%%%%%%%%%%%%%%%%%%
\section{Conclusion \label{Sec:Conclusions}}
We are currently ushering in the fourth wave of human development and are in the critical transition from the information society to the human-physics-information integration of the intelligent society. In this transition, computing technologies are undergoing transformative, even disruptive, changes. Intelligent computing is believed to be the future direction for computing, not only intelligence-oriented computing but also intelligence-empowered computing. It will provide universal, efficient, secure, autonomous, reliable, and transparent computing services to support large-scale and complex computational tasks in today's smart society. This paper presents a comprehensive review of intelligent computing, covering its theory fundamentals, the technological fusion of intelligence and computing, important applications, challenges, and future directions. We hope this review provides a good reference to researchers and practitioners and fosters future theoretical and technological innovations in intelligent computing.

%In the face of the challenges as well as the opportunities posed by intelligent computing, Zhejiang Lab has also laid out the blueprint for its comprehensive strategic research endeavors in intelligent computing, covering the research on the fundamental theories, platforms, and standards of intelligent computing. Notably, the ``digital reactor" scientific equipment project, a flagship project in intelligent computing, has been launched by the lab recently, aiming to construct large-scale equipment platforms of intelligent computing to support various high-impact interdisciplinary scientific and social applications.

\section*{Acknowledgement}
This work was partially supported by the Key Research Project (No. K2022PD1BB01, No. 2020NB2GA01, No. U21A20488, No. 2022PI0AC01), the Exploratory Research project (No. 2022KG0AN01), and the Center-initiated Research Project (No. 2022ME0AL02) of Zhejiang Lab, the National Key Research and Development Program of China (No. 2022YFB4500300), the National Natural Science Foundation of China under Grant (No. 62271452, No. 62172372), and the Natural Science Foundation of Zhejiang Province, China (No. LZ21F030001).

\printbibliography

\end{document}